\documentclass[twoside,11pt]{article}

\usepackage{amsmath}
\usepackage{amssymb}
\usepackage{amsthm}  %
\usepackage[colorlinks=false,allbordercolors={1 1 1}]{hyperref}  %
\usepackage[capitalise,nameinlink]{cleveref}   %
\usepackage[nohyperref, preprint]{jmlr2e}

\usepackage[USenglish]{babel}

\usepackage{mathtools}
\usepackage{comment}
\usepackage[shortlabels]{enumitem}
\usepackage{csquotes}
\usepackage{tikz}
\usepackage{tikz-qtree}
\usetikzlibrary{trees,positioning,shapes}
\usetikzlibrary{arrows.meta}
\usepackage{pgfplots}
\usepackage{xcolor,colortbl}

\usepackage{chngcntr}
\usepackage{stmaryrd}
\usepackage{dsfont}  

\usepackage{marginnote}

\usepackage[retainorgcmds]{IEEEtrantools}

\pgfplotsset{compat=1.11}

\newlength\Origarrayrulewidth

\newtheorem{assumption}[theorem]{Assumption}

\providecommand{\BlackBox}{\rule{1.5ex}{1.5ex}}  

\renewenvironment{proof}[1][\proofname]{\par
  \pushQED{\qed}%
  \normalfont\noindent{\bf #1\ }
}{%
  \popQED
}
\providecommand{\proofname}{Proof}

\newcommand{\bbE}{\mathbb{E}}

\newcommand{\bbN}{\mathbb{N}}

\newcommand{\bbR}{\mathbb{R}}

\newcommand{\bbone}{\mathds{1}}

\newcommand{\calF}{\mathcal{F}}

\newcommand{\calL}{\mathcal{L}}

\newcommand{\calN}{\mathcal{N}}

\newcommand{\calS}{\mathcal{S}}

\newcommand{\calU}{\mathcal{U}}

\providecommand{\bfa}{\boldsymbol{a}}
\providecommand{\bfb}{\boldsymbol{b}}

\providecommand{\bfe}{\boldsymbol{e}}
\providecommand{\bff}{\boldsymbol{f}}

\providecommand{\bfu}{\boldsymbol{u}}
\providecommand{\bfv}{\boldsymbol{v}}
\providecommand{\bfw}{\boldsymbol{w}}
\providecommand{\bfx}{\boldsymbol{x}}
\providecommand{\bfy}{\boldsymbol{y}}
\providecommand{\bfz}{\boldsymbol{z}}

\providecommand{\bfA}{\boldsymbol{A}}
\providecommand{\bfB}{\boldsymbol{B}}
\providecommand{\bfC}{\boldsymbol{C}}
\providecommand{\bfD}{\boldsymbol{D}}

\providecommand{\bfF}{\boldsymbol{F}}
\providecommand{\bfG}{\boldsymbol{G}}
\providecommand{\bfH}{\boldsymbol{H}}
\providecommand{\bfI}{\boldsymbol{I}}

\providecommand{\bfK}{\boldsymbol{K}}

\providecommand{\bfM}{\mathbf{M}}

\providecommand{\bfP}{\boldsymbol{P}}
\providecommand{\bfQ}{\boldsymbol{Q}}

\providecommand{\bfU}{\boldsymbol{U}}

\providecommand{\bfW}{\boldsymbol{W}}
\providecommand{\bfX}{\boldsymbol{X}}

\providecommand{\bfZ}{\boldsymbol{Z}}

\providecommand{\bftheta}{\boldsymbol{\theta}}

\providecommand{\bftau}{\boldsymbol{\tau}}

\providecommand{\bfSigma}{\boldsymbol{\Sigma}}

\providecommand{\bfone}{\boldsymbol{1}}

\providecommand{\bfzero}{\boldsymbol{0}}

\newcommand{\diff}{\,\mathrm{d}}
\newcommand{\quot}[1]{\enquote{#1}}

\newcommand{\equalDef}{\coloneqq}
\newcommand{\defEqual}{\eqqcolon}

\DeclareMathOperator{\sgn}{sgn}
\DeclareMathOperator{\tr}{tr}
\DeclareMathOperator{\Beta}{Beta}

\DeclareMathOperator{\Span}{Span}
\DeclareMathOperator{\Var}{Var}
\DeclareMathOperator{\cond}{cond}
\DeclareMathOperator{\eig}{eig}

\newcommand{\ovl}{\overline}
\newcommand{\udl}{\underline}

\allowdisplaybreaks

\newcommand{\lmin}{\lambda_{\mathrm{min}}}
\newcommand{\lmax}{\lambda_{\mathrm{max}}}

\newcommand{\vopt}{\bfv^{\mathrm{opt}}}
\newcommand{\tildevopt}{\tilde{\bfv}^{\mathrm{opt}}}
\newcommand{\popt}{p^{\mathrm{opt}}}
\newcommand{\qopt}{q^{\mathrm{opt}}}

\newcommand{\matleq}{\preceq}
\newcommand{\matgeq}{\succeq}
\newcommand{\matless}{\prec}
\newcommand{\matgr}{\succ}

\newcommand{\tGw}{\tilde{\bfG}^{\mathrm{w}}}
\newcommand{\tGab}{\tilde{\bfG}^{\mathrm{ab}}}
\newcommand{\tGwab}{\tilde{\bfG}^{\mathrm{wab}}}
\newcommand{\Gw}{\bfG^{\mathrm{w}}}
\newcommand{\Gab}{\bfG^{\mathrm{ab}}}
\newcommand{\Gwab}{\bfG^{\mathrm{wab}}}

\newcommand{\dQa}{p_{\mathrm{a}}}
\newcommand{\dQw}{p_{\mathrm{w}}}
\newcommand{\Za}{Z_{\mathrm{a}}}
\newcommand{\Zw}{Z_{\mathrm{w}}}
\newcommand{\tZa}{\tilde{Z}_{\mathrm{a}}}
\newcommand{\tZw}{\tilde{Z}_{\mathrm{w}}}

\newcommand{\bdQwa}{B_Z^{\mathrm{wa}}}

\newcommand{\vara}{\Var(\Za)}
\newcommand{\varw}{\Var(\Zw)}

\newcommand{\Aref}{\bfA^{\mathrm{ref}}}

\newcommand{\Pdata}{P^{\mathrm{data}}}
\newcommand{\tPdata}{\tilde{P}^{\mathrm{data}}}

\newcommand{\Pinitm}{P^{\mathrm{init}}_m}

\newcommand{\CP}{C_P}
\newcommand{\CW}{C_{\mathrm{weights}}}
\newcommand{\Ch}{C_{\mathrm{lr}}}

\newcommand{\Cx}{K_{\mathrm{data}}}
\newcommand{\Cpsi}{K_\psi}
\newcommand{\CM}{K_M}
\newcommand{\Cn}{K_{\mathrm{param}}}

\newcommand{\gx}{\gamma_{\mathrm{data}}}
\newcommand{\gpsi}{\gamma_\psi}
\newcommand{\gP}{\gamma_P}

\setlist[enumerate]{nosep}
\setlist[itemize]{nosep}

\makeatletter
\newcommand\ackname{Acknowledgements}
\if@titlepage
   \newenvironment{acknowledgements}{%
       \titlepage
       \null\vfil
       \@beginparpenalty\@lowpenalty
       \begin{center}%
         \bfseries \ackname
         \@endparpenalty\@M
       \end{center}}%
      {\par\vfil\null\endtitlepage}
\else
   
\fi
\makeatother

\newlength{\fixboxwidth}
\hypersetup{ hidelinks }

\usepackage{lastpage}

\jmlrheading{23}{2022}{1-\pageref{LastPage}}{7/20; Revised
2/22}{5/22}{20-830}{David Holzmüller and Ingo Steinwart}

\ShortHeadings{Inconsistency of Neural Networks}{Holzmüller and Steinwart}

\firstpageno{1}

\begin{document}

\title{Training Two-Layer ReLU Networks with Gradient Descent is Inconsistent}

\author{\name David Holzmüller \email david.holzmueller@mathematik.uni-stuttgart.de \\
       \addr University of Stuttgart\\
       Faculty of Mathematics and Physics\\
       Institute for Stochastics and Applications
       \AND
       \name Ingo Steinwart \email ingo.steinwart@mathematik.uni-stuttgart.de \\
       \addr University of Stuttgart\\
       Faculty of Mathematics and Physics\\
       Institute for Stochastics and Applications}

\editor{Yoshua Bengio}

\maketitle

\begin{abstract}%
We prove that two-layer (Leaky)ReLU networks initialized by e.g.\ the widely used method proposed by \citet{he_delving_2015} and trained using gradient descent on a least-squares loss are not universally consistent. 
Specifically, we describe a large class of one-dimensional data-generating distributions for which, with high probability, gradient descent only finds a bad local minimum of the optimization landscape, since it is unable to move the biases far away from their initialization at zero.
It turns out that in these cases, the found network essentially performs linear regression even if the target function is non-linear.
We further provide numerical evidence that this happens in practical situations, for some multi-dimensional distributions and that stochastic gradient descent exhibits similar behavior. We also provide empirical results on how the choice of initialization and optimizer can influence this behavior.
\end{abstract}


\section{Introduction}

In recent years, 
neural networks (NNs) have achieved various success stories in areas such
as image classification and natural language processing. 
For this reason, NNs are commonly viewed as one of the state-of-the-art machine 
learning algorithms. Unlike for other machine learning algorithms, however,
our theoretical understanding of their learning behavior, e.g.~in terms
of a-priori learning guarantees, is still rather limited.

We consider the classical setting of statistical learning theory, where a random data set $D = ((x_1, y_1), \hdots, (x_n, y_n))$ consists of $n$ i.i.d.\  pairs $(x_j, y_j)$ sampled from an unknown probability distribution $\Pdata$ on $\bbR^d\times \bbR$. In our case, we consider the empirical,   respectively population   least-squares risk of a function $f: \bbR^d \to \bbR$, that is, the quantities
\begin{IEEEeqnarray}{+rCl+x*}\label{def-risks}
R_D(f) & \equalDef & \frac{1}{2n} \sum_{j=1}^n (y_j - f(x_j))^2\qquad \mbox{ and}  \qquad R_{\Pdata}(f) \equalDef \frac{1}{2} \bbE_{(x, y) \sim \Pdata} (y - f(x))^2~.
\end{IEEEeqnarray}
The Bayes risk is the lowest possible risk, which might be nonzero due to noise in the $y$ component, that is
\begin{IEEEeqnarray*}{+rCl+x*}
R_{\Pdata}^* & \equalDef & \inf_{f: \bbR^d \to \bbR} R_{\Pdata}(f)~.
\end{IEEEeqnarray*}
It is well known that
this infimum is achieved by the conditional mean function, i.e., by
 $f_{\Pdata}^*(x) \equalDef \bbE_{\Pdata}(Y|X=x)$. 
 Now, a
 learning method,
 i.e., a method that assigns  a function $f_D$ to each data set $D$, is called consistent for a distribution $\Pdata$, if its population risk converges in probability to the Bayes risk as the number $n$ of samples goes to infinity. Or, to phrase it more formally, if  for each $\varepsilon > 0$, the probability of observing a \quot{bad} data set $D$ with
\begin{IEEEeqnarray*}{+rCl+x*}
R_{\Pdata}(f_D) - R_{\Pdata}^* & \geq & \varepsilon
\end{IEEEeqnarray*}
converges to zero for $n\to \infty$. The quantity $R_{\Pdata}(f_D) - R_{\Pdata}^*$ is also called \emph{excess risk}. Arguably the simplest a-priori guarantee for a learning method is the notion of universal consistency, which requires the learning method  to be consistent for all bounded, or more generally, for all distributions $\Pdata$ with $R_{\Pdata}(0)< \infty$. For simple learning methods such as histogram rules, kernel regression, and $k$-nearest neighbor rules, universal consistency has has been long known, see for example the books by \citet{devroye_probabilistic_1996} and \citet{gyorfi_distribution-free_2002} for a variety of universally consistent learning methods. Similarly, kernel-based based methods including support vector machines are 
universally consistent, see for example the book by \citet{steinwart_support_2008}. 

\subsection{Contribution}

We prove that training under-parameterized ReLU or Leaky ReLU networks with one hidden layer using gradient descent (GD) on a least-squares loss does not yield an universally consistent estimator if e.g.\ the common initialization method by \citet{he_delving_2015} is used. To this end, we specify a large family of one-dimensional data-generating distributions $\Pdata$, for which we prove that the probability of getting stuck in a bad local minimum converges to one as the width of the network and the number of data points %
simultaneously go to infinity in a controlled manner.
We further show that these one-dimensional distributions, when embedded into higher-dimensional spaces, also provide examples of inconsistency for NNs with multi-dimensional input. 
Moreover, we prove that there also exist a multitude of data sets 
for which over-parameterized versions of such NNs cannot be properly trained by gradient 
descent. 

We experimentally investigate the effect of deviating from our theoretically investigated assumptions by changing bias initialization, optimizer, network parameterization, and the dimensionality of the input distribution. We observe similar shortcomings of the zero bias initialization used in \citet{he_delving_2015} even with stochastic gradient descent and moderate-dimensional input distributions. In addition, our experimental results give rise to practical recommendations on ways to mitigate the investigated problems of zero bias initialization. %

This paper is an improved version of the first author's master's thesis \citep{holzmuller2019convergence}.

\subsection{Related Work} \label{sec:related_work}

Despite our result, there do exist some consistency results for certain classes of NNs, 
see e.g.~\citep{white_connectionist_1990} for regression, 
\citep{farago_strong_1993} for classification, 
as well as Chapter 30 in the book by \citet{devroye_probabilistic_1996}
and Chapter 16 in the book by  \citet{gyorfi_distribution-free_2002}.
 However, all
these results consider a training algorithm that finds a global minimum, as well as an \emph{under-parameterized regime}, in which
the number $m$ of hidden neurons grows more slowly than
the sample size. This shows that consistency does not require $m$ to grow fast, although it does require $m$ to converge to infinity such that the neural networks are able to approximate any target function $f_{\Pdata}^*$ arbitrarily well in the limit. 
Unfortunately, finding a global minimum for small network sizes 
can be NP-hard, see e.g.~the classical paper by \citet{blum_training_1989}
as well as \citet{boob_complexity_2020}, who establish similar results for 
certain ReLU-networks. 
Moreover, 
\citet{safran_spurious_2018} have empirically shown that the probability of finding a bad local minimum in certain
two-layer ReLU NNs with gradient descent can increase with increasing numbers of samples and neurons. Finally, coming from another angle, \cite{lee_consistency_2000} 
established a consistency result for Bayesian inference over NN parameters, which comes with a huge computational overhead. 
Consequently, it remains possible that the consistency results mentioned above only apply to computationally 
infeasible NN training algorithms.
On the other hand, a consistent NN training algorithm does not necessarily 
need to find a global optimum, and hence it also remains an open question 
whether practical NN algorithms such as variants of 
(stochastic) GD are consistent in the under-parameterized regime when our assumptions are not satisfied.
In this respect we finally note that the described theoretical gap resulting from the difficulty of solving the 
optimization problem is somewhat specific to neural networks:
For example,
beginning with \cite{yao_early_2007}
optimizing kernel methods with gradient descent and early stopping has been well understood in the sense that e.g.~both finite sample 
guarantees and learning rates, i.e., convergence rates for the excess risk, have been established.

Based on the empirical observations by \citet{zhang_understanding_2017}, 
it has been more recently  shown that finding a global minimum for
\emph{over-parameterized} NNs, i.e., for NNs whose number of
neurons (significantly) exceeds the number of samples, is easier.
For example, \citet{arora_understanding_2018} present a poly-time algorithm for 
finding a global minimum for NNs with one hidden layer  and \citet{mucke_empirical_2019}
present a quadratic time training algorithm for NNs with two hidden layers.
However, both algorithms are not based on (stochastic) GD.
By imposing rather mild assumptions 
on the data set, \citet{du_gradient_2019, du_gradient_2019-1}, \citet{allen-zhu_convergence_2019}, and \citet{zou_stochastic_2018}
show that 
(stochastic) GD also reaches a global optimum with high probability. Their analysis is based on the so-called  neural tangent kernel (NTK) approach proposed by  \citet{jacot_neural_2018}.
Many subsequent works have established bounds requiring a smaller amount of overparameterization, see e.g.~the papers by \citet{zou_improved_2019, ji_polylogarithmic_2019, song_quadratic_2019, ge_mildly_2019, chen_how_2021}. While these papers address the optimization problem of NNs, generalization guarantees such as consistency are either not discussed, or are only established 
under special assumptions on the data-generating distribution. In fact, 
\citet{mucke_empirical_2019} show that over-parameterized NNs of sufficient size
always have both global minima with good generalization performance in the sense of 
consistency
and global minima with very bad generalization performance. 
However, their global minima are
more interesting from a theoretical than a practical
point of view. 
Finally, 
 \citet{zhang_understanding_2017} and \citet{belkin_overfitting_2018}
discuss in detail why common techniques from statistical learning theory
cannot work for the analysis of over-parameterized NNs. 

There are some partial results regarding consistency of over-parameterized networks. \citet{arora_fine-grained_2019} present a generalization bound for certain ReLU-NNs
with one hidden layer
that involves a novel complexity measure in 
terms of a Gram matrix of a neural tangent kernel.
However, their results are only stated for a special class of \emph{noise-free} data-generating distributions. \citet{cao_generalization_2020} investigate a similar scenario for deeper networks. 
\cite{belkin_overfitting_2018} have argued that generalization properties of unregularized kernel methods should be studied, which has become even more interesting since \cite{arora_exact_2019} have shown that regression with sufficiently wide neural networks essentially behaves like unregularized kernel regression with the NTK.
For the Laplace kernel, \cite{rakhlin_consistency_2019} have shown inconsistency of unregularized kernel regression for a realistic class of data-generating distributions \emph{with label noise.} On the contrary, \cite{liang_multiple_2020} show that for NTK-like kernels, the excess risk can converge to zero, if one considers a sequence of \quot{uniformly easy} learning problems for which the dimension increases with the number of data points. 
Clearly, 
their setting is different from classical statistical learning questions such as 
universal consistency, and although label noise is permitted, very strong assumptions are imposed on the optimal regression function given by $\Pdata$. 
In summary, 
it can thus be hypothesized that very wide (over-parameterized) unregularized neural networks are not universally consistent, i.e., their excess risk does not converge to zero with increasing number of samples, although the excess risk may become very small if the input-dimension is high.

Similar to us, 
\cite{williams_gradient_2019} analyze two-layer ReLU networks with one-dimensional input. They define certain quantities $\delta_i$ that are based on the network weights and identify a \quot{kernel} regime for $\delta_i \to -\infty$ and an \quot{adaptive} regime for $\delta_i \to \infty$. They analyze the NN behavior in these idealized regimes and argue heuristically, why NNs that are close to these regimes should yield solutions similar to the idealized solutions. In our case, the NN behaves similar to their \quot{kernel} regime with a finite-rank kernel, although a simple statistical analysis reveals that $\delta_i = -O(1)$ for most $i$ and $\delta_i > 0$ for some $i$.

Finally,
\citet{steinwart_sober_2019} observed experimentally that
on one-dimensional data sets, 
two-layer ReLU NNs with the initialization method of \cite{he_delving_2015} often get stuck in bad local minima, 
and this was the initial motivation for our theoretical analysis. 
However, 
the existence of local minima in the loss landscape of neural networks has been investigated before in various settings with various results, see e.g.\ the papers by \cite{sontag_backpropagation_1989, gori_problem_1992, soudry_no_2016, yun_small_2019, fukumizu_local_2000, safran_spurious_2018, he_piecewise_2019}. Except for some experiments in \cite{safran_spurious_2018}, 
however,
the probability of reaching a spurious local minimum is not investigated in these works. In this respect it is interesting to note that \cite{yun_small_2019} investigate a type of spurious local minimum that is similar to the ones that are found by NNs in our scenario. Moreover,  \cite{he_piecewise_2019} prove the existence of similar local minima in a fairly general 
setting that includes deeper architectures and other loss functions.

The rest of this work is organized as follows: In \Cref{sec:setup}, we define our considered NN architecture and training process. In \Cref{sec:overview}, we then present some intuition behind our result. Our main theorem is then presented in \Cref{sec:main_result}, with applications to the over-parameterized case in \Cref{sec:over-parameterization} and the under-parameterized (inconsistent) case in \Cref{sec:inconsistency}. We outline the ideas behind the proof of our main theorem in \Cref{sec:proof_idea} and discuss the relation to NTK-based analyses in \Cref{sec:ntk}. We support our theory with experimental evidence in \Cref{sec:experiments}, provide further experiments on ways to resolve the problems discussed in this paper in \Cref{sec:practical}, and discuss further research questions in \Cref{sec:conclusion}. Most proofs are deferred to the appendix.

\section{Neural Network Architecture and Training} \label{sec:setup}

In this section we present the considered network architecture, the initialization, and the training process. 
For simplicity, we will mostly focus on one-dimensional inputs, but show in \Cref{rem:main:multi-d} that our results for one-dimensional inputs can be easily generalized to multi-dimensional inputs.

\begin{definition} \label{def:nn}
Let $\varphi: \bbR \to \bbR$ be the Leaky ReLU function with fixed parameter $\alpha \in \bbR \setminus \{\pm 1\}$, that is 
\begin{IEEEeqnarray*}{+rCl+x*}
\varphi(x) & \equalDef & \begin{cases}
x &, x \geq 0 \\
\alpha x &, x \leq 0~.
\end{cases} %
\end{IEEEeqnarray*}
We consider two-layer single-input single-output neural networks with $m \in \bbN$ hidden neurons. Such a neural network defines a function $f_W: \bbR \to \bbR$ via
\begin{IEEEeqnarray}{+rCl+x*}\label{def-NN-function}
f_W(x) & \equalDef & c + \sum_{i=1}^m w_i \varphi(a_i x + b_i)~,
\end{IEEEeqnarray}
where $W = (\bfa, \bfb, c, \bfw) \in \bbR^{3m+1}$ with $\bfa, \bfb, \bfw \in \bbR^m$ and $c \in \bbR$. 
Note  that  $\varphi$ is piecewise linear with a \emph{kink} (non-differentiable point) at $0$, and therefore the function $f_W$ is piecewise affine linear with potential \emph{kinks} at %
$-b_i / a_i$.
\end{definition}

\begin{assumption}[Initialization] \label{ass:init}
The components of the initial parameter vector $W_0$ of \eqref{def-NN-function}
are initialized independently with distributions
\begin{IEEEeqnarray*}{+rCl+x*}
b_{i, 0} & = & 0, \quad c_0 = 0, \quad a_{i, 0} \sim \Za, \quad w_{i, 0} \sim \frac{1}{\sqrt{m}} \Zw~,
\end{IEEEeqnarray*}
where $\Za, \Zw$ are $\bbR$-valued random variables satisfying:
\begin{enumerate}[(Q1), wide=0pt, leftmargin=*]

\item $\Za$ and $\Zw$ have symmetric  and bounded probability density functions $\dQa, \dQw: \bbR \to [0, \bdQwa]$,
where $\bdQwa$ is a suitable constant.
\item $\bbE |\Za|^p < \infty$ and $\bbE |\Zw|^p < \infty$ for all $p \in (0, \infty)$.
\end{enumerate}
We denote the distribution of $W_0$ by $\Pinitm$.
\end{assumption}

Importantly, \Cref{ass:init} is satisfied, for example, by the initialization method of \citet{he_delving_2015}, where $\Za, \Zw \sim \calN(0, 2)$. Our assumption is also satisfied for e.g.~uniformly distributed
$\Za$ and $\Zw$. Moreover, recall that the assumed zero bias initialization is 
 the default in both \texttt{Tensorflow} and \texttt{Keras}.
Finally, note that $\Za$ and $\Zw$ do not necessarily need to have distributions that belong to the same family of distributions.

\begin{assumption}[Data distribution I] \label{ass:pdata-1}
 Let $\Pdata$ be a distribution on $\bbR\times \bbR$ for which all moments are finite, that is 
 \begin{align*}
  \int |x|^p + |y|^p \diff \Pdata(x, y) < \infty
 \end{align*}
for all $p \in (0, \infty)$. Note that this is in particular satisfied if $\Pdata$ is \emph{bounded},
 that is, if there exist bounded (measurable) subsets $X,Y\subset \bbR$ with $\Pdata(X\times Y) = 1$.
 Finally,  $\Pdata_X$ denotes the marginal distribution of $\Pdata$ with respect to the $x$-component. 
\end{assumption}

\begin{definition} \label{def:loss}
For a weight vector $W \in \bbR^{3m+1}$, 
a data set $D = ((x_1, y_1), \hdots, (x_n, y_n)) \in (\bbR \times \bbR)^n$, and 
a bounded distribution
$P$ on $\bbR \times \bbR$, 
we define the empirical, respectively population
(least-squares) loss 
of the function $f_W: \bbR \to \bbR$ in \eqref{def-NN-function}
by %
\begin{IEEEeqnarray*}{+rClrCl+x*}
L_D(W) & \equalDef & R_D(f_W) \qquad \mbox{ and}  \qquad 
L_P(W) \equalDef R_P(f_W)  \, ,
\end{IEEEeqnarray*}
where the risks $R_D(f_W)$ and $R_P(f_W)$ have already been introduced in \eqref{def-risks}.  
\end{definition}

\begin{assumption}[Training] \label{def:gd}
For $m, n \in \bbN$, step size $h > 0$, a distribution $\Pinitm$ as in \Cref{ass:init}, and a data set $D \in (\bbR \times \bbR)^n$, training the neural network \eqref{def-NN-function} produces a random sequence of functions $(f_{W_k})_{k \in \bbN_0}$ via initializing $W_0 \sim \Pinitm$ and applying GD, i.e., for 
all $k\geq 0$ we have 
\begin{IEEEeqnarray*}{+rCl+x*}
W_{k+1} & \equalDef & W_k - h\nabla L_D(W_k)~. & \qedhere
\end{IEEEeqnarray*}
\end{assumption}

Note that for a \emph{fixed} data set $D$ the randomness of the training sequence $(f_{W_k})_{k \in \bbN_0}$ 
only comes from the initialization. When we investigate inconsistency, however, we also need to 
view the data set $D$ as a random variable.

\begin{remark}[Multi-dimensional input] \label{rem:main:multi-d}
We show in \Cref{sec:multi-d} that NNs with multi-dimensional input behave similarly in the 
following sense: For a fixed $\bfz \in \bbR^d$ with $\|\bfz\|_2 = 1$ and $D$ as above, consider the data set $\tilde{D} \equalDef ((x_1 \bfz, y_1), \hdots, (x_n \bfz, y_n)) \in (\bbR^d \times \bbR)^n$,
that is, we embed $D$ into $\bbR^d$ by mapping all $x_j$ to a point on the line 
 $\Span \{\bfz\}$ without changing the mutual  distances between the samples. Similarly, 
 a distribution $\Pdata$ on $\bbR\times \bbR$ can be mapped to $\bbR^d\times \bbR$
 by considering the image measure $\tPdata$
 of $\Pdata$ under the mapping $(x,y)\mapsto (x\bfz, y)$. Note that the marginal distribution $\tPdata_X$
 of such an image measure lives on a one-dimensional (linear)
 manifold in $\bbR^d$, and if its   conditional mean function is smooth, 
 conventional wisdom thus suggests that it is particularly easy to learn from such a probability measure $\tPdata$. In this respect note that the distributions 
 $\Pdata$ on which inconsistency occurs, see 
 \Cref{ass:P} for a detailed description, 
 include distributions with arbitrarily smooth 
 conditional mean function and by the transformation described above, the same applies to 
 $\tPdata$.

 Now
consider a neural network with multivariate input $\bfx \in \bbR^d$ defined by
\begin{IEEEeqnarray*}{+rCl+x*}
f_{\tilde{W}}(\bfx) & = & \tilde{c} + \sum_{i=1}^m \tilde{w}_i \varphi \left(\tilde{b}_i + \sum_{j=1}^d \tilde{a}_{ij} x_j\right)~,
\end{IEEEeqnarray*}
where $\tilde{W} = (\tilde{\bfa}, \tilde{\bfb}, \tilde{c}, \tilde{\bfw}) \in \bbR^{(d+2)n + 1}$ are the parameters of the NN. Again, we assume that the NN is trained using gradient on the same loss function and initialized via
\begin{IEEEeqnarray*}{+rCl+x*}
\tilde{b}_{i, 0} & = & 0, \quad \tilde{c}_0 = 0, \quad \tilde{a}_{i, j, 0} \sim \Za, \quad \tilde{w}_{i, 0} \sim \frac{1}{\sqrt{m}} \Zw~,
\end{IEEEeqnarray*}
where $\Za$ and $\Zw$ satisfy the conditions in \Cref{ass:init}. Again, this is satisfied for example by the initialization method of \cite{he_delving_2015}.

We show in \Cref{sec:multi-d} that there exists a random $W_0 \in \bbR^{3n+1}$ satisfying \Cref{ass:init} such that the GD iterates
\begin{IEEEeqnarray*}{+rCl+x*}
\tilde{W}_{k+1} & = & \tilde{W}_k - h\nabla L_{\tilde{D}}(\tilde{W}_k), \\
W_{k+1} & = & W_k - h\nabla L_D(W_k)
\end{IEEEeqnarray*}
satisfy
\begin{IEEEeqnarray*}{+rCl+x*}
f_{\tilde{W}_k}(x \bfz) = f_{W_k}(x)~.
\end{IEEEeqnarray*}
for all $x \in \bbR, k \in \bbN_0$. In other words, if we have a data set where all $x_j$ lie on a line $\Span \{\bfz\} \subseteq \bbR^d$, then an NN on the $d$-dimensional input space behaves on this line like an NN with one-dimensional input space.
\end{remark}

\section{A Closer Look at the Training Behavior} \label{sec:overview}

In this section we illustrate why 
the combination of zero bias initialization and GD training  
may produce poorly predicting NNs. 
In the subsequent sections we then rigorously prove that 
the illustrated training behavior does occur with high probability. 

\begin{figure}[bt]
\centering
\includegraphics[scale=1.0]{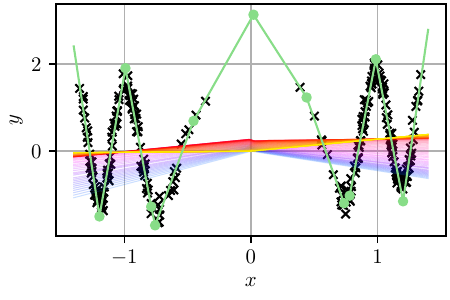}
\caption{A data set with $n=300$ samples (black crosses), a close-to-optimal NN predictor (green line, kinks marked as circles) with $m = 16$, and functions $f_{W_k}$ for $k = \lfloor 1.1^l \rfloor - 1, l \in \{0, \hdots, 120\}$, $m = 16$, $h = 0.002$, 
trained on the visualized data set. The colors of $f_{W_k}$ transition from blue to red to yellow, i.e., the lower blue function is $f_{W_0}$ and the upper red line is reached at an intermediate stage before the NN converges to the prominent yellow line.} \label{fig:nn_convergence_fused}
\end{figure}

Our first, rather basic observation is that zero bias initialization $b_{i,0}=0$  and $c_0=0$ as in 
\Cref{ass:init} places all kinks $-b_{i,0}/a_{i,0}$ of the initial $f_{W_0}$
at zero. Consequently, $f_{W_0}$ is  linear on both $(-\infty,0]$ and $[0,\infty)$.
In contrast, the function to be learned is typically nonlinear on these two sets, and 
finding a suitable NN approximation during training may thus require to substantially 
move at least some of the kinks. To illustrate this statement, consider 
\Cref{fig:nn_convergence_fused},
in which a data set that requires a nonlinear predictor is depicted.
It is obvious from 
\Cref{fig:nn_convergence_fused}
 that any reasonable NN approximation requires at least a few kinks in both $[-1.5, -0.5]$ and 
$[0.5, 1.5]$. The training algorithm thus needs to move a few kinks from 0 into these two areas. 
Unfortunately, such a behavior can in general  not be guaranteed. For example,
\Cref{fig:loss_and_kinks} illustrates that on the data set $D$ of 
\Cref{fig:nn_convergence_fused},
GD does not move the kinks outside the interval $[-0.2, 0.2]$ and in particular,
no kink is moved across a sample, since no $x$-component of a sample of $D$ falls inside $[-0.2, 0.2]$.
As a consequence, the corresponding NN predictors $f_{W_k}$, for $k\geq 1$, 
remain affine linear on the  left  part $D_{-1}$ and the 
right  part $D_1$   of the data set $D$,
as 
\Cref{fig:nn_convergence_fused}
illustrates. This problem has been observed experimentally by \citet{steinwart_sober_2019}, who suggests to use a data-dependent initialization method that places kinks randomly in between the data points.

A closer look at 
\Cref{fig:nn_convergence_fused}
further indicates that 
$f_{W_k}$ approaches the optimal affine linear regression lines 
for the two data sets $D_{-1}$ and $D_{1}$ with increasing $k$. As a consequence, GD gets stuck in a 
bad local minimum of the loss surface.
Note that the existence of structurally similar bad
local minima has already been shown in \citet{yun_small_2019}, but there
it remained an open question whether GD can avoid such bad minima.
In the following sections, we rigorously show that  
under some 
assumptions on $D$ or $P$, on the step size $h$, and on $n$ and $m$,
the predictors  $f_{W_k}$ produced by 
GD remain affine linear on the negative and positive 
parts of $D$ with high probability. Consequently, GD does not escape from the corresponding 
bad local minima in such situations.

\begin{figure}[bt]
\centering

\input{loss_and_kinks_short.pgf} 
\input{loss_and_kinks.pgf}

\caption{This figure shows the evolution of several quantities during training in 
\Cref{fig:nn_convergence_fused}, where the two plots show a different number of epochs.
The thick red line shows $L_D(W_k) - \inf_{k' \in \bbN_0} L_D(W_{k'})$, where the latter infimum is the minimal loss that can be reached by fitting $D_1$ and $D_{-1}$ with linear regression similar to the yellow line in 
\Cref{fig:nn_convergence_fused}.
The thin black lines show the 16 NN kinks in 
\Cref{fig:nn_convergence_fused}
(one can see only 14 kinks since there are two almost identical pairs of paths). The right $y$-axis in this plot corresponds to the $x$-positions of the kinks in 
\Cref{fig:nn_convergence_fused}. 
} \label{fig:loss_and_kinks}
\end{figure}

\Cref{fig:loss_and_kinks} also shows that the kinks initially move very fast but then slow down.
Empirically, 
this slowdown is related to the loss, whose evolution relative to the reached optimum in our example is also shown in \Cref{fig:loss_and_kinks}. 
In our theoretical analysis, we find that the convergence of the neural network and the movement of the kinks are related to a four-dimensional linear iteration equation. Here, the system matrix has two eigenvalues of order $-\Theta(m)$ leading to a fast convergence and two eigenvalues of order $-\Theta(1)$ leading to a slow convergence. In the example of this section,
the initialization of this system is close to the fast eigenvectors, which leads to the two-step decay of the loss in \Cref{fig:loss_and_kinks}. 

Our theory suggests the following intuitive explanation of this two-step decay: Since the weights in the second layer have a lower standard deviation ($O(1/\sqrt{m})$) than the weights in the first layer, updates to second-layer weights have a stronger effect and the gradient with respect to second-layer weights is larger.  Since the first-layer biases are initially zero, the faster learning of second-layer weights can only adjust the slopes of the affine network parts. Adjusting the first-layer biases and the single second-layer bias happens at a substantially slower speed. In cases where almost no bias adjustment is needed for learning the optimal affine regression lines, learning in the second layer is so fast that kinks do not succeed in moving far. The relative training speeds of the layers can be adjusted, for example, using the NTK parameterization \citep{jacot_neural_2018}, which we further discuss in \Cref{sec:practical}.

A necessary condition for consistency is that the NN is able to approximate the target function $f_{\Pdata}^*$ arbitrarily well. In general, this requires the width $m = m_n$ of the NN, which can depend on the number of samples $n$, to converge to infinity:
\begin{IEEEeqnarray*}{+rCl+x*}
\lim_{n \to \infty} m_n = \infty~.
\end{IEEEeqnarray*}
Note that wide networks are more likely to have neurons with very small $|a_i|$, for which the kinks $-b_i/a_i$ can move fast. Because of this, we find that in order to prove the failure case described in this section, the intercept of the optimal affine regression lines needs to be closer to zero the more neurons are used. When considering the setting of consistency, where the variance of this intercept decreases with the size of the randomly sampled data set, our negative results only apply to the under-parameterized regime $m_n < o(n)$. As explained in \Cref{sec:related_work}, this is also the regime where universal consistency results are known for NNs with perfect optimization. For suitable fixed data sets, our negative results also extend to the over-parameterized regime $m_n > n$.

The insights above suggest that making the first layer learn faster via a different initialization / parameterization and initializing the kinks randomly should improve the training behavior of neural networks. Indeed, \cite{du_gradient_2019-1} consider such a setting and show convergence to a global optimum with high probability. However, their result is not sufficient for universal consistency since only the training error is analyzed. In \Cref{sec:practical}, we provide further experimental results on the benefits of such modifications.

\section{Main Result} \label{sec:main_result}

Before we can state the main result of our work, we need to introduce some more notions.

\begin{definition} \label{def:D_sigma}
For given data set $D = ((x_1, y_1), \hdots, (x_n, y_n)) \in (\bbR \times \bbR)^n$, the subsequence  of samples $(x_j, y_j)$ with $x_j > 0$ is denoted by $D_1$. 
Analogously, $D_{-1}$ denotes the samples with $x_j < 0$.
Moreover, for a 
distribution $\Pdata$, we define the measures $P_1$ and $P_{-1}$ via
\begin{IEEEeqnarray*}{+rCl+x*}
P_1(E) & \equalDef & \Pdata(E \cap ((0, \infty) \times \bbR)) \\
P_{-1}(E) & \equalDef & \Pdata(E \cap ((-\infty, 0) \times \bbR))~. & \qedhere
\end{IEEEeqnarray*}
Throughout this work,  we require $D_1$ and $D_{-1}$ to be non-empty and $P_1$ and $P_{-1}$
to be non-trivial.
However, it is also possible to prove analogous results for the cases where $D$ or $\Pdata$ are solely concentrated on $(-\infty,0)$ or $(0,\infty)$, cf.\ \Cref{rem:alternative_systems}.
\end{definition}

With the help of $D_{\pm 1}$ and $P_{\pm 1}$ 
we can now define the \quot{half-sided  linear regression optima}, which will be crucial for our analysis.

\begin{definition} \label{def:main:quantities}
For $P:=\Pdata$ and $D$ as in \Cref{def:loss}, $x, y \in \bbR$, and   $\sigma \in \{\pm 1\}$, we define
\begin{IEEEeqnarray*}{+rClrCl+x*}
\bfM_x & \equalDef & \begin{pmatrix}
x^2 & x \\
x & 1
\end{pmatrix}, \qquad \qquad \qquad\qquad& \hat{\bfu}^0_{(x, y)} & \equalDef & \begin{pmatrix}
xy \\
y
\end{pmatrix}  ,\\
\bfM_{D, \sigma} & \equalDef & \frac{1}{n} \sum_{(x, y) \in D_\sigma} \bfM_x, & \hat{\bfu}^0_{D, \sigma} & \equalDef & \frac{1}{n} \sum_{(x, y) \in D_\sigma} \hat{\bfu}^0_{(x, y)} , \\
\bfM_{P, \sigma} & \equalDef & \int \bfM_x \diff P_\sigma(x, y), & \hat{\bfu}^0_{P, \sigma} & \equalDef & \int \hat{\bfu}^0_{(x, y)} \diff P_\sigma(x, y).
\end{IEEEeqnarray*}

Moreover, if the $2\times 2$ matrix $\bfM_{D, \sigma}$ is invertible, 
see \Cref{rem:MD-invert} for a simple characterization of this situation,
we write
\begin{IEEEeqnarray*}{+rCl+x*}
\vopt_{D, \sigma} & \equalDef & \begin{pmatrix}
\popt_{D, \sigma} \\
\qopt_{D, \sigma}
\end{pmatrix} \equalDef \bfM_{D, \sigma}^{-1} \hat{\bfu}^0_{D, \sigma}\, .
\end{IEEEeqnarray*}
Note that $\popt_{D, \sigma}$ and $\qopt_{D, \sigma}$ are the 
slope and intercept of the optimal linear regression line for $D_\sigma$, see \Cref{rem:affine_regression} for a proof  
and \Cref{fig:vopt} for an illustration. 

In the sequel, we are mostly interested in the maximal absolute slopes and intercepts as well as
the distance of $D$ to $0$, i.e., 
\begin{IEEEeqnarray*}{+rCl+x*}
\psi_{D, p} & \equalDef & \max \left\{\left|\popt_{D, 1}\right|, \left|\popt_{D, -1}\right|\right\}, \\
\psi_{D, q} & \equalDef & \max \left\{\left|\qopt_{D, 1}\right|, \left|\qopt_{D, -1}\right|\right\}, \\
\underline{x}_D & \equalDef & \min \{|x_1|, \hdots, |x_n|\}~.  %
\end{IEEEeqnarray*}
If $\bfM_{P, \sigma}$ is invertible,  
the quantities $\vopt_{P, \sigma}, \qopt_{P, \sigma}, \popt_{P, \sigma}, \psi_{P, p}$ and $\psi_{P, q}$ are defined and interpreted analogously.
Finally, the smallest and the largest eigenvalue of a symmetric matrix $A$ are denoted by 
$\lmin(A)$ and $\lmax(A)$. Since the matrix $\bfM_{D, \sigma}$ is positive semi-definite, we note that 
it is invertible if and only if $\lmin(\bfM_{D, \sigma})>0$. 
\end{definition}

\begin{figure}[bt]
\centering
\begin{tikzpicture}
\begin{axis}[axis lines=middle, ticks=none, xlabel=$x$, ylabel=$y$, ymin=-3.2, ymax=2.1]
\addplot[color=orange, mark=None] coordinates {
(-7, 1) (7, 1)
};
\addplot[color=blue, mark=None] coordinates {
(-7, -2.5) (7, -0.166)
};

\addplot[only marks, mark options={scale=0.5}] coordinates {
(-6, -2.333+0.5)
(-4, -2.0-1)
(-2, -1.666+0.5)

(-5, -2.166+0.2)
(-3, -1.833-0.4)
(-1, -1.5+0.2)

(3, 1.25)
(4, 0.75)
(5, 0.75)
(6, 1.25)

(2.5, 1.0 - 0.2)
(3.5, 1.0 + 0.4)
(4.5, 1.0 - 0.2)
};

\node[anchor=north west, orange] (a) at (-7, 1) {$y = \popt_{D, 1} x + \qopt_{D, 1}$};
\node[anchor=north west, blue] (b) at (0.2, -1.1) {$y = \popt_{D, -1}x + \qopt_{D, -1}$};
\node[anchor=north west, orange] (c) at (4.2, 1.8) {$D_1$};
\node[anchor=north west, blue] (d) at (-4.7, -0.6) {$D_{-1}$};
\draw[orange, dashed] (4.2, 1) circle[radius=1.05cm];
\draw[blue, dashed] (-3.5, -1.8) circle[radius=1.4cm];

\end{axis}
\end{tikzpicture}
\caption{Optimal affine regression lines for $D_1$ (orange, upper line) and $D_{-1}$ (blue, lower line) for an example data set $D$. The data points are shown in black. The slope and intercept of the optimal affine regression line for $D_\sigma$ are given by $\popt_{D, \sigma}$ and $\qopt_{D, \sigma}$, respectively.} \label{fig:vopt}
\end{figure}

The following remark, whose proof can be found in \Cref{rem-check-MD-invert}, shows that the matrices $\bfM_{D, \pm 1}$ are invertible for all 
    interesting data sets $D$. Since the eigenvalues of $\bfM_{D, \pm 1}$
    will play an important role in our main result, the following remark also 
     provides simple estimates for the eigenvalues of 
$\bfM_{D, \pm 1}$.

\begin{remark}\label{rem:MD-invert}
Given a $\sigma \in \{\pm 1\}$, the matrix $\bfM_{D, \sigma}$ is invertible, if and only if the data set $D_{\sigma}$
contains at least two samples $(x_i,y_i)$ and $(x_j,y_j)$ with $x_i \neq x_j$.
Moreover, if $D_\sigma$ contains $n_\sigma$ samples and $D_{X, \sigma}$ are the $x$ components of the samples in $D_\sigma$, we have
\begin{IEEEeqnarray*}{+rCl+x*}
\lmin(\bfM_{D, \sigma}) & \geq & \frac{n_\sigma}{n} \cdot \frac{\Var D_{X,\sigma}} {\Var D_{X,\sigma} + (\bbE D_{X, \sigma})^2 + 1}~, \\
\lmax(\bfM_{D, \sigma}) & \leq & \frac{n_\sigma}{n} \left(\Var D_{X,\sigma} + (\bbE D_{X, \sigma})^2 + 1\right)~,
\end{IEEEeqnarray*}
and these bounds are off by a factor of at most $2$. Therefore, as long as $n_\sigma/n = \Theta(1)$, $\bbE D_{X, \sigma} = \Theta(1)$ and $\Var D_{X, \sigma} = \Theta(1)$, the eigenvalues of $\bfM_{D, \sigma}$ are also $\Theta(1)$. Especially, \Cref{prop:sampling} shows that this holds with high probability if $D$ is sampled from a suitable distribution $\Pdata$. For example, it holds for the distribution used in \Cref{fig:nn_convergence_fused}.
\qedhere
\end{remark}

We can now state our main theorem, which is proven at the end of \Cref{sec:training_dynamics}.
In \Cref{thm:main:fixed_D} and \Cref{thm:main:inconsistency} we then apply this main theorem to 
over-parameterized and under-parameterized NNs, respectively. Note that in the formulation of \Cref{thm:main:general}, the constants $\CP, \Ch, \CW$ can depend on all previous constants and especially on $\varepsilon$, which prohibits removing the term $m^\varepsilon$ by taking the limit $\varepsilon \to 0$.

\begin{theorem} \label{thm:main:general}
Let $\varepsilon > 0$ and let $\gpsi, \gx, \gP \geq 0$ with $\gpsi + \gx + \gP < 1/2$. Then, for 
all given constants $\Cx, \Cpsi, \CM > 0$, there exist constants $\CP, \Ch, \CW > 0$ such that the following statement holds:

For all 
neural networks \eqref{def-NN-function} with 
width $m \in \bbN$ that are initialized and   trained according to 
\Cref{def:gd} with step size $h$ satisfying 
\begin{align}\label{step-size-assumption}
0< h \leq \Ch m^{-1}
\end{align}
and 
all data sets $D$ satisfying:
\begin{enumerate}[(a), wide=0pt, leftmargin=*]
\item $\udl{x}_D \geq \Cx m^{-\gx}$
\item $\CM^{-1} \leq \lmin(\bfM_{D, \pm 1}) \leq \lmax(\bfM_{D, \pm 1}) \leq \CM$
\item $\psi_{D, p} \leq \Cpsi$
\item $\psi_{D, q} \leq \Cpsi m^{\gpsi - 1}$
\end{enumerate}
the random training sequence $(f_{W_k})_{k \in \bbN_0}$ has the following properties with probability not less
than $1 - \CP m^{-\gP}$:
\begin{enumerate}[(i), wide=0pt, leftmargin=*]

\item For all $i=1,\dots,m$ and $k\geq 0$ it holds  $|a_{i, k} - a_{i, 0}| \leq \CW m^{\gpsi + \varepsilon - 3/2}$.
\item For all $i=1,\dots,m$ and $k\geq 0$ it holds $|b_{i, k} - b_{i, 0}| \leq \CW m^{\gpsi + \varepsilon - 3/2}$.
\item For all $i=1,\dots,m$ and $k\geq 0$ it holds $|w_{i, k} - w_{i, 0}| \leq \CW m^{\gpsi + \varepsilon - 1}$.
\item For all $k\geq 0$ it holds $|c_k - c_0| \leq \CW m^{\gpsi + \varepsilon - 1}$.
\item For all $k \geq 0$, the neural network function $f_{W_k}$ is affine linear on the intervals $(-\infty, -\Cx m^{-\gx}]$ and $[\Cx m^{-\gx}, \infty)$.
\end{enumerate}
\end{theorem}

To fully appreciate  \Cref{thm:main:general} a couple remarks are necessary:
Conditions \emph{(a)} and \emph{(b)} are conditions on the $x$-parts $D_{X,\sigma}$ 
of the data sets 
$D_{\pm 1}$, whereas \emph{(c)} and \emph{(d)} impose conditions on the 
optimal half-sided affine linear regression estimators. In a nutshell, 
\emph{(c)} only requires bounded slopes of these linear estimates, whereas 
\emph{(d)} demands the intercepts to become closer to $0$ for increasing network sizes. 
Moreover, note that the exponents $\gpsi, \gx, \gP$ 
describe a balance between data set characteristics \emph{(a)} and \emph{(d)} on the one-hand
and the accuracy of the guarantees on the other hand. For example, in the extreme case, in which 
we are only interested in data sets that are bounded away from the origin and whose 
intercepts vanish, we may choose $\gx = \gpsi = 0$ and $\gP = 1/2 - \varepsilon$
for all sufficiently small $\varepsilon>0$. Ignoring the nuisance term $\varepsilon$, the 
statements \emph{(i)} -- \emph{(v)} then  hold with probability $\geq 1-C_Pm^{-1/2}$,
the guarantees \emph{(i)} and \emph{(ii)} for the parameters of the hidden layer are of the 
order $O(m^{-3/2})$, and guarantees \emph{(iii)} and \emph{(iv)} for the parameters of the output neuron are of the 
order $O(m^{-1})$. In particular, the parameters ``typically'' move less during training the larger the network size is, and in fact, the probability of this event also increases with the network size.
Moreover, the guarantees for the hidden layer are stronger in $m$ suggesting that the 
parameters in the hidden layer move less during training than those of the output neuron.
The guarantee \emph{(v)} simply confirms the illustrative example in \Cref{fig:nn_convergence_fused}.

In a nutshell,
\Cref{thm:main:general} thus states that if $D$ is inside certain bounds and the step size $h$ is sufficiently small, the NN function $f_{W_k}$ remains affine on $D_1$ and $D_{-1}$ with high probability and the weights $W_k$ do not change much. This behavior is independent of when the iteration is stopped.

\begin{remark} \label{rem:step_size}
The data-independent condition \eqref{step-size-assumption} involves a conservative   constant $\Ch$. In \Cref{prop:ref_sum} and the proof of \Cref{thm:main:general}, we show that \eqref{step-size-assumption} can be replaced by $h \leq \lmax(\bfH)^{-1}$, where 
$\bfH \in \bbR^{4 \times 4}$ is a symmetric positive (semi-)definite matrix, which is  defined in \Cref{def:Aref} and which can be computed from $D$ and $W_0$.
\end{remark}

\section{Over-parameterization} \label{sec:over-parameterization}

In this case section we investigate the consequences of \Cref{thm:main:general}
for neural networks of arbitrary width $m$. We begin with the  
following corollary that describes these consequences for a \emph{fixed} data set $D$ of size $n\geq 4$, where $n \geq 4$ is necessary for $\Var D_{X, \pm 1} > 0$ as defined in \Cref{rem:MD-invert}. Note that in particular it applies to the over-parameterized case $m>n$.

\begin{corollary} \label{thm:main:fixed_D}
Let $D$ be fixed data set such that $\udl{x}_D > 0$,  $\Var D_{X,\pm 1} > 0$ %
and $\psi_{D, q} = 0$. 
Then for all $\varepsilon > 0$ and $0 < \gP < 1/2$, there 
exist constants $\CP, \Ch, \CW > 0$ such that  %
for all widths $m \geq 1$ and all step sizes
\begin{IEEEeqnarray*}{+rCl+x*}
0 < h & \leq & \Ch m^{-1}~,
\end{IEEEeqnarray*}
the random  sequence $(f_{W_k})_{k \in \bbN_0}$ obtained by initializing and training according to \Cref{def:gd}
satisfies (i) -- (v) in \Cref{thm:main:general} for $\Cx = \udl{x}_D$
and $\gx= \gpsi  = 0$ with probability not less
than $1 - \CP m^{-\gP}$.

\begin{proof}
By \Cref{rem:MD-invert} we known that there is a constant $\CM > 0$ such that \emph{(b)} of 
\Cref{thm:main:general} is satisfied. By choosing $\Cpsi := \psi_{D, p} + 1$, where 1 is added
to ensure $\Cpsi > 0$,
we then see that $D$ also satisfies the remaining assumptions of \Cref{thm:main:general}.
\end{proof}
\end{corollary}

\begin{remark}
In \Cref{thm:main:fixed_D}, it is also possible to choose $\gx > 0$, since then conclusion (v) of \Cref{thm:main:general} tells us that the kinks move less with increasing $m$. The price for this is that $\gP$, and therefore the probability of the kinks moving less, decreases accordingly.
\end{remark}

\begin{remark}\label{rem:attack}
\Cref{lemma:adding_points} shows that for one-dimensional data sets
$D$ with $x_j \neq 0$ for all $j$, the assumptions on 
the data set in \Cref{thm:main:fixed_D}
can always be satisfied by suitably  adding three more points to $D$.
By \Cref{thm:main:fixed_D},  these additional   points can play the role of \emph{adversarial} training samples
in the sense that they provably hinder the neural network training algorithm to converge to a good predictor if the 
best possible predictor is not half-sided affine linear, see again \Cref{fig:nn_convergence_fused}. 
Moreover, by \Cref{thm:main:fixed_D}  the success probability of such an attack increases with increasing 
network size $m$.
\qedhere
\end{remark}

Let us finally present a particularly simple data set to which \Cref{thm:main:fixed_D} applies and for which 
the best possible predictor is not half-sided affine linear.

\begin{example} \label{ex:D} 
The data set 
\begin{IEEEeqnarray*}{+rCl+x*}
D \equalDef ((-3, -1), (-2, 2), (-1, -1), (1, 1), (2, -2), (3, 1))
\end{IEEEeqnarray*}
satisfies $\udl{x}_D =1$ and $\Var D_{X,\pm 1} > 0$. Moreover, 
$\hat{\bfu}^0_{D, \pm 1} = \bfzero$ implies
$\psi_{D, p} = \psi_{D, q}  = 0$. 
In particular, the best possible empirical risk a half-sided affine linear
predictor can achieve is $1$.
Now assume that $m\geq 6$. Then one can easily construct a network  $f_W$ of the form \eqref{def-NN-function}
with $R_D(f_W) = 0$.
On the other hand, 
\Cref{thm:main:fixed_D}  shows that with high probability  $f_{W_k}$ is affine linear on 
$(-\infty, -1]$ and $[1, \infty)$ for all $k \geq 1$, and hence we have $R_D(f_{W_k})\geq 1$.
Consequently, training with gradient descent does not  come even close to a global optimum.
\end{example}

\section{Inconsistency} \label{sec:inconsistency}

In this section we present the inconsistency result for initializing and training a neural network according to 
\Cref{def:gd}.
To this end, we 
 describe a  class of  distributions that 
 produce data sets satisfying the assumptions of \Cref{thm:main:general}
with high probability, 
and for which the conditional mean function is not half-sided affine linear.

To describe the failure of gradient descent  in this case, we need to recall the  classical notion of consistency
from statistical learning theory, see 
e.g.~\citet[Chapter 6]{steinwart_support_2008}. Since 
the predictors obtained by our training scheme described in \Cref{def:gd} 
are probabilistic in both the data set $D$ and the 
initialization $W_0$, the following definition of consistency has been adapted in this respect.

\begin{definition}
\label{def:consistency}
A learning method, i.e., a method that produces for each data set $D$ a potentially random predictor $f_D$,
is called \emph{consistent} for a bounded distribution $\Pdata$,
if for all $\varepsilon > 0$, the probability of sampling a data set $D$ with $n$ i.i.d.\ samples $(x_j, y_j) \sim \Pdata$ satisfying
\begin{IEEEeqnarray*}{+rCl+x*}
R_{\Pdata}(f_D) \geq R_{\Pdata}^* + \varepsilon
\end{IEEEeqnarray*}
converges to $0$ as $n \to \infty$, where $R_{\Pdata}^*$ is the optimal risk, i.e., $R_{\Pdata}^* :=\inf_{f: \bbR \to \bbR} R_{\Pdata}(f)$.
The learning method is \emph{universally consistent} if this holds for all such $\Pdata$.
\end{definition}

Roughly speaking, 
universally consistent learning methods are guaranteed to produce close-to-optimal predictors for $n\to \infty$, \emph{independently} of the 
data generating distribution $\Pdata$. Universal consistency is therefore widely accepted as a minimal requirement
for   statistically sound learning methods, see e.g.~the books by \citet{devroye_probabilistic_1996}
and \citet{gyorfi_distribution-free_2002}.

Next we will show that neural networks initialized and trained as in \Cref{def:gd}
 are not universally consistent in a strong sense. Namely, we show that inconsistency occurs for 
 all distributions satisfying the following assumption.
For its formulation we denote the set of all $f:\bbR\to \bbR$ that are affine linear on both 
$(-\infty,0)$ and $(0,\infty)$ by $\calF_{\mathrm{hsal}}$.

\begin{assumption}[Data distribution II] \label{ass:P}
The  distribution $\Pdata$ satisfies \Cref{ass:pdata-1} and the following four conditions:
\begin{enumerate}[(P1), wide=0pt, leftmargin=*]

\item For all $\sigma \in \{\pm 1\}$, the matrix $\bfM_{\Pdata, \sigma}$ is invertible.
\item There exists an $\eta \in (4, \infty]$ such that we have
\begin{IEEEeqnarray*}{+rCl+x*}
\Pdata_X([-x, x]) & = & O(x^\eta) ~\text{ for $x \searrow 0$}, & $(\text{if }\eta < \infty)$ \\
\Pdata_X((-\delta, \delta)) & = & 0 ~\text{ for some $\delta > 0$}. & $(\text{if } \eta = \infty)$
\end{IEEEeqnarray*}
\item The intercepts of \Cref{def:main:quantities} satisfy $\psi_{\Pdata, q} = 0$. 
\item We have $\inf_{f \in \calF_{\mathrm{hsal}}} R_{\Pdata} (f) > 
R_{\Pdata}^*$.\qedhere
\end{enumerate}
\end{assumption}

The next remarks show, for example, that  $\Pdata$  
satisfies (P1) and (P2) if 
the distribution $\Pdata_X$ of the $x$ component has a density that is 
sufficiently small around $0$. They further show that we can always enforce (P3) by suitably modifying 
$\Pdata$
and that (P4)  means that the target function to be learned is not 
half-sided
affine linear in the sense of 
$\calF_{\mathrm{hsal}}$. As an example, we note that 
the data set   in \Cref{fig:nn_convergence_fused} has been sampled from a distribution 
satisfying \Cref{ass:P}.

\begin{remark}
Assumption (P1) is satisfied if $\Pdata_\sigma((\bbR \setminus \{x\}) \times \bbR) > 0$ for all $x \in \bbR$.
Indeed,  the kernel of the  matrix $\bfM_x$ is  $\Span \{(1, -x)^\top\}$, and hence there is, for any
vector $0 \neq \bfv \in \bbR^2$, at most one $x\in \bbR$ with $\bfv^\top \bfM_x \bfv=0$. This gives 
\begin{IEEEeqnarray*}{+rCl+x*}
\bfv^\top \bfM_{\Pdata, \sigma} \bfv = \int \bfv^\top \bfM_x \bfv \diff \Pdata_\sigma(x, y) > 0\, .
\end{IEEEeqnarray*}
In particular, (P1) is satisfied if, for example, $\Pdata_X$ has a density that does not completely
vanish on $(-\infty,0)$ or $(0,\infty)$.
\end{remark}

\begin{remark}
Assumption (P2) holds, for example, if $\Pdata_X$ has a density $p$ with 
\begin{IEEEeqnarray*}{+rCl+xx*}
p(x) & = & O(|x|^{\eta-1})~ \text{ for $x \to 0$}, & $(\text{if }\eta < \infty)$ \\
p(x) & = & 0 ~\text{ for $x \in (-\delta, \delta)$ for some $\delta > 0$.} & $(\text{if }\eta = \infty)$
\end{IEEEeqnarray*}
Verifying this claim is a straight-forward exercise.
\end{remark}

\begin{remark} \label{rem:half_shifting}
Let $\Pdata$ be a   distribution satisfying  \Cref{ass:pdata-1}, (P1), (P2), and (P4), and let
us fix a pair of random variables $(X, Y) \sim \Pdata$. Then the 
distribution $\tPdata$  of the vertically shifted random variables
\begin{align*}
 (X, Y - \qopt_{\Pdata, \sgn(X)})
\end{align*}
 satisfies \Cref{ass:P}.
This assertion immediately follows from the meaning of the intercepts $\popt_{\Pdata, -1}$ and $\popt_{\Pdata, 1}$ and the definition of the maximal absolute intercept
$\psi_{\Pdata, q}$.
\end{remark}

\begin{remark}
Recall from e.g.~\citet[Example 2.6]{steinwart_support_2008} that the risk $R_{\Pdata}$ is minimized by the conditional expectation $f^*(x) := \bbE_{\Pdata}(Y|x)$
and that the excess risk of a predictor $f:\bbR\to \bbR$ is 
\begin{align*}
   R_{\Pdata} (f) -  R_{\Pdata}^* = \frac 12\int |f(x) - f^*(x) |^2 \diff\Pdata_X(x)\, .
\end{align*}
Assumption
(P4) thus states that the least squares target function $f^*$ cannot be approximated by 
half-sided
affine linear functions in the sense of 
$\calF_{\mathrm{hsal}}$.
\end{remark}

\begin{figure}[tb]
\centering
\begin{tikzpicture}
\begin{axis}[axis lines = middle, width=\textwidth, height=4cm, xlabel=$x$, ylabel=$y$, xtick distance=0.5]
\addplot[domain=-1.75:-1.5, blue, very thick]{-1};
\addplot[domain=-1.25:-0.75, blue, very thick]{1};
\addplot[domain=-0.5:-0.25, blue, very thick]{-1};

\addplot[domain=1.75:1.5, blue, very thick]{1};
\addplot[domain=1.25:0.75, blue, very thick]{-1};
\addplot[domain=0.5:0.25, blue, very thick]{1};
\end{axis}
\end{tikzpicture}

\caption{Support of the distribution $\Pdata$ from \Cref{ex:class_P} for $\lambda = 1/4$. More specifically, if $(x, y) \sim \Pdata$, then $(x, y)$ is uniformly distributed on the blue intervals. Note that this distribution is symmetric for simplicity, but symmetry is not necessary for the conclusions of \Cref{ex:class_P}.} \label{fig:class_P_support}
\end{figure}

\Cref{ass:P} can be satisfied even by very simple distributions:

\begin{example} \label{ex:class_P}
Let $\lambda > 0$. Consider the distribution $\Pdata$ shown in \Cref{fig:class_P_support}, which is given by $\Pdata(y=1) = \Pdata(y=-1) = \frac{1}{2}$ and 
\begin{IEEEeqnarray*}{+rCl+x*}
\Pdata(x | y = \sigma) = \calU([\sigma \lambda, 2 \sigma \lambda] \cup [-3\sigma \lambda, -5\sigma \lambda] \cup [6\sigma \lambda, 7\sigma \lambda])~,
\end{IEEEeqnarray*}
for $\sigma \in \{\pm 1\}$. In other words, given $y=\sigma$, $y$ is uniformly distributed across four intervals. The intervals are also visualized in \Cref{fig:class_P_support}. By construction, we have $\Var(y) = 1$ and by a suitable choice of $\lambda$, we can even achieve that $\Var(x) = 1$. The distribution $\Pdata$ can be interpreted as a simple binary 
classification problem with classes $\pm 1$ without label noise where the classes are separated by a positive margin. 
For many classical learning algorithms, the classification error on such a problem behaves asymptotically like $O(n^{-1})$, or even like $O(0)$. For example, a simple calculation shows that 
histogram rules with fixed width smaller than $\lambda$ achieve zero classification error with high probability even for moderately sized data sets.

It is not hard to show that $\Pdata$ satisfies \Cref{ass:P} with $\eta = \infty$. For example, we can compute $\hat{\bfu}^0_{\Pdata, \pm 1} = \bfzero$, which implies $\psi_{\Pdata, p} = \psi_{\Pdata, q} = 0$. When employing the quadratic loss function used in this paper for this binary classification problem, \Cref{thm:main:inconsistency} will show that the training setup in \Cref{sec:notation} with high probability produces a NN with affine linear predictions on the left and right halves $\Pdata_{-1}$ and $\Pdata_{+1}$ of $\Pdata$. However, such affine linear predictions must have a classification error of at least $25\%$ since at least one of the three intervals on each side of the $y$ axis must be entirely misclassified. As mentioned in \Cref{def:D_sigma}, using a modification of the arguments in this paper, it is also possible to show the same result if only the part of $\Pdata$ with $x > 0$ or only the part with $x < 0$ is considered.
\end{example}

The next proposition, whose proof is delegated to \Cref{prop:sampling:appendix},
helps us to show that the assumptions of \Cref{thm:main:general} are
satisfied with high probability for data sets $D$ sampled from any distribution $\Pdata$ satisfying \Cref{ass:P}.
In its formulation we use the convention $\infty \cdot 0 \equalDef \infty$.

\begin{proposition} \label{prop:sampling}
Let $\Pdata$ satisfy  \Cref{ass:pdata-1} and (P1) -- (P3) from \Cref{ass:P} and let $\varepsilon>0, \Cx > 0$, $m \geq 1$, and $\gx, \gamma' \geq 0$.  
In the case $\eta = \infty$, we further assume that $\Cx$ is chosen such that it satisfies $\Pdata((-\Cx, \Cx) \times \bbR) = 0$.
Finally, let $D$ be a data set with $n\geq 1$ i.i.d.~samples $(x_j, y_j) \sim \Pdata$. Then with 
probability $1 - O(n^{-\gamma'} + nm^{-\eta \gx})$ the following statements simultaneously hold:
\begin{enumerate}[(D1), wide=0pt, leftmargin=*]

\item  $\underline{x}_D \geq \Cx m^{-\gx}$\, ,
\item For all $\sigma \in \{\pm 1\}$ we have 
\begin{align*}
 \frac{1}{2} \lmin(\bfM_{\Pdata, \sigma}) & \leq \lmin(\bfM_{D, \sigma}) \\
 \lmax(\bfM_{D, \sigma}) & \leq 2\lmax(\bfM_{\Pdata, \sigma})\, ,
\end{align*}
\item $\|\vopt_D - \vopt_{\Pdata}\|_\infty \leq n^{\varepsilon-1/2}$\, .

\end{enumerate}
\end{proposition}

By combining \Cref{prop:sampling} with \Cref{thm:main:general}, we obtain the following theorem, which 
is the key result for proving our inconsistency result. Due to the imposed constraint $m \leq \Cn n^{\frac{1}{2 - 2\gpsi}} < o(n)$, it is situated in the under-parameterized regime.

\begin{theorem}\label{thm:main:inconsistency}
Let $\Pdata$ satisfy \Cref{ass:P}, let $\varepsilon > 0$, and let $\gpsi, \gx, \gP \geq 0$ with $\gpsi + \gx + \gP < 1/2$. 
Then, for given constants  $\Cn, \Cx > 0$, where $\Cx$
needs to satisfy 
 $\Pdata((-\Cx, \Cx) \times \bbR) = 0$ in the case $\eta = \infty$,
there exist constants $\CP, \Ch, \CW > 0$ such that the following statement holds:

For all data sets
$D$  with $n\geq 1$ i.i.d.~samples $(x_j, y_j) \sim \Pdata$
and all
neural networks \eqref{def-NN-function} with 
width $m \geq 1$ that are initialized and   trained according to 
\Cref{def:gd} with step size $h$,
the random sequence $(f_{W_k})_{k \in \bbN_0}$ satisfies the conclusions (i) -- (v) of \Cref{thm:main:general}
with probability not less
than $1 - \CP(m^{-\gP} + nm^{-\eta \gx})$
provided  that $h$ and $m$ satisfy the constraints:
\begin{IEEEeqnarray*}{+rCl+x*} 
0 < h & \leq & \Ch m^{-1} \\
m & \leq & \Cn n^{\frac{1}{2 - 2\gpsi}}\, .
\end{IEEEeqnarray*}

\begin{proof}  
Let $\tilde{\varepsilon} > 0$ be small enough such that $\gpsi' \equalDef 2\tilde{\varepsilon} + \gpsi$ satisfies $\gpsi' + \gx + \gP < 1/2$. Under the assumptions above, there is a constant $\CM > 0$ such that for any data set $D$ satisfying (D1) -- (D3) 
of \Cref{prop:sampling}, we have the following three estimates:
\begin{IEEEeqnarray*}{+rCl+x*}
\udl{x}_D & \geq & \Cx m^{-\gx} \, ,\\
\frac{1}{\CM} & \leq & \lmin(\bfM_{D, \pm 1}) \leq \lmax(\bfM_{D, \pm 1}) \leq \CM\, , \\
\|\vopt_D - \vopt_{\Pdata}\|_\infty & \leq & n^{(\tilde{\varepsilon} - 1/2)} \leq O(m^{(2 - 2\gpsi)(\tilde{\varepsilon} - 1/2)}) \leq O(m^{2\tilde{\varepsilon} + \gpsi - 1}) = O(m^{\gpsi' - 1})~.
\end{IEEEeqnarray*}
It follows that
\begin{IEEEeqnarray*}{+rCl+x*}
\psi_{D, q} & \leq & \psi_{\Pdata, q} + \|\vopt_D - \vopt_{\Pdata}\|_\infty \stackrel{\text{(P3)}}{\leq} O(m^{\gpsi' - 1}) \\
\psi_{D, p} & \leq & \psi_{\Pdata, p} + \|\vopt_D - \vopt_{\Pdata}\|_\infty \leq \psi_{\Pdata, p} + O(m^{\gpsi' - 1}) \leq O(1)~.
\end{IEEEeqnarray*}
Then, by \Cref{thm:main:general}, there exists $\Ch > 0$ such that the conclusions
(i) -- (v) of \Cref{thm:main:general}
hold with probability $\geq 1 - O(m^{-\gP})$ if $0 < h \leq \Ch m^{-1}$. 
Let us now fix a $\gamma' \geq \gP$.
Because of (P1) and (P2), \Cref{prop:sampling} then shows that the conditions (D1) -- (D3) hold with probability 
\begin{align*}
 1 - O(n^{-\gamma'} + nm^{-\eta \gx}) \, .
\end{align*}
Moreover, $\gamma' \geq \gP$ implies 
$n^{-\gamma'} \leq O(m^{-(2 - 2\gpsi) \gamma'}) \leq O(m^{-\gamma'}) \leq O(m^{-\gP})$. By the union bound, the conclusions therefore hold with the specified probabilities.
\end{proof}
\end{theorem}

\begin{remark}\label{rem:mn-cond}
A simple calculation shows that the condition $\eta >4$ imposed in 
\Cref{ass:P}
is necessary and sufficient for the existence of 
 a  sequence $(m_n)_{n\geq 1}$ that satisfies the following two conditions:
 \begin{align*}
  nm_n^{-\eta \gx} \to 0
  \qquad \qquad \mbox{ and } \qquad \qquad 
  m_n \leq \Cn n^{\frac{1}{2 - 2\gpsi}}\, .
 \end{align*}
Note that the first condition ensures that the probability considered in 
\Cref{thm:main:inconsistency} converges to 1, while the second condition is a prerequisite 
in \Cref{thm:main:inconsistency}. Finally, note that such a sequence necessarily 
satisfies $m_n\to \infty$, and hence the approximation error of the considered networks
converge to zero by the universal approximation theorem. In other words, the 
considered neural networks can represent predictors $f_{m_n}:\bbR\to \bbR$ such that 
$R_{\Pdata}(f_{m_n}) \to R_{\Pdata}^*$ for $n\to \infty$, but gradient descent 
together with the considered initialization scheme is not able to find such predictors
as the following three corollaries, whose proofs are given at the end of \Cref{sec:appendix:inconsistency}, show.
\end{remark}

The first inconsistency result applies to all distributions 
$\Pdata$ satisfying \Cref{ass:P} for $\eta=\infty$, where we recall that 
$\eta=\infty$ simply means that $\Pdata_X$ has no mass in a small vicinity around 0. 

\begin{corollary} \label{cor:main:inconsistency}
Consider the learning method that, given a data set $D \in (\bbR\times \bbR)^n$, chooses a function $f_D = f_{W_k}$, where $k$ can be arbitrarily chosen and  
 $f_{W_k}$ is found according to \Cref{def:gd} for network size $m_n$ and step size $h_n$.
If $m_n$ and $h_n$ satisfy 
\begin{align*}
 m_n \leq O(n^{1-\varepsilon})\, ,
 \qquad \qquad 
 \lim_{n \to \infty} m_n = \infty\, ,
 \qquad \qquad \mbox{ and } \qquad \qquad
  h_n < o(m_n^{-1})\, ,
\end{align*}
then this 
learning method is not consistent for every distribution $\Pdata$ satisfying \Cref{ass:P} for $\eta=\infty$.
\end{corollary}

Note that the first condition $m_n \leq O(n^{1-\varepsilon})$ in \Cref{cor:main:inconsistency}
means that we are   in the under-parameterized regime of neural networks. 
However, since we may consider $\varepsilon \to 0$
we can get at least arbitrarily close to the limiting regime, in which the number of parameters 
grows linearly in $n$.

Our next inconsistency result applies to distributions satisfying \Cref{ass:P} for
some $4 < \eta <\infty$, i.e., for distributions that only have a small positive mass around 
$0$.

\begin{corollary}\label{cor:main:inconsistency-finite-eta}
Let $\eta \in (4, \infty)$ and 
consider the neural network  learning method of \Cref{cor:main:inconsistency}, but with 
$m_n \leq O(n^{1-\varepsilon})$ replaced by $m_n = \Theta(n^\gamma)$ for some $\gamma \in (\frac{2}{\eta} , 1 - \frac{2}{\eta})$.
Then this 
learning method is not consistent for every distribution $\Pdata$ satisfying \Cref{ass:P} for 
the chosen $\eta$.
\end{corollary}

Note that the price for extending the inconsistency 
from $\eta = \infty$ to $\eta <\infty$ is that we can no longer get arbitrarily close to the limiting regime discussed above. In fact, for $\eta \to 4$, the constraint on the network size
$m_n$ becomes stronger. This is consistent with the observations made in 
\Cref{rem:mn-cond}.

Our final result shows that neural networks acting on higher dimensional data are also not 
consistent provided that they are initialized and  trained in analogy to 
\Cref{def:gd}.

\begin{corollary} \label{cor:main:inconsistency:multi-d}
Consider a learning method as in \Cref{cor:main:inconsistency} but for $d$-dimensional data sets $D\in (\bbR^d\times \bbR)^n$ and with the initialization and training adaptations 
discussed in 
\Cref{rem:main:multi-d}.
Then this learning method is not universally consistent.
\end{corollary}

Let us finally set our results in a broader context of learning guarantees for other 
learning methods derived in statistical learning theory. 
Here, besides the worst-case approach leading to the notion of universal
consistency, refined results that, for example, establish learning rates for restricted classes 
of distributions exists. For such rates, one is typically interested in minmax
optimal rates, that is, in rates that, independent of the algorithm, cannot be improved
for the considered class of distributions.
In the regression context such restricted classes are almost always described by smoothness assumptions 
on the target function to be learned.
Typical and rather classical  
results in this direction show that many learning algorithms learn faster
the smoother this target function is. This corresponds to the intuition that 
learning should be easier if, on average, close-by
input points lead to close-by labels.
Now, 
the  classes of distributions
considered in this section
contain examples with arbitrarily high smoothness, meaning that at least some
traditional learning methods including gradient descent 
with early stopping in suitable reproducing kernel Hilbert spaces can learn these examples with up to the rate ${\cal O}(n^{-1})$, 
see  \cite{yao_early_2007}. In contrast, our results show that the considered neural networks  
do not learn at all in the case $d=1$.
In higher dimensions, the well-known curse of dimensionality
prevents fast learning rates for moderately smooth target functions unless additional assumptions are imposed.
One such assumption is the so-called manifold assumption, which, roughly speaking, assumes that the input data lies on a manifold with 
low intrinsic dimension $\rho<d$ and it is widely believed that at least some real-world data sets satisfy this assumption 
approximately. Theoretical results, see e.g.~\cite{hamm_adaptive_2021} and the references therein, show that 
in this case actually $\rho$ plays the role of the dimension in the resulting rates, or to phrase 
it differently, the smaller the intrinsic dimension is the faster some learning algorithms can learn. 

Both the smoothness and the manifold assumption were postulated to better describe common features of real-world
distributions, and our results show that the considered neural networks do not learn at all for some of these distributions, at least if $\rho = 1$. While the case $\rho = 1$ may seem unrealistic, our experiments in Section 10 indicate that postulating
lower bounds of the form $\rho\geq 2$ or even $\rho \geq 4$
may not solve the problem for (stochastic) gradient descent with zero bias initialization.

\section{Proof Idea} \label{sec:proof_idea}

Here, we want to give an overview over the proof of \Cref{thm:main:general}. We omit technical terms with exponent $\varepsilon$ for simplicity and choose a different order than in the appendix. %

As explained in \Cref{sec:overview}, we want to show that the kinks $-b_{i, k}/a_{i, k}$ do not move much  during training. 
To this end, we show that $|a_{i, 0}| \geq \Omega(m^{-1-\gP})$ with probability $1 - O(m^{-\gP})$. Hence, if $|a_{i, k} - a_{i, 0}| < o(m^{-1-\gP})$, then we still have $|a_{i, k}| \geq \Omega(m^{-1-\gP})$. Moreover, if $|b_{i, k}| = |b_{i, k} - b_{i, 0}| \leq O(m^{-1-\gP-\gx})$, then $|b_{i, k}/a_{i, k}| \leq O(m^{-\gx})$ and the main conclusion (v) of \Cref{thm:main:general} follows.

Note that the (Leaky)ReLU
$\varphi$ satisfies 
$\varphi(a_i x + b_i) = \varphi'(\sgn(a_i x + b_i)) \cdot (a_i x + b_i)$. 
We then investigate GD on a modified loss function $L_{D, \bftau}$, where we replace $\varphi'(\sgn(a_i x + b_i))$ by $\varphi'(\sgn(a_{i, 0} x + b_{i, 0}))$. If GD on this modified loss   does not move the kinks much, then $\sgn(a_i x + b_i)$ remains constant and $\nabla L_{D, \bftau}(W_k) = \nabla L_D(W_k)$. On both loss functions GD will thus  yield the same result. Such a strategy has also been used by \citet{li_learning_2018}.

Next, we will explain how to bound the change in $b_{i, k}$, the situation for $a_{i, k}$ is analogous. One can show that with $\sigma \equalDef \sgn(a_{i, 0})$, we have $b_{i, l+1} = b_{i, l} + hw_{i, l}s_{\sigma, l}$ for a quantity $s_{\sigma, l}$ defined in \Cref{def:derived_quantities}. We can then derive
\begin{IEEEeqnarray*}{+rCl+x*}
|b_{i, k} - b_{i, 0}| & \leq & h\sum_{l=0}^{k-1} |w_{i, l} s_{\sigma, l}| \leq \left(\sup_{0 \leq l < k} |w_{i, l}|\right) \cdot h\sum_{l=0}^{k-1} |s_{\sigma, l}| \\
& \leq & \left(|w_{i, 0}| + \sup_{0 \leq l < k} |w_{i, l} - w_{i, 0}|\right) \cdot h\sum_{l=0}^{k-1} |s_{\sigma, l}|~. \IEEEyesnumber \label{eq:example_diff_bound}
\end{IEEEeqnarray*}
Given a bound on $h\sum_{l=0}^{k-1} |s_{\sigma, l}|$, bounding $|b_{i, k} - b_{i, 0}|$ in this way requires bounding $|w_{i, l} - w_{i, 0}|$. Bounding the latter with a similar argument would require bounding $|b_{i, l'} - b_{i, 0}|$ and so on. While one can proceed by proving bounds using induction, we resolve the problem in a different but similar fashion in \Cref{prop:difference_bound} which does not require guessing an induction hypothesis.

As mentioned before, the neural network functions $f_{W_k}$ are piecewise affine. Hence there are affine functions $f_{W_k, 1}(x) = p_{1, k}x + q_{1, k}$ and $f_{W_k, -1}(x) = p_{-1, k}x + q_{-1, k}$ such that $f_{W_k}(x) = f_{W_k, 1}(x)$ for sufficiently large $x > 0$ and $f_{W_k}(x) = f_{W_k, -1}(x)$ for sufficiently small $x < 0$. A central quantity in our proof is the vector
\begin{IEEEeqnarray*}{+rCl+x*}
\ovl{\bfv}_k \equalDef \begin{pmatrix}
p_{1, k} - \popt_{D, 1} \\
p_{-1, k} - \popt_{D, -1} \\
q_{1, k} - \qopt_{D, 1} \\
q_{-1, k} - \qopt_{D, -1}
\end{pmatrix}
\end{IEEEeqnarray*}
containing the difference of the slope and intercept parameters to their affine regression optimum. We show in \Cref{sec:gradient_descent_equations} that $s_{\sigma, l}$ is a linear combination of components of $\ovl{\bfv}_l$. Thus, any bound on $h\sum_{l=0}^{k-1} \|\ovl{\bfv}_l\|$ directly yields a bound on $h\sum_{l=0}^{k-1} |s_{\sigma, l}|$. We also show that 
\begin{IEEEeqnarray*}{+rCl+x*}
\ovl{\bfv}_{k+1} = (\bfI_4 - h\bfA_k \bfM_D) \ovl{\bfv}_k~, \IEEEyesnumber \label{eq:system}
\end{IEEEeqnarray*}
where $\bfI_4$ denotes the four-dimensional identity matrix, 
$\bfA_k \in \bbR^{4 \times 4}$ depends on $W_k$, and $\bfM_D \in \bbR^{4 \times 4}$ is assembled using $\bfM_{D, 1}$ and $\bfM_{D, -1}$. Under the hypothesis that $W_k$ is close to $W_0$, we have $\bfA_k \approx \Aref$, where $\Aref \in \bbR^{4 \times 4}$ 
is a suitable matrix 
only depending on $W_0$.

We first consider a reference system $\ovl{\bfv}_{k+1} = (\bfI_4 - h\Aref \bfM_D) \ovl{\bfv}_k$. It can be shown that $\Aref$ and $\bfM_D$ are symmetric and positive definite (s.p.d.)\ with probability one. %
By a change of basis, we obtain the s.p.d.\ matrix $\bfH \equalDef \bfM_D^{1/2} \Aref \bfM_D^{1/2} = \bfM_D^{1/2} (\Aref \bfM_D) \bfM_D^{-1/2}$. Hence, $\Aref \bfM_D$ has positive real eigenvalues $\lambda_1, \hdots, \lambda_4$ with eigenvectors $\bfv_1, \hdots, \bfv_4$. If $\ovl{\bfv}_0 = \sum_i C_i \bfv_i$, then
\begin{IEEEeqnarray*}{+rCl+x*}
\ovl{\bfv}_k & = & \sum_{i=1}^4 (1 - h\lambda_i)^k C_i \bfv_i
\end{IEEEeqnarray*}
and for $0<h \leq (\max_i \lambda_i)^{-1}$, i.e., $1 - h\lambda_i \in [0, 1)$, we thus find
\begin{IEEEeqnarray*}{+rCl+x*}
h\sum_{l=0}^\infty \|\ovl{\bfv}_l\| & \leq & \sum_{i=1}^4 h\sum_{l=0}^\infty (1 - h\lambda_i)^l |C_i| \|\bfv_i\| = \sum_{i=1}^4 \lambda_i^{-1} |C_i| \|\bfv_i\|~.
\end{IEEEeqnarray*}
The idea is now to show that, with high probability, $\lambda_1, \lambda_2 = \Theta(m)$ and $\lambda_3, \lambda_4 = \Theta(1)$, while $\|\bfv_i\| \leq O(1)$, $|C_1|, |C_2| \leq O(1)$ and $|C_3|, |C_4| \leq O(m^{\gpsi-1})$ in order to obtain the bound
\begin{IEEEeqnarray*}{+rCl+x*}
h\sum_{l=0}^\infty \|\ovl{\bfv}_l\| & \leq & O(m^{\gpsi - 1})~.
\end{IEEEeqnarray*}
Indeed, we show in \Cref{prop:ref_sum} that $\Span \{\bfv_1, \bfv_2\} \approx \Span \{\bfe_1, \bfe_2\}$ with the first two standard unit vectors $\bfe_1, \bfe_2 \in \bbR^4$. With $q_{\sigma, 0} = 0$ and $\psi_{D, q} \leq O(m^{\gpsi-1})$, we also show that $\ovl{\bfv}_0$ is close to $\Span \{\bfe_1, \bfe_2\}$ and therefore $|C_3|, |C_4| \leq O(m^{\gpsi-1})$.

In \Cref{prop:sum-O}, we perform an induction showing that $W_k$ is close to $W_0$ and that the solution of $\ovl{\bfv}_{k+1} = (\bfI_4 - h\bfA_k \bfM_D) \ovl{\bfv}_k$ behaves similar to the solution of the reference system $\ovl{\bfv}_{k+1} = (\bfI_4 - h\Aref \bfM_D) \ovl{\bfv}_k$. Inserting the result into Eq.~\eqref{eq:example_diff_bound} yields the asymptotics
\begin{IEEEeqnarray*}{+rCl+x*}
|b_{i, k} - b_{i, 0}| & = & (O(m^{-1/2}) + o(m^{-1/2})) O(m^{\gpsi - 1}) \\
& = & O(m^{\gpsi - 3/2})~.
\end{IEEEeqnarray*} 
By our assumption $\gpsi + \gx + \gP < 1/2$, we have $m^{\gpsi - 3/2} < o(m^{-1-\gx-\gP})$ and as outlined in the beginning of this section, all kinks only move by $O(m^{-\gx})$.

The idea of deriving a system as in Eq.~\eqref{eq:system} and using induction to prove that $W$ does not change much over time has already been used by e.g.\  \citet{du_gradient_2019-1}. The main novelties in this part of our proof are:
\begin{itemize}

\item In our scenario, we are able to use a four-dimensional system instead of a $n$-dimensional system. We explain the relation between these systems in \Cref{sec:ntk}.
\item We find different eigenvalue asymptotics and results for the alignment of the corresponding eigenvectors depending on the data set, which requires more sophisticated arguments to exploit.
\item We prove different bounds on the change of weights in different layers, which is a consequence of using a more standard parameterization of the NN. We also prove strong bounds on certain \quot{second-moment} weight statistics, cf.\ \Cref{rem:second_moment_diff_bounds}.
\end{itemize}

\section{Relation to Neural Tangent Kernels} \label{sec:ntk}

In this section, we illustrate that a part of our approach essentially consists of factoring and analyzing the singular NTK matrix associated with our neural network. To this end, consider the continuous gradient flow dynamics
\begin{IEEEeqnarray*}{+rCl+x*}
\frac{\diff}{\diff t} W(t) & = & -\nabla L_D(W(t))~.
\end{IEEEeqnarray*}
As shown e.g.\ by \cite{du_gradient_2019-1}, the vector
\begin{IEEEeqnarray*}{+rCl+x*}
\ovl{\bff}(t) & \equalDef & \begin{pmatrix}
f_{W(t)}(x_1) - y_1 \\
\vdots \\
f_{W(t)}(x_n) - y_n \end{pmatrix}
\end{IEEEeqnarray*}
then satisfies the differential equation
\begin{IEEEeqnarray*}{+rCl+x*}
\dot{\ovl{\bff}}(t) & = & -\frac{1}{n} \bfK(t) \ovl{\bff}(t)~,
\end{IEEEeqnarray*}
where $\bfK(t) \in \bbR^{n \times n}$ is the empirical NTK matrix defined by
\begin{IEEEeqnarray*}{+rCl+x*}
[\bfK(t)]_{ij} & = & \left \langle \frac{\partial f_{W(t)}(x_i)}{\partial W}, \frac{\partial f_{W(t)}(x_j)}{\partial W}\right\rangle~.
\end{IEEEeqnarray*}
Assuming without loss of generality that $x_1, \hdots, x_{n'} > 0$ and $x_{n'+1}, \hdots, x_n < 0$ for suitable $n'$, define
\begin{IEEEeqnarray*}{+rCl+x*}
\bfX & \equalDef & \begin{pmatrix}
x_1 & 0 & 1 & 0 \\
\vdots & \vdots & \vdots & \vdots \\
x_{n'} & 0 & 1 & 0 \\
0 & x_{n'+1} & 0 & 1 \\
\vdots & \vdots & \vdots & \vdots \\
0 & x_n & 0 & 1
\end{pmatrix}~.
\end{IEEEeqnarray*}
We show in \Cref{prop:ntk_relations} that in our scenario with no kink reaching a data point (setting $h = 0$),
\begin{IEEEeqnarray*}{+rCl+x*}
\bfK(t) & = & \bfX \bfA(t) \bfX^\top \\
\bfM_D & = & \frac{1}{n}\bfX^\top \bfX \\
\ovl{\bfv} & = & (\bfX^\top \bfX)^{-1} \bfX^\top \ovl{\bff}~.
\end{IEEEeqnarray*}
Therefore $\frac{1}{n}\bfK(t)$ has the same non-zero eigenvalues and the same rank as $\frac{1}{n}\bfA(t) \bfX^\top \bfX = \bfA(t) \bfM_D$, i.e., it has rank four, two $\Theta(m)$ eigenvalues and two $\Theta(1)$ eigenvalues.\footnote{In general, if $v$ is an eigenvector of $AB$ with eigenvalue $\lambda \neq 0$, then $Bv \neq 0$ is an eigenvector of $BA$ with eigenvalue $\lambda$.}
It follows that
\begin{IEEEeqnarray*}{+rCl+x*}
\dot{\ovl{\bfv}} & = & (\bfX^\top \bfX)^{-1} \bfX^\top \dot{\ovl{\bff}} \\
& = & -\frac{1}{n} \bfA \bfX^\top \ovl{\bff} = -\frac{1}{n} \bfA (\bfX^\top \bfX) (\bfX^\top \bfX)^{-1} \bfX^\top \ovl{\bff} \\
& = & -\frac{1}{n} \bfA \bfX^\top \bfX \ovl{\bfv} = -\bfA \bfM_D\ovl{\bfv}~,
\end{IEEEeqnarray*}
which is the continuous-time analog of our update equation $\ovl{\bfv}_{k+1} = (\bfI_4 - h\bfA_k \bfM_D) \ovl{\bfv}_k$. In conclusion, after removing its null space, the rescaled kernel matrix $\frac{1}{n} \bfK$ is related to $\bfA \bfM_D$ via a change of basis. Working with $\bfA \bfM_D$ is more comfortable in our case, because $\bfA \bfM_D$ is invertible and it facilitates bounding the eigenvalues and eigenvectors.

\section{Inconsistency Experiments} \label{sec:experiments}

In this section, we present empirical evidence from some  Monte Carlo experiments that:
\begin{enumerate}[(1), wide=0pt, leftmargin=*]
\item The failure of kinks to move sufficiently far to reach data points can occur for realistic network sizes with high probability.
\item Stochastic gradient descent (SGD) can exhibit a similar behavior when combined with an early stopping rule.
\item A similar failure also occurs on multi-dimensional data sets that do not lie in a one-dimensional subspace as in \Cref{rem:main:multi-d}.
\end{enumerate}

\begin{figure}[tb]
\centering
\pgfplotscreateplotcyclelist{custom comparison}{%
solid, every mark/.append style={solid, fill=blue!50!white}, blue, mark=*\\%
dotted, every mark/.append style={solid, fill=blue!50!white}, blue, mark=square*\\%
densely dotted, every mark/.append style={solid, fill=orange!50!white}, orange, mark=triangle*\\%
dashdotted, every mark/.append style={solid, fill=orange!50!white}, orange, mark=star\\%
dashed, every mark/.append style={solid, fill=gray},mark=diamond*\\%
loosely dashed, every mark/.append style={solid, fill=gray},mark=*\\%
densely dashed, every mark/.append style={solid, fill=gray},mark=square*\\%
dashdotted, every mark/.append style={solid, fill=gray},mark=otimes*\\%
dashdotdotted, every mark/.append style={solid},mark=star\\%
densely dashdotted,every mark/.append style={solid, fill=gray},mark=diamond*\\%
}

\begin{tikzpicture}
\begin{loglogaxis}[xmin = , xmax = , ymin = 0.001, ymax = 1.0, xlabel = $m$ (Number of hidden neurons), ylabel = $P($not affine$)$, legend pos = south west, width = 14cm, height = 7cm, cycle list name = custom comparison, grid=major]
\addplot coordinates {
(16, 0.3611)
(32, 0.2501)
(64, 0.1824)
(128, 0.1303)
(256, 0.0919)
(512, 0.0659)
(1024, 0.0474)
(2048, 0.0335)
};
\addlegendentry{GD}
\addplot coordinates {
(16, 0.3621)
(32, 0.2493)
(64, 0.1866)
(128, 0.1287)
(256, 0.0749)
(512, 0.0427)
(1024, 0.0267)
(2048, 0.0134)};
\addlegendentry{GD, ES}
\addplot coordinates {
(16, 0.8612)
(32, 0.5653)
(64, 0.3468)
(128, 0.24450000000000005)
(256, 0.17179999999999995)
(512, 0.125)
(1024, 0.09819999999999995)
(2048, 0.07140000000000002)};
\addlegendentry{SGD, ES}
\addplot coordinates {
(16, 0.24280000000000002)
(32, 0.12970000000000004)
(64, 0.09050000000000002)
(128, 0.05279999999999996)
(256, 0.03410000000000002)
(512, 0.02429999999999999)
(1024, 0.01849999999999996)
(2048, 0.011499999999999955)};
\addlegendentry{SGD, ES, $h = \frac{0.01}{m}$}

\addplot[domain=16:2048] {1/sqrt(x)};
\addlegendentry{$y = x^{-1/2}$}
\end{loglogaxis}
\end{tikzpicture}
\caption{Monte-Carlo estimates ($10^4$ repetitions) of the probability of a kink 
crossing a sample
for different values of $m$, $n = m^2$, and different optimization and termination strategies. 
The data set 
is described in the text and 
 ES stands for early stopping. The line $y=x^{-1/2}$ is added to better illustrate the asymptotic behavior of the probabilities  in $m$. } \label{fig:comparison}
\end{figure}

Data for the figures in this section can be reproduced using the code at
\begin{IEEEeqnarray*}{+rCl+x*}
\text{\url{github.com/dholzmueller/nn_inconsistency}}~,
\end{IEEEeqnarray*}
which is archived at \url{https://doi.org/10.18419/darus-2978}.

\subsection{One-dimensional Distribution}

For (1) and (2), we use the following experimental setup: We compute each estimated probability using $10^4$ Monte Carlo trials. We choose $\Pdata$ as the uniform distribution on the data set $D$ from \Cref{ex:D} and sample a data set $D'$ of size $n = m^2$ from $\Pdata$. We initialize the weights independently as
\begin{IEEEeqnarray*}{+rCl+x*}
a_{i, 0} & \sim & \calN(0, 2), \quad b_{i, 0} = 0, \quad c_0 = 0, \quad w_{i, 0} \sim \calN(0, 2/m)~.
\end{IEEEeqnarray*}
We then use either gradient descent or stochastic gradient descent with batch size 16 in order to train the network and check whether a kink $-b_{i, k}/a_{i, k}$ leaves the interval $(-1, 1)$, in which case we can stop training.\footnote{Ignoring the unlikely case $a_{i, k} = 0$, $f_{W_k}$ remains affine on $(-\infty, -1]$ and $[1, \infty)$ iff no kink leaves $[-1, 1]$.} For some experiments, we also stop training if an early stopping (ES) criterion is 
satisfied.\footnote{We use early stopping as implemented in \texttt{Keras} \citep{chollet_keras_2015} with $\mathtt{patience} = 10$ and $\mathtt{min\_delta} = 10^{-8}$: Every 1000 epochs (GD) or 1000 batches (SGD), we monitor the loss on an independently drawn validation set of size $n$. Whenever $L_{\mathrm{val}} < L_{\mathrm{ref}} - 10^{-8}$, where $L_{\mathrm{val}}$ is the validation loss, we set $L_{\mathrm{ref}} \equalDef L_{\mathrm{val}}$. Training is stopped, when $L_{\mathrm{ref}}$ did not decrease within the last ten checks.
} For GD, we can also stop, if the techniques used in \Cref{prop:ref_sum}, \Cref{prop:difference_bound}, and \Cref{prop:sum-O} guarantee that no kink will ever leave 
the interval $(-1, 1)$.

For the step size $h$, unless specified otherwise, we use our maximal upper bound $h = \lmax(\bfH)^{-1}$ as mentioned in \Cref{rem:step_size} in order to reduce the number of iterations needed. We observe experimentally that $\lmax(\bfH)^{-1} \approx 0.4 m^{-1}$ for our choice of $\Pdata$.

\begin{figure}[tb]
\centering
\input{mc-plot-gd.tex}
\caption{Monte-Carlo estimates ($10^4$ trials) of the probability of a kink crossing a sample for different values of $m$. We use $n = m^2$ and GD without early stopping. Here, the data distribution $\Pdata$ of \Cref{fig:comparison}
is shifted upwards in $y$-direction  by $\Delta$. 
As described in the text, this change means that the condition 
$\psi_{D, q}\leq \Cpsi m^{\gpsi - 1}$ in \Cref{thm:main:general} is eventually 
violated for increasing $m$.
} \label{fig:shift}
\end{figure}

\Cref{fig:comparison} shows how the probabilities behave as $m$, and thus $n = m^2$, increases. In our scenario, we can apply \Cref{thm:main:inconsistency} with $\gpsi = \gx = 0$ and obtain that the probabilities should behave like $O(m^{-\gP})$ for all $\gP < 1/2$. In \Cref{fig:comparison}, this behavior can be observed even for small $m$ and also for SGD.\footnote{Without early stopping, SGD might still be able to move kinks far enough after a large number of iterations due to noisy gradients.} 
In this respect recall that it is 
shown in \Cref{fig:loss_and_kinks} that kinks may also move by significant amounts in later stages of the optimization. It is therefore not surprising that compared to considering $f_{W_k}$ for all $k \in \bbN_0$, using early stopping can significantly reduce the probability that a kink leaves $(-1, 1)$ during training. We see this especially for the small step size $h = 0.01m^{-1}$ in \Cref{fig:comparison}. Finally, if we shift $\Pdata$ upwards by adding some $\Delta \in \bbR$ to all $y$ values, we have $\psi_{\Pdata, q} = |\Delta|$ and assumption (P3) from \Cref{ass:P} is violated. We can see in \Cref{fig:shift} that this changes the asymptotic behavior, but for small $m$ and $|\Delta|$, the probabilities are still similarly low.
Note that this behavior is due to $\psi_{D, q} \leq O(m^{-1} + |\Delta|)$, where the $|\Delta|$ term dominates once we have entered the regime $m \gg 1/|\Delta|$, as it can be seen in \Cref{fig:shift}.

\subsection{Multi-dimensional Distribution} 
In order to show (3), i.e., that NNs can also perform poorly on multi-dimensional data sets, we choose the following experimental setup: We consider the uniform distribution $\Pdata$ on the data set from \Cref{fig:star_dataset}, which consists of the 33 points $(x_{ij}, y_{ij})$ defined by
\begin{IEEEeqnarray*}{+rCl+x*}
x_{ij} & = & (1 + 0.1 \cdot i) \begin{pmatrix}
\cos(2\pi j/11) \\
\sin(2\pi j/11)
\end{pmatrix}, \quad y_{ij} = 3|i| - 2, \quad j \in \{0, \hdots, 10\}, \quad i \in \{-1, 0, 1\}~.
\end{IEEEeqnarray*}
As above, we sample $n = m^2$ data points from $\Pdata$, but for $m \in \{128, 256, 512, \hdots, 8192\}$ with $1000$ Monte Carlo runs on $10^6$ epochs to ensure that the NNs have sufficiently converged. In the 1D case, we picked the step size $h = \lmax(\bfH)^{-1} = \lmax(\Aref \bfM_D) = n\lmax(\bfK_0)^{-1}$, where the last equality follows from \Cref{sec:ntk} for the kernel matrix $\bfK_0$ at initialization. Since we have not defined $\bfH$ for multi-dimensional data sets, we choose $h = n\lmax(\bfK_0)^{-1}$ in this case. As in the previous experiments this approach again results in different values for each considered NN instance. For the initialization, we again follow \cite{he_delving_2015} and use
\begin{IEEEeqnarray*}{+rCl+x*}
a_{i, j, 0} & \sim & \calN(0, 1), \quad b_{i, 0} = 0, \quad c_0 = 0, \quad w_{i, 0} \sim \calN(0, 2/m)~.
\end{IEEEeqnarray*}

\begin{figure}[tb]
\centering
\begin{tikzpicture}
\begin{axis}[grid, width=7cm, height=7cm, ymin=-1.3, ymax=1.3, xmin=-1.3, xmax=1.3, xlabel=$x_1$, ylabel=$x_2$]
\addplot[domain=0:360*10/11, samples=11, only marks, orange] ({0.9*cos(x)}, {0.9*sin(x)});
\addplot[domain=0:360*10/11, samples=11, only marks, blue] ({cos(x)}, {sin(x)});
\addplot[domain=0:360*10/11, samples=11, only marks, orange] ({1.1*cos(x)}, {1.1*sin(x)});
\end{axis}
\end{tikzpicture}
\caption{Illustration of $\Pdata$ of the  two-dimensional data sets used for the experiments in \Cref{fig:star_results}. The middle blue points have a $y$ value of $-2$ and the inner and outer orange points have a $y$ value of $1$.} \label{fig:star_dataset}
\end{figure}

\Cref{fig:star_results} shows that for this multi-dimensional distribution, the probability of a kink reaching a data point still decreases with increasing width of the network, although the probabilities are higher than in the one-dimensional setting above. Moreover, we also monitor the probability that the mean squared error (MSE) drops below $0.999999$ times the best MSE achievable by functions that are affine on each of the 11 \quot{arms} $\{x_{j, -1}, x_{j, 0}, x_{j, 1}\}$. We observe that even when a kink reaches a data point, the MSE often does not drop below this threshold.

\begin{figure}[tb]
\centering
\input{mc-plot-2d_star_11.tex}
\caption{Monte-Carlo estimates ($1000$ trials) of the probability of a kink crossing a sample / the MSE passing below $0.999999$ times the best possible \quot{affine} MSE for different values of $m$, $n = m^2$ using GD over $10^6$ epochs. Here, a uniform distribution on the data set from \Cref{fig:star_dataset} is used.} \label{fig:star_results}
\end{figure}

\section{Exploring Possible Improvements} \label{sec:practical} %

The inconsistency of the considered ReLU networks prompts the question on how they can be improved. In this section, we experimentally compare two-layer ReLU networks with He initialization and (stochastic) gradient descent to other two-layer ReLU Networks. When the input dimension $d$ is larger than one, we use SGD instead of GD in order to deal with larger data sets efficiently. Note that in our experiments in \Cref{sec:experiments}, SGD exhibits similar behavior to GD.

For possible improvements of the considered two-layer ReLU networks, a wide variety of options is available, such as
\begin{enumerate}[(1)]
\item changing the initialization / parameterization of the weights,
\item changing the optimizer,
\item using a different bias initialization method,
\item making the network deeper or wider, and
\item using a different activation function.
\end{enumerate}
For all of these options, it is currently unknown whether the resulting learning method will be universally consistent. For this reason, we investigate options (1) -- (3) using numerical experiments,\footnote{Our results can be reproduced using the code at \url{github.com/dholzmueller/nn_inconsistency}, which is archived at \url{https://doi.org/10.18419/darus-2978}.} keeping a two-layer ReLU NN structure: For input $\bfx \in \bbR^d$, let
\begin{IEEEeqnarray*}{+rCl+x*}
f_W(\bfx) & \equalDef & c + \alpha_2 \sum_{i=1}^m w_i \varphi(\alpha_1 \langle \bfa_i, \bfx \rangle + b_i) 
\end{IEEEeqnarray*}
with parameters $c, w_i, b_i \in \bbR$ and $\bfa_i \in \bbR^d$ and constants $\alpha_1, \alpha_2 \in \bbR$. In our experiments, we set $m=512$ and we always initialize the second-layer bias $c$ to zero since it is not followed by a ReLU activation $\varphi$. We consider the following options:

\subsection{Parameterization}
We consider two parameterizations with corresponding weight initializations:
\begin{itemize}
\item Standard parameterization (\textbf{He}): Following \cite{he_delving_2015}, we set $\alpha_1 = \alpha_2 = 1$ and initialize $\bfa_i \sim \calN(0, \frac{2}{d} \bfI_d)$ and $w_i \sim \calN(0, \frac{2}{m})$ independently. This parameterization is used in our theoretical analysis.
\item NTK parameterization (\textbf{NTK}): Similar to \cite{jacot_neural_2018}, we set $\alpha_1 = \sqrt{\frac{2}{d}}, \alpha_2 = \sqrt{\frac{2}{m}}$ and initialize $\bfa_i \sim \calN(0, \bfI_d)$ and $w_i \sim \calN(0, 1)$ independently. For gradient-based optimization, this parameterization usually increases the relative training speed of the first layer compared to the second layer.\footnote{
When going from standard parameterization to NTK parameterization, the magnitude of GD updates of $\alpha_1 \bfa_i$ is multiplied by $\alpha_1^2 = \frac{2}{d}$, while the magnitude of updates of $\alpha_2 w_i$ is multiplied by $\alpha_2^2 = \frac{2}{m}$. In the regime $m \gg d$ considered here, this means that the relative training speed of the first layer to the second layer is multiplied by $m/d \gg 1$. After a corresponding increase of the learning rate, the first layer will train faster than for the standard parameterization. Since we especially want the biases to train faster, we do not employ the option to multiply the biases by a (small) constant to reduce the training speed of the biases to a level similar to the standard parameterization \citep{jacot_neural_2018}.}
Compared to \cite{jacot_neural_2018}, we include the factor $\sqrt{2}$ in $\alpha_1$ and $\alpha_2$ such that the initial network function is distributed in the same way as for the standard parameterization. 
\end{itemize}

\subsection{Optimization} We consider the following optimizers, using hyperparameter optimization for the learning rate (see Training procedure below):
\begin{itemize}
\item Stochastic Gradient Descent (\textbf{SGD}): The non-stochastic version of this optimizer is used in our theoretical analysis and in the experiments for $d=1$.
\item SGD with Momentum (\textbf{SGDM}): We set the momentum hyperparameter to $0.9$, which corresponds to the default momentum used for Adam.
\item Adam (\textbf{Adam}): A very popular adaptive optimizer proposed by \cite{kingma_adam_2015}. Except for the learning rate, we use the standard hyperparameters $\beta_1 = 0.9$, $\beta_2 = 0.999$ and $\epsilon = 10^{-8}$.
\end{itemize}

\subsection{Bias Initialization} For the initialization of the first-layer biases, we consider multiple choices:
\begin{itemize}
\item Zero initialization (\textbf{Zero}): This is the initialization $b_i = 0$ that we have used throughout this paper.
\item PyTorch default (\textbf{PyTorch}): The default initialization in PyTorch initializes $b_i \sim \calU[-1/\sqrt{d}, 1/\sqrt{d}]$.
\item Uniform initialization (\textbf{$U(1)$}): We initialize $b_i \sim \calU[-\sqrt{6}, \sqrt{6}]$. For a normalized input distribution and for both parameterizations, we have $\bbE_{\bfx \sim P_X} \bbE_{\bfa_i} (\alpha_1 \langle \bfa_i, \bfx \rangle)^2 = 2 = \bbE b_i^2$.
\item Positive uniform initialization (\textbf{$U_+(1)$}): We initialize $b_i \sim \calU[0, \sqrt{6}]$.
\item Negative uniform initialization (\textbf{$U_-(1)$}): We initialize $b_i \sim \calU[-\sqrt{6}, 0]$.
\item Negative uniform kink initialization (\textbf{$U_{k-}(1)$}): We initialize $b_i = \alpha_1 \|\bfa_i\|_2 \beta_i$, where $\beta_i \sim \calU[-\sqrt{3}, 0]$ is independent of $\bfa_i$. This again ensures $\bbE b_i^2 = (\bbE (\alpha_1\|\bfa_i\|)^2)(\bbE \beta_i^2) = 2 \cdot 1 = 2$. Here, $|\beta_i|$ is the distance of the hyperplane $\{\bfx \in \bbR^d \mid \alpha_1 \langle \bfa_i, \bfx \rangle + b_i = 0\}$ to zero. This initialization method is inspired by the initialization methods considered by \cite{steinwart_sober_2019}.
\end{itemize}

\subsection{Training Procedure} We choose a training set with $256 \cdot d$ samples, use a batch size of $256$ and train for $8192 / d$ epochs such that we always use $8192$ iterations. Hence, the optimization only uses stochastic minibatches for $d \geq 2$. Every $64/d$ epochs, we compute errors on the validation and test sets, which contain $1024$ samples each. Test errors are reported for those epochs with the best validation errors. We run each hyperparameter configuration for $100$ times to obtain more accurate results. This is repeated for learning rates on a grid $\{\eta_0 \cdot 2^{k/2}, k \in \{-12, -11, \hdots, 10\}\}$, where $\eta_0$ is a sensible default learning rate depending on the parameterization and optimization. For the learning rate $\eta$ on the grid where the $100$ repetitions have the lowest RMSE on average, another grid search is performed on the finer grid $\{\eta \cdot 2^{k/8}, k \in \{-3, -2, \hdots, 3\}\}$ to find the finally used best learning rate. Using the same optimized learning rate for all $100$ repetitions resembles a practitioner using a well-tuned default value for the learning rate. %
\emph{For training, we normalize the training data such that $y$ and each component of $\bfx$ has zero mean and unit variance.}

\begin{figure}[tb]
\centering
\includegraphics[scale=1.0]{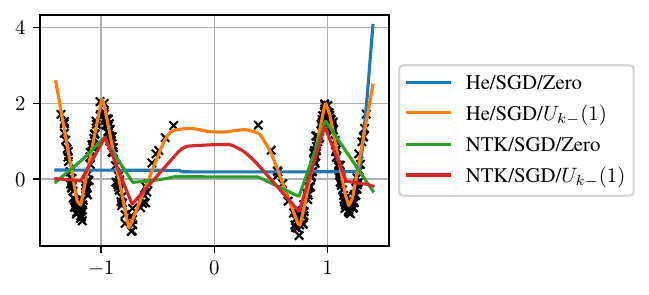}
\caption{Median-performing NNs for various configurations on a different data set that was drawn randomly from the distribution $\Pdata_{\mathrm{ex}}$ specified in \Cref{sec:appendix_experiments}. The training set is shown as black crosses and is identical to \Cref{fig:nn_convergence_fused} except that in \Cref{fig:nn_convergence_fused}, $\qopt_{D, \pm 1} = 0$ was ensured by shifting both halves of the data set as in \Cref{rem:half_shifting}. Training setup and learning rate optimization are identical to the other results in \Cref{sec:practical} for $d=1$. For each configuration, eleven NNs were trained using the best found learning rate without early stopping. After 8192 epochs, the NN function realizing the median of the eleven validation RMSEs was plotted. It is apparent that the combination He/SGD/Zero performs badly although not all assumptions of our theoretical results are satisfied: Here, the $x$ and $y$ values of the data set are normalized before training, which violates (P2) and (P3) from \cref{ass:P} as well as the assumption that the $(x_i, y_i)$ are i.i.d. Moreover, the learning rate is not constrained to a small regime during hyperparameter optimization.} \label{fig:nn_ex}
\end{figure}

\subsection{One-Dimensional Example} \Cref{fig:nn_ex} shows typical results of various training configurations on a one-dimensional data set similar to \Cref{fig:nn_convergence_fused}. In the remainder of this section, we consider a different set of data-generating distributions in order to study the influence of the input dimension on the performance of training configurations:

\begin{figure}[p]
\centering
\includegraphics[width=\textwidth]{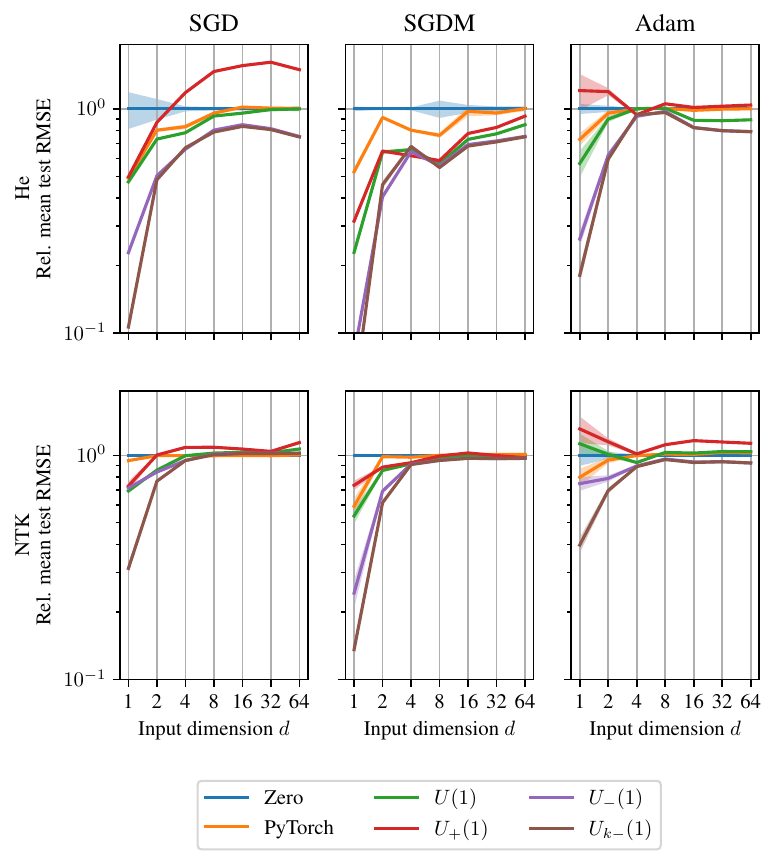}
\caption{Relative Mean Test RMSEs of trained NNs using the data-generating distribution from Eq.~\eqref{eq:rad_distr} for varying input dimension $d$, bias initialization method, optimizer and parameterization. The Mean Test RMSEs are plotted relative to the corresponding NNs with zero bias initialization on a logarithmic scale.} \label{fig:bias_inits}
\end{figure}

\subsection{Data Distribution} For $d \in \{1, 2, 4, 8, 16, 32, 64\}$, we randomly draw samples $(\bfx, y)$ as
\begin{IEEEeqnarray*}{+rCl+x*}
\bfx & = & u \frac{\tilde \bfx}{\|\tilde \bfx\|}, \qquad \tilde \bfx \sim \calN(0, \bfI_d), \quad u \sim \calU[0, 1] \\
y & = & \cos(2\pi \|\bfx\|) = \cos(2\pi u)~. \IEEEyesnumber \label{eq:rad_distr}
\end{IEEEeqnarray*}
Here, $\tilde \bfx$ and $u$ are independent random variables. This makes the distribution of $\bfx$ rotationally invariant, with its radius being uniformly distributed on $[0, 1]$. The definition of $y$ implies that no label noise is used. We chose $y$ such that it only depends on the radial component of the data and hence cannot be fit well with zero biases. We denote the distribution of $(\bfx, y)$ by $\Pdata_d$.

\begin{figure}[bt]
\centering
\includegraphics[width=\textwidth]{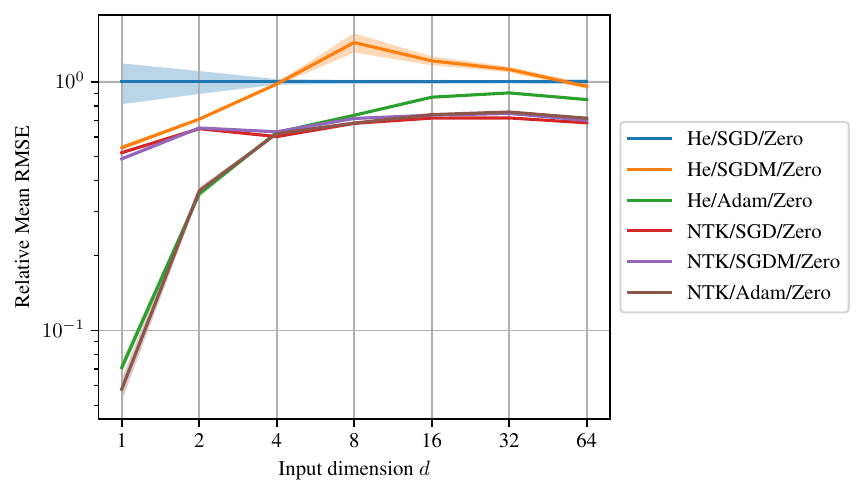}
\caption{Relative Mean Test RMSEs of trained NNs with zero-bias initialization using the data-generating distribution from Eq.~\eqref{eq:rad_distr} for varying input dimension $d$,  optimizer and parameterization. The Mean Test RMSEs are plotted relative to the combination of He parameterization/initialization and SGD.} \label{fig:param_opts}
\end{figure}

\subsection{Discussion} 
\begin{figure}[tb]
\centering
\includegraphics[scale=1.0]{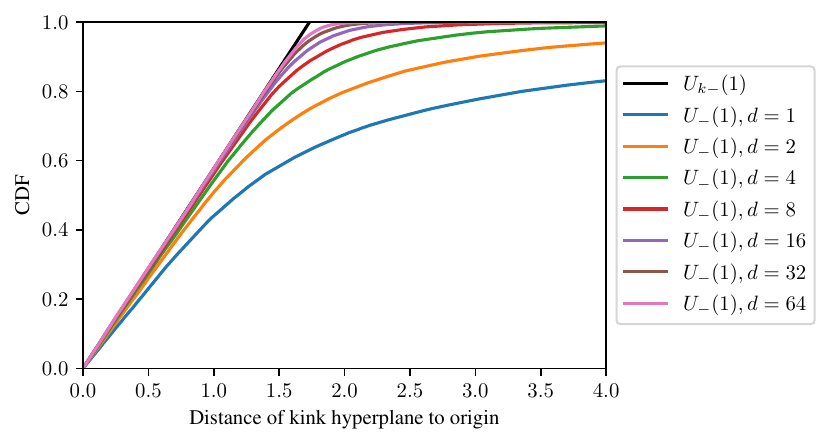}
\caption{CDFs of the distribution of kink distances to the origin given by $\frac{|b_i|}{|\alpha_1| \|\bfa_i\|}$ for different bias initializations and input dimensions $d$. The CDF is independent of $d$ when using $U_{k-}(1)$.} \label{fig:kink_distrs}
\end{figure}

For each combination of optimizer and parameterization, relative comparisons of the bias initialization methods are shown in \Cref{fig:bias_inits}. A direct comparison of optimizers and parameterizations for zero bias initialization is shown in \Cref{fig:param_opts}. Detailed results are provided in \Cref{sec:appendix_experiments}. We observe the following trends in the experimental results:
\begin{itemize}
\item \textbf{Bias initialization matters less for NTK parameterization.} This is likely because the first layer trains faster in NTK parameterization than in standard parameterization.
\item \textbf{NTK parameterization mostly outperforms standard parameterization.} A similar trend has been observed by \cite{arora_harnessing_2019}. Counterexamples to this trend occur mostly for good bias initializations and small dimension $d \in \{1, 2\}$. This can also be observed in \Cref{sec:appendix_experiments} and also in \Cref{fig:nn_ex}. We conjecture that in this case, the faster training speed of the first layer when using NTK parameterization is detrimental because it allows the kinks to move away from their initially well-distributed locations.
\item \textbf{Zero bias initialization performs badly for small input dimensions.} This is in agreement with our theoretical observations even though the data-generating distribution is more strongly concentrated around zero than (P2) in \Cref{ass:P} allows.
\item \textbf{Negative bias initialization helps.} In our experiments, negative uniform bias initialization $U_-(1)$ performs better than uniform bias initialization $U(1)$, which in turn performs better than positive uniform bias initialization $U_+(1)$. Note that the two neurons $\varphi(\langle \bfa, \bfx\rangle + b)$ and $\varphi(\langle -\bfa, \bfx\rangle + (-b))$ have the same \quot{kink hyperplane} $\{\bfx \in \bbR^d \mid \langle \bfa, \bfx\rangle + b = 0\}$ but are nonzero on opposite sides of the hyperplane. Since our distribution of the weights $\bfa$ is invariant under negation, the bias initializations $U_-(1)$, $U(1)$ and $U_+(1)$ generate the same distribution of kink hyperplanes. However, negative biases cause less data points to fall in the active region of the ReLU activation, creating a sparser activation pattern. This appears to be helpful, potentially because it eases optimization. A potential downside of negative bias initialization is that too small biases can create dead neurons, i.e., neurons that are always zero. Hence, the variance of the biases at initialization should not be too large.
\item \textbf{Initializing the kinks directly helps for small input dimensions.} For $d \in \{1, 2\}$, initializing the biases according to $U_{k-}(1)$ instead of $U_-(1)$, i.e., drawing the kinks uniformly instead of the biases, is mostly beneficial. For larger input dimensions, the differences between $U_{k-}(1)$ and $U_-(1)$ are very small. This is plausible: If $\xi \sim \calN(0, 1)$, then
\begin{IEEEeqnarray*}{+rCl+x*}
\Var(\alpha_1^2 \|\bfa_i\|_2^2) & = & \sum_{k=1}^d \Var(\alpha_1^2 a_{i,k}^2) = \sum_{k=1}^d \Var(\frac{2}{d} \xi^2) = \frac{4}{d} \Var(\xi^2) = \frac{8}{d}~.
\end{IEEEeqnarray*}
Therefore, the factor $\alpha_1 \|\bfa_i\|_2$ becomes almost constant for high input dimensions $d$ and the distributions $U_-(1)$ and $U_{k-}(1)$ become very similar. This is shown in \Cref{fig:kink_distrs}, which also shows that for small $d$ and bias initialization $U_-(1)$, the distribution of distances of kink hyperplanes from the origin has heavy tails and can therefore create many dead neurons, which might explain the superior performance of $U_{k-}(1)$.
\item \textbf{Adam is beneficial for small input dimensions.} In low dimensions, we observe that Adam performs better than SGDM and SGD. In higher dimensions, where the discrepancy between train and test errors becomes much larger, no clear picture emerges.
\end{itemize}

\subsection{Limitations} We have demonstrated that negative kink-based bias initializations can improve the test errors of NNs on a family of synthetic data sets. While encouraging, this result still leaves unanswered questions for future research. First of all, the practical use of such bias initialization methods can only be answered by a large-scale empirical study. \cite{steinwart_sober_2019} performed such a large-scale study on tabular data sets and found that data-dependent bias initialization methods can, on average, outperform zero initialization. As shown in \Cref{sec:appendix_experiments}, $U_{k-}(1)$ usually yields better results on the synthetic data sets considered here than one of the initialization methods proposed by \cite{steinwart_sober_2019}. Second, while our synthetic data distributions allow us to vary the input dimension in our experiments, distinguishing the effects of varying dimension from varying amounts of data and varying number of minibatches per epoch would require significantly more experiments. Third, a large-scale study including real-world data sets would also be necessary to determine the best kink distribution. In our experiments, we were not able to find a clearly best (data-dependent) kink distribution. For example, placing the kinks on random data points is not always better than $U_{k-}(1)$, cf. \Cref{sec:appendix_experiments}. We also do not study the effects of bias initialization methods in deeper layers of deeper NNs. The inputs of deeper layers can exhibit different characteristics than normalized input data in the first layer and might hence require different bias initialization methods. For example, a good data-dependent bias initialization might help to counteract sample variance decay \citep{luther_sample_2019} in deep NNs and the resulting dying ReLU problem.

\section{Conclusion} \label{sec:conclusion}

We have proven that NNs can fail with high probability in the over-parameterized regime for
certain data sets and in the under-parameterized regime even for sampled data sets. In these cases, the NN converges to a local \quot{linear regression} optimum. In particular, our analysis reveals that the difference of the NN to this linear regression optimum consists of a fast-decaying and a slow-decaying component and that for certain data sets, the slow-decaying component is already small at initialization. In essence, the reason is that learning is done mainly by the last layer, which contains $m$ weights but only one bias and therefore, the bias is learned more slowly than the weights. Especially, in our case, the NN operates in a \quot{lazy regime} \citep{chizat_lazy_2019}, where the NN stays close to a \quot{linearized version} around its initialization.

Using slightly different assumptions\footnote{For example, they use no biases $\bfb, c$ and they use the NTK parameterization. The NTK parameterization is essentially equivalent to using a smaller learning rate for the second layer, which might lead to a different optimization result than the usual parameterization.}, \citet{du_gradient_2019-1} show convergence of over-para\-me\-ter\-ized NNs to a global minimum. One of their assumptions is that no two data points $x_i, x_j$ should be parallel. This assumption is violated in our scenario, where \emph{all} $x_j$ are parallel.\footnote{While we use biases, which can be emulated in NNs without biases by transforming the inputs to $\tilde{x}_j \equalDef (x_j, 1)$, our biases are initialized differently from the weights, and hence such a transformation is not equivalent to the NNs considered here.}
Ironically, the case where samples lie on a low-dimensional submanifold of the input space is often considered a strength of deep learning methods, but at least in the 
extreme case of non-curved one-dimensional submanifolds our results show that neural networks with 
one hidden layer and zero bias initialization can be inconsistent. In this respect we 
also like to note that for a 
data set with parallel $x_j$, we can satisfy the assumption of \citet{du_gradient_2019-1}
by considering the transformed covariate samples
$\tilde{x}_j \equalDef (x_j, 1)$. In fact, this transformation essentially corresponds to 
considering the original data set but
initializing the hidden biases by $b_i \sim \calN(0, 1)$ instead of by $b_i = 0$. 
Note that this ensures that the kinks are now distributed randomly over $\bbR$ after initialization. Consequently, given enough over-parameterization, there are already enough kinks available such that an NN could achieve zero training loss without moving them at all.

The training of NNs can be viewed as a race between linear and nonlinear dynamics trying to fit the data. In both our setting and the setting of \citet{du_gradient_2019-1}, linear dynamics reach a critical point of the optimization landscape before nonlinear dynamics can contribute substantially. 
From a theoretical viewpoint, a major contribution of this paper is to show that this can happen with high probability even in a setting where the critical point is a spurious local minimum and nonlinear dynamics would be necessary to find a global optimum. While our result is a worst-case result in the sense that it only considers a very specific class of data-generating distributions, our experiments demonstrate that the considered training configuration performs badly beyond the assumptions of our theoretical analysis.

It should be noted that our result does not exclude the possibility of proving worst-case generalization guarantees, such as universal consistency, for NNs. On the contrary, it might assist the search for such guarantees by showing that a necessary assumption of such a guarantee would be to rule out our considered training configuration. While the training configuration of \citet{du_gradient_2019-1} allows to find a global optimum and therefore provides an interesting candidate setting for worst-case guarantees, it is yet unknown whether this leads to a universally consistent learning method. Moreover, \citet{du_gradient_2019-1} require the number of hidden neurons to grow polynomially with the number of training samples, which can quickly become infeasible even for moderatly sized data sets.
It is still an open question whether kinks move suitably in an intermediate case, e.g.\ if they are initialized randomly but the NN is not strongly over-parameterized. It also remains an open question for future research whether or not our results can be extended to deep NNs, other loss functions, or other optimization methods. 

While our analysis focuses on the role of biases during training, as we have already discussed above, biases can be interpreted as weights by considering inputs of the form $(x, 1)$. Hence, certain weight initialization methods such as sparse initialization methods might suffer from similar problems as zero bias initialization, as they can lead to similar degeneracies in the distribution of kink hyperplanes. Our insights might therefore help to explain why such weight initialization techniques are not frequently used. Moreover, it might be interesting to consider implications for NNs that are forced to be sparse, either by pruning after training or by imposing a sparse structure during training. %

From a practical point of view, we provide experimental results on the effect of different alterations to our training setup on synthetic data sets in different dimensions. These data sets are generated using cosine-type functions of the norm of the input, such that the learning of biases is crucial to achieve good results. For high-dimensional learning problems, where neural networks have been employed with great success, the radial behavior of the target function may not be very interesting. For example, in image classification, changing the norm of the input can be interpreted as changing the illumination of the image, which should not change the desired classification outcome. On the other hand, our experimental results are potentially relevant to the training of NNs on tabular data, a domain where other learning methods are still competitive with NNs \citep{shwartz-ziv_tabular_2022, kadra_regularization_2021, gorishniy_revisiting_2021}. Here, the input dimensions can be much smaller and the prediction result can strongly depend on the norm of the input. This can be intuitively seen for tasks such as the prediction of house prices from features such as floor area, or the task of predicting the runtime of a program from features such as the size of its input. Similarly, the learning of surrogate models for parameterized simulations may involve non-trivial radial behavior of the target function in low dimensions. Indeed, for tabular data, an extensive experimental evaluation by \cite{steinwart_sober_2019} has revealed that the results of zero bias initialization can be improved considerably by certain random bias initializations, which is in agreement with our experimental results on synthetic data sets. We hope that our analysis motivates further research to improve bias initializations and the distributions of ReLU kinks in general.

\acks{Funded by Deutsche Forschungsgemeinschaft (DFG, German Research Foundation) under Germany's Excellence Strategy - EXC 2075 - 390740016. The authors thank the International Max Planck Research School for Intelligent Systems (IMPRS-IS)
for supporting David Holzmüller.}

\newpage

\appendix

\numberwithin{theorem}{section}
\counterwithin{figure}{section}
\counterwithin{table}{section} %

\section{Notation and Matrix Algebra} \label{sec:notation}

In this section, we will introduce some notation that is used throughout the appendix. We will also list some results about matrices, especially involving matrix norms and eigenvalues, cf.\ e.g.\ \citet{bhatia_matrix_2013} as well as \citet{golub_matrix_1989}. The rest of the appendix is structured as follows:  In \Cref{sec:fixed_act_pattern}, we will define a modification of the NN with fixed activation pattern and prove some elementary results. We will then investigate gradient descent dynamics for derived quantities like $\ovl{\bfv}_k$ in \Cref{sec:gradient_descent_equations} and discuss them in \Cref{sec:comments}. In \Cref{sec:stochastics}, we prove stochastic properties of the initialization $W_0$ and the data set $D$. We then investigate a simplified \quot{linearized} case in \Cref{sec:reference_dynamics} and show in \Cref{sec:training_dynamics} that the true behavior is close to the linearized case, which concludes the proof of our main theorem. In \Cref{sec:multi-d}, we show how NNs with one-dimensional input and NNs with multi-dimensional input are related. Our inconsistency corollaries are proved in \Cref{sec:appendix:inconsistency}, a miscellaneous statement in \Cref{sec:misc} and relations of our quantities to NTK-related quantities in \Cref{sec:ntk_proofs}. Finally, detailed results accompanying the experiments in \Cref{sec:practical} are given in \Cref{sec:appendix_experiments}. %

\begin{definition}
We denote the sign of a real number $x \in \bbR$ by
\begin{IEEEeqnarray*}{+rCl+x*}
\sgn(x) & = & \begin{cases}
1 &\text{, if $x > 0$} \\
0 &\text{, if $x = 0$} \\
-1 &\text{, if $x < 0$}~.
\end{cases}
\end{IEEEeqnarray*}
For a set $S$, we denote its indicator function by $\bbone_S$, i.e.,
\begin{IEEEeqnarray*}{+rCl+x*}
\bbone_S(x) & = & \begin{cases}
1 &\text{, if $x \in S$} \\
0 &\text{, otherwise.}
\end{cases} & \qedhere
\end{IEEEeqnarray*}
\end{definition}

\begin{definition}[Asymptotic notation] \label{def:O-notation}
We use standard asymptotic notation $f < o(g)$, $f \leq O(g)$, $f = \Theta(g)$, $f \geq \Omega(g)$ (we do not need $f > \omega(g)$). The constant in such an asymptotic (in)equality should not depend on
\begin{itemize}

\item the number $m$ of hidden neurons,
\item the number $n$ of data points,
\item the step size $h > 0$,
\item step count variables such as $k, l, l' \in \bbN_0$,
\item the initialization $W_0$,
\item the data set $D$,
\end{itemize}
as long as these variables satisfy the imposed assumptions.\footnote{Sometimes, we also need to assume that $m, n$ is sufficiently large to be able to write $f \leq O(g)$ even if $f$ is infinite or undefined for small $m, n$.} For example, we could write $\Cx m^{-\gx} = \Theta(m^{-\gx})$ for $\Cx > 0$, but not $nm^{-\gx} \leq O(m^{-\gx})$.
\end{definition}

\begin{definition} \label{def:matrix_notation}
Let $n, m \geq 1$, let $\bfA, \bfB \in \bbR^{m \times m}$ and $\bfC \in \bbR^{n \times m}$. (We sometimes use $m, n$ to denote arbitrary vector space dimensions instead of numbers of hidden neurons and data points.)

\begin{enumerate}[(1), wide=0pt, leftmargin=*]

\item We write $\bfA \matgr 0$ iff $\bfA$ is symmetric and positive definite and $\bfA \matgeq 0$ iff $\bfA$ is symmetric and positive semidefinite. We define $\matleq$ and $\matless$ analogously.

\item A symmetric matrix $\bfA$ has an orthogonal eigendecomposition $\bfA = \bfU\bfD\bfU^\top$ with $\bfU \in \bbR^{m \times m}$ orthogonal and $\bfD \in \bbR^{m \times m}$ diagonal such that $\bfD$ contains the (real) eigenvalues of $\bfA$. We denote the set of eigenvalues of $\bfA$ by $\eig(\bfA)$ and define
\begin{IEEEeqnarray*}{+rCl+x*}
\lmax(\bfA) & \equalDef & \max \eig(\bfA) \\
\lmin(\bfA) & \equalDef & \min \eig(\bfA)~.
\end{IEEEeqnarray*}
The matrix $\bfA$ is invertible iff $0 \notin \eig(\bfA)$ and we have $\bfA \matgeq 0$ iff $\eig(\bfA) \subseteq [0, \infty)$. In the latter case, we can define the (symmetric) square root of $\bfA$ as $\bfA^{1/2} \equalDef \bfU\bfD^{1/2}\bfU^\top$, where $\bfD^{1/2}$ contains the square roots of the entries of $D$. Similarly, $\bfA^{-1} = \bfU \bfD^{-1} \bfU^\top$, which yields
\begin{IEEEeqnarray*}{+rClrCl+x*}
\lmax(\bfA^{1/2}) & = & \lmax(\bfA)^{1/2}, \quad & \lmin(\bfA^{1/2}) & = & \lmin(\bfA)^{1/2}, \quad \\
\lmax(\bfA^{-1}) & = & \lmin(\bfA)^{-1}, \quad & \lmin(\bfA^{-1}) & = & \lmax(\bfA)^{-1}~.
\end{IEEEeqnarray*}

\item As matrix norms, we use the Frobenius norm as well as the induced $2$- and $\infty$-norms:
\begin{IEEEeqnarray*}{+rCl+x*}
\|\bfC\|_F & = & \left(\sum_{i, j} C_{i, j}^2\right)^{1/2} \\
\|\bfC\|_2 & = & \sup_{x \neq 0} \frac{\|\bfC x\|_2}{\|x\|_2} \\
\|\bfC\|_\infty & = & \sup_{x \neq 0} \frac{\|\bfC x\|_\infty}{\|x\|_\infty} = \max_i \sum_j |C_{ij}|~.
\end{IEEEeqnarray*}
If $\bfC \matgeq 0$, then $\|\bfC\|_2 = \lmax(\bfC)$. 

These matrix norms satisfy the following inequalities (cf.\ e.g.\ Section 2.3 in \citet{golub_matrix_1989}):
\begin{IEEEeqnarray*}{+rCl+x*}
\|\bfC\|_2 & \leq & \|\bfC\|_F \leq \sqrt{m} \|\bfC\|_2 \\
\frac{1}{\sqrt{m}} \|\bfC\|_\infty & \leq & \|\bfC\|_2 \leq \sqrt{n} \|\bfC\|_\infty~.
\end{IEEEeqnarray*}
Moreover, if $\bfC'$ is a subblock of $\bfC$, then $\|\bfC'\|_p \leq \|\bfC\|_p$ for $p \in \{2, F, \infty\}$.

\item We define the condition number of a matrix $\bfA \matgr 0$ by
\begin{IEEEeqnarray*}{+rCl+x*}
\cond(\bfA) \equalDef \|\bfA\|_2 \cdot \|\bfA^{-1}\|_2 = \lmax(\bfA) \lmax(\bfA^{-1}) = \frac{\lmax(\bfA)}{\lmin(\bfA)}~.
\end{IEEEeqnarray*}

\item We occasionally use element-wise operations on matrices. For example, $|\bfA|$ is the matrix containing as entries the absolute values of the entries of $\bfA$ and $\sup_s \bfA(s)$ consists of the element-wise suprema. Also, $\bfA \leq \bfB$ means that $\bfA_{ij} \leq \bfB_{ij}$ for all $i, j$. \qedhere
\end{enumerate}
\end{definition}

There are some more facts about matrices that we will use during some proofs. We show some typical arguments here:
\begin{itemize}

\item We will use the fact that for symmetric $\bfA$,
\begin{IEEEeqnarray*}{+rCl+x*}
\lmax(\bfA) = \sup_{\|\bfv\|_2 = 1} \bfv^\top \bfA \bfv, \qquad \lmin(\bfA) = \inf_{\|\bfv\|_2 = 1} \bfv^\top \bfA \bfv~,
\end{IEEEeqnarray*}
which is a special case of the Courant-Fischer-Weyl min-max principle (e.g.\ Corollary III.1.2 in \citet{bhatia_matrix_2013}). This shows $\bfA \matgeq 0 \Leftrightarrow \lmin(\bfA) \geq 0$. For $\bfA, \bfB \matgeq 0$, we can use such an argument to show that
\begin{IEEEeqnarray*}{+rCl+x*}
\lmax(\bfA + \bfB) & \leq & \lmax(\bfA) + \lmax(\bfB) \\
\lmin(\bfA + \bfB) & \geq & \lmin(\bfA) + \lmin(\bfB)~.
\end{IEEEeqnarray*}

\item If
\begin{IEEEeqnarray*}{+rCl+x*}
\bfM = \begin{pmatrix}
\bfM_{11} & \bfM_{12} \\
\bfM_{12}^\top & \bfM_{22}
\end{pmatrix} \matgeq 0~,
\end{IEEEeqnarray*}
we know that
\begin{IEEEeqnarray*}{+rCl+x*}
\bfx^\top \bfM_{11} \bfx = \begin{pmatrix}
\bfx \\ 0
\end{pmatrix}^\top \bfM \begin{pmatrix}
\bfx \\ 0
\end{pmatrix} \geq \lmin(\bfM)\|\bfx\|_2^2 \geq 0~,
\end{IEEEeqnarray*}
hence $\bfM_{11} \matgeq 0$ with $\lmin(\bfM_{11}) \geq \lmin(\bfM)$. Similarly, $\lmax(\bfM_{11}) \leq \lmax(\bfM)$ and analogous identities hold for $\bfM_{22}$. We also have
\begin{IEEEeqnarray*}{+rCl+x*}
\eig \begin{pmatrix}
\bfM_1 \\
& \bfM_2
\end{pmatrix} = \eig(\bfM_1) \cup \eig(\bfM_2)
\end{IEEEeqnarray*}
and therefore
\begin{IEEEeqnarray*}{+rCl+x*}
\begin{pmatrix}
\bfM_1 \\
& \bfM_2
\end{pmatrix} \matgr 0 \text{ iff } \bfM_1, \bfM_2 \matgr 0~.
\end{IEEEeqnarray*}
\end{itemize}

\section{Gradient Descent with Fixed Activation Pattern} \label{sec:fixed_act_pattern}

In this section, we construct a modified loss function $L_{D, \bftau}$ which fixes the activation pattern of the neurons to its state at initialization. We also show that $L_{D, \bftau}(W) = L_D(W)$ for $W \approx W_0$ and introduce a shorter notation for gradient descent updates.

\begin{definition}[Fixed activation pattern]
Define
\begin{IEEEeqnarray*}{+rCl+x*}
\tau_i & \equalDef & \sgn(a_{i, 0}) \\
I & \equalDef & \{1, \hdots, m\} \\
J & \equalDef & \{1, \hdots, n\} \\
I_\sigma & \equalDef & \{i \in I \mid \tau_i = \sigma\} \\
J_\sigma & \equalDef & \{j \in J \mid \sgn(x_j) = \sigma\} \\
f_{W, \bftau, \sigma}(x) & \equalDef & c + \sum_{i \in I} w_i \varphi'(\sigma \tau_i) \cdot (a_i x + b_i) \\ %
L_{D, \bftau}(W) & \equalDef & \frac{1}{2n} \sum_{j \in J} (y_j - f_{W, \bftau, \sgn(x_j)}(x_j))^2~. & \qedhere
\end{IEEEeqnarray*}
\end{definition}

The previous definition is motivated by the following lemma:

\begin{lemma} \label{lemma:linearization_region}
For $\udl{x} > 0$ and $W_0 \in \bbR^{3m+1}$, consider the open set
\begin{IEEEeqnarray*}{+rCl+x*}
\calS_{W_0}(\udl{x}) \equalDef \{W \in \bbR^{3m+1} \mid \forall i \in I: |b_i| < (|a_{i, 0}| - |a_{i, 0} - a_i|) \udl{x}\}~.
\end{IEEEeqnarray*}
The functions $f_{W, \bftau, \sigma}$ are affine and for all $W \in \calS_{W_0}(\udl{x})$ and all $x \in \bbR$ with $|x| \geq \udl{x}$, we have
\begin{IEEEeqnarray*}{+rCl+x*}
f_W(x) & = & f_{W, \bftau, \sgn(x)}(x)~.
\end{IEEEeqnarray*}
If $\udl{x}_D > 0$, we have 
\begin{IEEEeqnarray*}{+rCl+x*}
L_D(W) & = & L_{D, \bftau}(W), \quad \nabla L_D(W) = \nabla L_{D, \bftau}(W)
\end{IEEEeqnarray*}
for all $W \in \calS_{W_0}(\udl{x}_D)$.

\begin{proof}
Trivially, $f_{W, \bftau, \sigma}$ is affine. Now, let $W \in \calS_{W_0}(\udl{x})$, let $i \in I$ and let $|x| \geq \udl{x}$. We then obtain that $|a_{i, 0}| - |a_{i, 0} - a_i| > 0$ and therefore
\begin{IEEEeqnarray*}{+rCl+x*}
|b_i| < (|a_{i, 0}| - |a_{i, 0} - a_i|)|x|~.
\end{IEEEeqnarray*}
Since $b_{i, 0} = 0$, we have
\begin{IEEEeqnarray*}{+rCl+x*}
|(a_i x + b_i) - (a_{i, 0} x + b_{i, 0})| & \leq & |a_i - a_{i, 0}| \cdot |x| + |b_i| < |a_{i, 0}| \cdot |x| = |a_{i, 0} x + b_{i, 0}|~.
\end{IEEEeqnarray*}
This shows $\sgn(a_i x + b_i) = \sgn(a_{i, 0}x + b_{i, 0})$, where
\begin{IEEEeqnarray*}{+rCl+x*}
\sgn(a_{i, 0} x + b_{i, 0}) & = & \sgn(a_{i, 0} x) = \sgn(a_{i, 0}) \sgn(x) = \tau_i \sgn(x)~.
\end{IEEEeqnarray*}
Due to our special choice of $\varphi$, we have
\begin{IEEEeqnarray*}{+rCl+x*}
\varphi(a_i x + b_i) & = & \varphi'(a_i x + b_i) \cdot (a_i x + b_i) = \varphi'(\sgn(a_i x + b_i)) \cdot (a_i x + b_i) \\
& = & \varphi'(\tau_i \sgn(x)) \cdot (a_i x + b_i)~,
\end{IEEEeqnarray*}
which yields
\begin{IEEEeqnarray*}{+rCl+x*}
f_W(x) & = & f_{W, \bftau, \sgn(x)}(x)~.
\end{IEEEeqnarray*}
Since all data points $x_j$ satisfy $|x_j| \geq \udl{x}_D$ by definition of $\udl{x}_D$, we find that
\begin{IEEEeqnarray*}{+rCl+x*}
L_{D, \bftau}(W) & = & \frac{1}{2n} \sum_{j \in J} (y_j - f_{W, \bftau, \sgn(x)}(x_j))^2 = \frac{1}{2n} \sum_{j \in J} (y_j - f_W(x_j))^2 = L_D(W)~.
\end{IEEEeqnarray*}
In addition, because $L_{D, \bftau}$ and $L_D$ are equal on the open set $\calS_{W_0}(\udl{x}_D)$, their derivatives must also be equal on $\calS_{W_0}(\udl{x}_D)$.
\end{proof}
\end{lemma}

We can now define gradient descent iterates with respect to the \quot{linearized} loss function $L_{D, \bftau}$.

\begin{definition}[Gradient descent]
Given the random initial vector $W_0$, its activation pattern $\bftau$ and a step size $h > 0$, we recursively define
\begin{IEEEeqnarray*}{+rCl+x*}
W_{k+1} & \equalDef & W_k - h\nabla L_{D, \bftau}(W_k)~.
\end{IEEEeqnarray*}
Moreover, we write $W_k = (\bfa_{\cdot, k}, \bfb_{\cdot, k}, c_k, \bfw_{\cdot, k})$ and we may implicitly omit the index $k$ when deriving identities that hold for each $k \in \bbN_0$. For any derived quantity $\xi \equalDef g(W)$, define
\begin{IEEEeqnarray*}{+rCl+x*}
\delta \xi & \equalDef & \delta g(W) \equalDef g(W - h\nabla L_{D, \bftau}(W)) - g(W)
\end{IEEEeqnarray*}
such that
\begin{IEEEeqnarray*}{+rCl+x*}
\xi_{k+1} = g(W_{k+1}) = g(W_k) + (g(W_{k+1}) - g(W_k)) = \xi_k + \delta \xi_k
\end{IEEEeqnarray*}
and hence
\begin{IEEEeqnarray*}{+rCl+x*}
\delta g(W) & = & g(W + \delta W) - g(W)~. & \qedhere
\end{IEEEeqnarray*}
\end{definition}

We can now write iteration rules differently: Instead of
\begin{IEEEeqnarray*}{+rCl+x*}
W_{k+1} = W_k - h\nabla L_{D, \bftau}(W_k)~,
\end{IEEEeqnarray*}
we will use the more convenient notation
\begin{IEEEeqnarray*}{+rCl+x*}
\delta W = -h\nabla L_{D, \bftau}(W)
\end{IEEEeqnarray*}
which suppresses the iteration index $k$ and reads more like the negative gradient flow ODE
\begin{IEEEeqnarray*}{+rCl+x*}
\dot{W} = -h\nabla L_{D, \bftau}(W)~.
\end{IEEEeqnarray*}

The following lemma introduces some convenient rules for using $\delta$.

\begin{lemma}[Differential calculus for $\delta$] \label{lemma:discrete_diffcalc}
Let $g: \bbR^{3m+1} \to \bbR^N$ for some $m, N \geq 1$.
\begin{enumerate}[(a), wide=0pt, leftmargin=*]

\item If $g$ is linear, then $\delta g(W) = g(\delta W) = -h g(\nabla L_{D, \bftau}(W))$. 
\item If $g$ is constant, then $\delta g = 0$.
\item If $g_1, g_2: \bbR^{3m+1} \to \bbR$ are linear, then
\begin{IEEEeqnarray*}{+rCl+x*}
\delta (g_1 \cdot g_2) = (\delta g_1) \cdot  g_2 + g_1 \cdot (\delta g_2) + (\delta g_1) \cdot (\delta g_2)~.
\end{IEEEeqnarray*}
\item If $g_1, g_2: \bbR^{3m+1} \to \bbR^N$, then
\begin{IEEEeqnarray*}{+rCl+x*}
\delta (g_1 + g_2) = \delta g_1 + \delta g_2~.
\end{IEEEeqnarray*}
\item If $g_2: \bbR^{3m+1} \to \bbR^N, g_1: \bbR^N \to \bbR^{N'}$ and $g_1$ is linear, then 
\begin{IEEEeqnarray*}{+rCl+x*}
\delta (g_1 \circ g_2) = g_1 \circ (\delta g_2)~.
\end{IEEEeqnarray*}
\item If $g_1, \hdots, g_N: \bbR^{3m+1} \to \bbR$, then
\begin{IEEEeqnarray*}{+rCl+x*}
\delta \begin{pmatrix}
g_1 \\
\vdots \\
g_N
\end{pmatrix} = \begin{pmatrix}
\delta g_1 \\ 
\vdots \\
\delta g_N
\end{pmatrix}~.
\end{IEEEeqnarray*}
\end{enumerate}

\begin{proof}
\leavevmode
\begin{enumerate}[(a), wide=0pt, leftmargin=*]

\item If $g$ is linear, then
\begin{IEEEeqnarray*}{+rCl+x*}
\delta g(W) & = & g(W + \delta W) - g(W) = g(\delta W) = g(-h\nabla L_{D, \bftau}(W)) = -h g(\nabla L_{D, \bftau}(W))~.
\end{IEEEeqnarray*}
\item Trivial.
\item In this case,
\begin{IEEEeqnarray*}{+rCl+x*}
\delta g(W) & = & g(W + \delta W) - g(W) \\
& = & g_1(W)g_2(\delta W) + g_1(\delta W) g_2(W) \\
&& ~+~ g_1(\delta W) g_2(\delta W) \\
& \stackrel{\text{(a)}}{=} & \delta g_1(W) g_2(W) + g_1(W) \delta g_2(W) + \delta g_1(W) \delta g_2(W)~.
\end{IEEEeqnarray*}
\item We have
\begin{IEEEeqnarray*}{+rCl+x*}
\delta (g_1 + g_2)(W) & = & (g_1 + g_2)(W + \delta W) - (g_1 + g_2)(W) \\
& = & (g_1(W + \delta W) - g_1(W)) + (g_2(W + \delta W) - g_2(W)) \\
& = & \delta g_1(W) + \delta g_2(W)~.
\end{IEEEeqnarray*}

\item For $W \in \bbR^{3m+1}$,
\begin{IEEEeqnarray*}{+rCl+x*}
\delta(g_1 \circ g_2)(W) & = & g_1(g_2(W + \delta W)) - g_1(g_2(W)) = g_1(g_2(W+\delta W) - g_2(W)) \\
& = & g_1(\delta g_2(W))~.
\end{IEEEeqnarray*}

\item This follows from
\begin{IEEEeqnarray*}{+rCl+x*}
\begin{pmatrix}
g_1 \\
\vdots \\
g_N
\end{pmatrix}(W + \delta W) - \begin{pmatrix}
g_1 \\
\vdots \\
g_N
\end{pmatrix}(W) & = & \begin{pmatrix}
g_1(W+\delta W) - g_1(W) \\
\vdots \\
g_N(W+\delta W) - g_N(W)
\end{pmatrix}~. & \qedhere
\end{IEEEeqnarray*}
\end{enumerate}
\end{proof}
\end{lemma}

\section{Reformulation of Gradient Descent} \label{sec:gradient_descent_equations}

In this section, we will derive equations that describe how different aspects of the neural network behave during gradient descent. A summary and interpretation of the derived equations is presented in \Cref{sec:comments}.

\begin{definition}[Derived quantities] \label{def:derived_quantities}
\leavevmode
\begin{enumerate}[(a), wide=0pt, leftmargin=*]

\item For $\sigma \in \{\pm 1\}$, we write $\Sigma_{\sigma, a^2} \equalDef \sum_{i \in I_\sigma} a_i^2$, $\Sigma_{\sigma, wa} \equalDef \sum_{i \in I_\sigma} w_i a_i$ and so on. 

\item The matrix $\bfM_\sigma \equalDef \bfM_{D, \sigma}$ from \Cref{def:main:quantities} helps in relating different interesting quantities. For $\bfM_\sigma \matgr 0$, let
\begin{IEEEeqnarray*}{+rClrCl+x*}
\hat{\bfv}_\sigma & \equalDef & \begin{pmatrix}
\hat{p}_\sigma \\ \hat{q}_\sigma
\end{pmatrix} \equalDef \begin{pmatrix}
\Sigma_{\sigma, wa} \\ \Sigma_{\sigma, wb}
\end{pmatrix} & \bfv_\sigma & \equalDef & \begin{pmatrix}
p_\sigma \\ q_\sigma
\end{pmatrix} \equalDef \begin{pmatrix}
\hat{p}_\sigma + \alpha \hat{p}_{-\sigma} \\
c + \hat{q}_\sigma + \alpha \hat{q}_{-\sigma}
\end{pmatrix} \\
\hat{\bfu}_\sigma & \equalDef & \begin{pmatrix}
\hat{r}_\sigma \\ \hat{s}_\sigma
\end{pmatrix} \equalDef \begin{pmatrix}
-\frac{1}{n} \sum_{j \in J_\sigma} (f_{W, \bftau, \sigma}(x_j) - y_j) x_j \\
-\frac{1}{n} \sum_{j \in J_\sigma} (f_{W, \bftau, \sigma}(x_j) - y_j)
\end{pmatrix} \quad & \bfu_\sigma & \equalDef & \begin{pmatrix}
r_\sigma \\ s_\sigma
\end{pmatrix} \equalDef \begin{pmatrix}
\hat{r}_\sigma + \alpha \hat{r}_{-\sigma} \\
\hat{s}_\sigma + \alpha \hat{s}_{-\sigma}
\end{pmatrix}
\end{IEEEeqnarray*}
and
\begin{IEEEeqnarray*}{+rCl+x*}
\ovl{\bfv}_\sigma \equalDef \bfv_\sigma - \vopt_\sigma~.
\end{IEEEeqnarray*}
We will show in \Cref{lemma:deriv:linear_relations} that $\hat{\bfu}_\sigma = -\bfM_\sigma \ovl{\bfv}_\sigma$. The $\bfu$-vectors are interesting since their components can be used to simplify $\delta W$. As we will see in \Cref{lemma:deriv:linear_relations}, $\bfv_\sigma$ is interesting since $f_{W, \bftau, \sigma}(x) = p_\sigma x + q_\sigma$ for $x \in \bbR$. The notation of the different variants is motivated as follows: Expressions with a hat such as $\hat{\bfv}_\sigma$ and $\hat{\bfu}_\sigma$ only sum over one sign $\sigma$ while hat-less expressions include both $\sigma = 1$ and $\sigma = -1$. %

We will also use the matrices
\begin{IEEEeqnarray*}{+rClrClrCl+x*}
\Gw_\sigma & \equalDef & \begin{pmatrix}
\Sigma_{\sigma, w^2} & 0 \\
0 & \Sigma_{\sigma, w^2}
\end{pmatrix}, \quad & \Gab_\sigma & \equalDef & \begin{pmatrix}
\Sigma_{\sigma, a^2} & \Sigma_{\sigma, ab} \\
\Sigma_{\sigma, ab} & \Sigma_{\sigma, b^2}
\end{pmatrix}, \quad & \Gwab_\sigma & \equalDef & (r_\sigma \Sigma_{\sigma, wa} + s_\sigma\Sigma_{\sigma, wb}) \bfI_2~,
\end{IEEEeqnarray*}
where $\bfI_2$ is the $2 \times 2$ identity matrix.

\item For any two vectors $\bfz_1, \bfz_{-1} \in \bbR^2$ defined in step (c) and any two matrices $\bfF_1, \bfF_{-1} \in \bbR^{2 \times 2}$ defined in step (b), we define
\begin{IEEEeqnarray*}{+rCl+x*}
\tilde{\bfz} \equalDef \begin{pmatrix}
\bfz_1 \\ \bfz_{-1}
\end{pmatrix} \in \bbR^4, \quad \tilde{\bfF} \equalDef \begin{pmatrix}
\bfF_1 \\
& \bfF_{-1}
\end{pmatrix} \in \bbR^{4 \times 4}~.
\end{IEEEeqnarray*}
For example, this means that
\begin{IEEEeqnarray*}{+rCl+x*}
\tilde{\bfu} & = & \begin{pmatrix}
\bfu_1 \\ \bfu_{-1}
\end{pmatrix} = \begin{pmatrix}
r_1 \\ s_1 \\ r_{-1} \\ s_{-1}
\end{pmatrix}~.
\end{IEEEeqnarray*}
In addition, we define new matrices
\begin{IEEEeqnarray*}{+rCl+x*}
\tilde{\bfC} & \equalDef & \begin{pmatrix}
0 & 0 & 0 & 0 \\
0 & 1 & 0 & 1 \\
0 & 0 & 0 & 0 \\
0 & 1 & 0 & 1
\end{pmatrix}, \quad \tilde{\bfB} \equalDef \begin{pmatrix}
1 & 0 & \alpha & 0 \\
0 & 1 & 0 & \alpha \\
\alpha & 0 & 1 & 0 \\
0 & \alpha & 0 & 1
\end{pmatrix} = \begin{pmatrix}
\bfI_2 & \alpha \bfI_2 \\
\alpha \bfI_2 & \bfI_2
\end{pmatrix}, \\
\tilde{\bfA} & \equalDef & \tilde{\bfB}(\tGw + \tGab + h\tGwab)\tilde{\bfB} + \tilde{\bfC}~.
\end{IEEEeqnarray*}
We will prove in \Cref{prop:deriv:AM} that $\delta \tilde{\ovl{\bfv}} = h\tilde{\bfA}\tilde{\hat{\bfu}} = -h\tilde{\bfA}\tilde{\bfM}\tilde{\ovl{\bfv}}$.

\item We want to perform a change of basis using the permutation matrix 
\begin{IEEEeqnarray*}{+rCl+x*}
\tilde{\bfP} & \equalDef & \begin{pmatrix}
1 & 0 & 0 & 0 \\
0 & 0 & 1 & 0 \\
0 & 1 & 0 & 0 \\
0 & 0 & 0 & 1
\end{pmatrix}~.
\end{IEEEeqnarray*}
which satisfies $\tilde{\bfP} = \tilde{\bfP}^\top = \tilde{\bfP}^{-1}$: For any vector $\tilde{\bfz} \in \bbR^4$ and any matrix $\tilde{\bfF} \in \bbR^{4 \times 4}$ defined in step (d), we define
\begin{IEEEeqnarray*}{+rCl+x*}
\bfz \equalDef \tilde{\bfP} \tilde{\bfz}, \quad \bfF \equalDef \tilde{\bfP} \tilde{\bfF} \tilde{\bfP}^{-1} = \tilde{\bfP} \tilde{\bfF} \tilde{\bfP}~.
\end{IEEEeqnarray*}
For example, this yields
\begin{IEEEeqnarray*}{+rCl+x*}
\bfu = \tilde{\bfP} \tilde{\bfu} = \begin{pmatrix}
r_1 \\ r_{-1} \\ s_1 \\ s_{-1}
\end{pmatrix}, \quad \bfC = \begin{pmatrix}
0 & 0 & 0 & 0 \\
0 & 0 & 0 & 0 \\
0 & 0 & 1 & 1 \\
0 & 0 & 1 & 1
\end{pmatrix}, \quad \bfB = \begin{pmatrix}
1 & \alpha \\
\alpha & 1 \\
& & 1 & \alpha \\
& & \alpha & 1
\end{pmatrix} \defEqual \begin{pmatrix}
\hat{\bfB} & \\
& \hat{\bfB}
\end{pmatrix}~.
\end{IEEEeqnarray*}
We see that this change of basis by $\tilde{\bfP}$ makes the matrices $\tilde{\bfB}$ and $\tilde{\bfC}$ block-diagonal while it destroys the block-diagonal structure of $\tGab$ and $\tilde{\bfM}$. We will see in \Cref{lemma:eig_Aref} that $\Gab$ is still block-diagonal at initialization. We will use the tilde quantities as an intermediate step to derive equations for the non-tilde quantities, since the latter will be more suitable for us to analyze eigenvectors and eigenvalues. %

Elementary arguments show that
\begin{IEEEeqnarray*}{+rCl+x*}
(\bfM_1 \matgr 0 \text{ and } \bfM_{-1} \matgr 0) & \Leftrightarrow & \tilde{\bfM} \matgr 0 \Leftrightarrow \bfM = \tilde{\bfP}\tilde{\bfM}\tilde{\bfP}^\top \matgr 0~.
\end{IEEEeqnarray*}
Therefore, we need to require $\bfM \matgr 0$ so that $\vopt$ and $\ovl{\bfv}$ can be defined.

\item Many of the quantities above depend on the data set $D$, which we may highlight later by indexing them with $D$. For example, we may write $\bfu_D$ instead of $\bfu$.

\item Finally, let
\begin{IEEEeqnarray*}{+rCl+x*}
\bftheta_i & \equalDef & \begin{pmatrix}
a_i \\ b_i \\ w_i
\end{pmatrix}, \quad \bfSigma_\sigma \equalDef \sum_{i \in I_\sigma} \bftheta_i \bftheta_i^\top = \begin{pmatrix}
\Sigma_{\sigma, a^2} & \Sigma_{\sigma, ab} & \Sigma_{\sigma, wa} \\
\Sigma_{\sigma, ab} & \Sigma_{\sigma, b^2} & \Sigma_{\sigma, wb} \\
\Sigma_{\sigma, wa} & \Sigma_{\sigma, wb} & \Sigma_{\sigma, w^2} 
\end{pmatrix}, \quad \bfQ_\sigma \equalDef \begin{pmatrix}
0 & 0 & r_\sigma \\
0 & 0 & s_\sigma \\
r_\sigma & s_\sigma & 0
\end{pmatrix}.
\end{IEEEeqnarray*}
These quantities will be analyzed in the next proposition. \qedhere
\end{enumerate}
\end{definition}

\begin{proposition} \label{prop:deriv:theta_Sigma}
For $i \in I_\sigma, \sigma \in \{\pm 1\}$, we have
\begin{IEEEeqnarray*}{+rCl+x*}
\delta \bftheta_i & = & h\bfQ_\sigma \bftheta_i \\
\delta c & = & h(\hat{s}_1 + \hat{s}_{-1}) \\
\delta \bfSigma_\sigma & = & h\bfQ_\sigma \bfSigma_\sigma + h\bfSigma_\sigma \bfQ_\sigma + h^2 \bfQ_\sigma \bfSigma_\sigma \bfQ_\sigma
\end{IEEEeqnarray*}
and the latter identity can also be written as
\begin{IEEEeqnarray*}{+rCl+x*}
\bfSigma_{\sigma, k+1} & = & (\bfI_3 + h\bfQ_{\sigma, k})\bfSigma_{\sigma, k}(\bfI_3 + h\bfQ_{\sigma, k})~.
\end{IEEEeqnarray*}

\begin{proof}
The first two equations can also be written as
\begin{IEEEeqnarray*}{+rCl+x*}
\delta a_i & = & hr_\sigma w_i \\
\delta b_i & = & hs_\sigma w_i \\
\delta w_i & = & hr_\sigma a_i + hs_\sigma b_i \\
\delta c & = & h(\hat{s}_1 + \hat{s}_{-1})~.
\end{IEEEeqnarray*}
We will prove the first of these equations, the other ones follow similarly. Set $g(W) \equalDef a_i$. With \Cref{lemma:discrete_diffcalc} (a), we obtain
\begin{IEEEeqnarray*}{+rCl+x*}
\delta a_i & = & \delta g(W) = -h g(\nabla L_{D, \bftau}(W)) = -h\frac{\partial L_{D, \bftau}}{\partial a_i}(W) \\
& = & -h\frac{1}{n}\sum_{j \in J} (f_{W, \bftau, \sgn(x_j)}(x_j) - y_j) \varphi'(\tau_i \cdot \sgn(x_j)) w_i x_j \\
& = & -h\frac{1}{n}\left(\sum_{j \in J_\sigma} (f_{W, \bftau, \sigma}(x_j) - y_j) w_i x_j + \alpha \sum_{j \in J_{-\sigma}} (f_{W, \bftau, -\sigma}(x_j) - y_j) w_i x_j\right) \\
& = & h(\hat{r}_\sigma + \alpha \hat{r}_{-\sigma}) w_i = hr_\sigma w_i~.
\end{IEEEeqnarray*}

Now for $\bfSigma_\sigma$: Since $\bfQ_\sigma = \bfQ_\sigma^\top$, we have
\begin{IEEEeqnarray*}{+rCl+x*}
\bfSigma_{\sigma, k+1} & = & \sum_{i \in I_\sigma} \bftheta_{i, k+1} \bftheta_{i, k+1}^\top = \sum_{i \in I_\sigma} (\bfI_3 + h\bfQ_{\sigma, k}) \bftheta_{i, k} \bftheta_{i, k}^\top (\bfI_3 + h\bfQ_{\sigma, k})^\top \\
& = & (\bfI_3 + h\bfQ_{\sigma, k}) \left(\sum_{i \in I_\sigma}\bftheta_{i, k} \bftheta_{i, k}^\top \right) (\bfI_3 + h\bfQ_{\sigma, k})^\top = (\bfI_3 + h\bfQ_{\sigma, k}) \bfSigma_{\sigma, k} (\bfI_3 + h\bfQ_{\sigma, k})~,
\end{IEEEeqnarray*}
which means that 
\begin{IEEEeqnarray*}{+rCl+x*}
\delta \bfSigma_k & = & \bfSigma_{k+1} - \bfSigma_k = h\bfQ_{\sigma, k} \bfSigma_{\sigma, k} + h \bfSigma_{\sigma, k} \bfQ_{\sigma, k} + h^2\bfQ_{\sigma, k} \bfSigma_{\sigma, k} \bfQ_{\sigma, k}~. & \qedhere
\end{IEEEeqnarray*}
\end{proof}
\end{proposition}

\begin{remark}
The term $h^2 \bfQ_\sigma \bfSigma_\sigma \bfQ_\sigma$ in \Cref{prop:deriv:theta_Sigma} corresponds to the term $\delta g_1 \cdot \delta g_2$ in the \quot{product rule} for $\delta$ (\Cref{lemma:discrete_diffcalc} (c)). It vanishes when using negative gradient flow. In our case, it does not affect the qualitative behavior of gradient descent.
\end{remark}

The following lemma shows relations between several quantities from \Cref{def:derived_quantities}.

\begin{lemma} \label{lemma:deriv:linear_relations}
Let $\bfM \matgr 0$. For $\sigma \in \{\pm 1\}$ and $x \in \bbR$, we have
\begin{IEEEeqnarray*}{+rCl+x*}
f_{W, \bftau, \sigma}(x) & = & p_\sigma x + q_\sigma \\
\hat{\bfu}_\sigma & = & -\bfM_\sigma \ovl{\bfv}_\sigma~.
\end{IEEEeqnarray*}
Moreover,
\begin{IEEEeqnarray*}{+rCl+x*}
\tilde{\bfu} = \tilde{\bfB} \tilde{\hat{\bfu}}, \quad \tilde{\hat{\bfu}} = -\tilde{\bfM} \tilde{\ovl{\bfv}}, \quad \tilde{\bfv} = \tilde{\bfB} \tilde{\hat{\bfv}} + \begin{pmatrix}
0 \\ c \\ 0 \\ c
\end{pmatrix}~.
\end{IEEEeqnarray*}

\begin{proof}
For $x \in \bbR$, 
\begin{IEEEeqnarray*}{+rCl+x*}
f_{W, \bftau, \sigma}(x) & = & c + \sum_{i \in I} w_i \varphi'(\tau_i \sigma) (a_i x + b_i) \\
& = & c + \sum_{i \in I_\sigma} (w_i a_i x + w_i b_i) + \alpha \sum_{i \in I_{-\sigma}} (w_i a_i x + w_i b_i) = p_\sigma x + q_\sigma~.
\end{IEEEeqnarray*}
Therefore, using \Cref{def:main:quantities},
\begin{IEEEeqnarray*}{+rCl+x*}
\hat{\bfu}_\sigma & = & \begin{pmatrix} \hat{r}_\sigma \\ \hat{s}_\sigma \end{pmatrix} = -\frac{1}{n} \begin{pmatrix}
\sum_{j \in J_\sigma} (f_{W, \bftau, \sigma}(x_j) - y_j) x_j \\
\sum_{j \in J_\sigma} (f_{W, \bftau, \sigma}(x_j) - y_j)
\end{pmatrix} \\
& = & -\frac{1}{n} \begin{pmatrix}
\sum_{j \in J_\sigma} (p_\sigma x_j + q_\sigma - y_j) x_j \\
\sum_{j \in J_\sigma} (p_\sigma x_j + q_\sigma - y_j)
\end{pmatrix} \\
& = & -\frac{1}{n} \begin{pmatrix}
p_\sigma \sum_{j \in J_\sigma} x_j^2 + q_\sigma \sum_{j \in J_\sigma} x_j - \sum_{j \in J_\sigma} x_j y_j \\
p_\sigma \sum_{j \in J_\sigma} x_j + q_\sigma \sum_{j \in J_\sigma} 1 - \sum_{j \in J_\sigma} y_j
\end{pmatrix} \\
& = & -\bfM_\sigma \begin{pmatrix}
p_\sigma \\ q_\sigma
\end{pmatrix} + \frac{1}{n} \sum_{j \in J_\sigma} \hat{\bfu}^0_{(x_j, y_j)} = -\bfM_\sigma \bfv_\sigma + \hat{\bfu}^0_\sigma = -\bfM_\sigma(\bfv_\sigma - \vopt_\sigma) = -\bfM_\sigma \ovl{\bfv}_\sigma~.
\end{IEEEeqnarray*}

We now obtain
\begin{IEEEeqnarray*}{+rCl+x*}
\tilde{\bfu} & = & \begin{pmatrix}
r_1 \\ s_1 \\ r_{-1} \\ s_{-1}
\end{pmatrix} = \begin{pmatrix}
\hat{r}_1 + \alpha \hat{r}_{-1} \\
\hat{s}_1 + \alpha \hat{s}_{-1} \\
\hat{r}_{-1} + \alpha \hat{r}_1 \\
\hat{s}_{-1} + \alpha \hat{s}_1
\end{pmatrix} = \tilde{\bfB} \tilde{\hat{\bfu}} \\
\tilde{\hat{\bfu}} & = & \begin{pmatrix}
\hat{\bfu}_1 \\ \hat{\bfu}_{-1}
\end{pmatrix} = \begin{pmatrix}
\bfM_1 \\
& \bfM_{-1}
\end{pmatrix} \begin{pmatrix}
\ovl{\bfv}_1 \\ \ovl{\bfv}_{-1}
\end{pmatrix} = \tilde{\bfM} \tilde{\ovl{\bfv}} \\
\tilde{\bfv} & = & \begin{pmatrix}
p_1 \\ q_1 \\ p_{-1} \\ q_{-1}
\end{pmatrix} = \begin{pmatrix}
\hat{p}_1 + \alpha \hat{p}_{-1} \\
c + \hat{q}_1 + \alpha \hat{q}_{-1} \\
\hat{p}_{-1} + \alpha \hat{p}_1 \\
c + \hat{q}_{-1} + \alpha \hat{q}_1
\end{pmatrix} = \tilde{\bfB} \tilde{\hat{\bfv}} + \begin{pmatrix}
0 \\ c \\ 0 \\ c
\end{pmatrix}~. & \qedhere
\end{IEEEeqnarray*}
\end{proof}
\end{lemma}

This enables us to compute another iteration equation:

\begin{proposition} \label{prop:deriv:AM}
Let $\bfM \matgr 0$. Then, %
\begin{IEEEeqnarray*}{+rCl+x*}
\delta \tilde{\ovl{\bfv}} & = & -h\tilde{\bfA}\tilde{\bfM}\tilde{\ovl{\bfv}} = -h(\tilde{\bfB}(\tGw + \tGab + h\tGwab)\tilde{\bfB} + \tilde{\bfC})\tilde{\bfM} \tilde{\ovl{\bfv}} \\
\delta \ovl{\bfv} & = & -h\bfA\bfM\ovl{\bfv} = -h(\bfB(\Gw + \Gab + h\Gwab)\bfB + \bfC)\bfM\ovl{\bfv}~.
\end{IEEEeqnarray*}
Hence,
\begin{IEEEeqnarray*}{+rCl+x*}
\ovl{\bfv}_{k+1} = \ovl{\bfv}_k + \delta \ovl{\bfv}_k = (\bfI_4 - h\bfA_k \bfM)\ovl{\bfv}_k~.
\end{IEEEeqnarray*}

\begin{proof}
Consider
\begin{IEEEeqnarray*}{+rCl+x*}
\hat{\bfv}_\sigma & = & \begin{pmatrix}
\hat{p}_\sigma \\ \hat{q}_\sigma
\end{pmatrix} = \begin{pmatrix}
\Sigma_{\sigma, wa} \\ \Sigma_{\sigma, wb}
\end{pmatrix} = \begin{pmatrix}
\bfI_2 & 0
\end{pmatrix} \bfSigma_{\sigma} \begin{pmatrix}
0 \\ 0 \\ 1
\end{pmatrix}~.
\end{IEEEeqnarray*}
Using \Cref{prop:deriv:theta_Sigma} and \Cref{lemma:discrete_diffcalc}, we obtain
\begin{IEEEeqnarray*}{+rCl+x*}
\delta \hat{\bfv}_\sigma & = & \begin{pmatrix}
\bfI_2 & 0
\end{pmatrix} \left(h\bfQ_\sigma \bfSigma_\sigma + h\bfSigma_\sigma \bfQ_\sigma + h^2\bfQ_\sigma \bfSigma_\sigma \bfQ_\sigma\right) \begin{pmatrix}
0 \\ 0 \\ 1
\end{pmatrix} \\
& = & h\begin{pmatrix}
0 & 0 & r_\sigma \\
0 & 0 & s_\sigma
\end{pmatrix} \begin{pmatrix}
\Sigma_{\sigma, wa} \\
\Sigma_{\sigma, wb} \\
\Sigma_{\sigma, w^2}
\end{pmatrix} + h\begin{pmatrix}
\Sigma_{\sigma, a^2} & \Sigma_{\sigma, ab} & \Sigma_{\sigma, wa} \\
\Sigma_{\sigma, ab} & \Sigma_{\sigma, b^2} & \Sigma_{\sigma, wb}
\end{pmatrix} \begin{pmatrix}
r_\sigma \\
s_\sigma \\
0
\end{pmatrix} \\
&& ~+~ h^2\begin{pmatrix}
0 & 0 & r_\sigma \\
0 & 0 & s_\sigma
\end{pmatrix} \begin{pmatrix}
\Sigma_{\sigma, a^2} & \Sigma_{\sigma, ab} & \Sigma_{\sigma, wa} \\
\Sigma_{\sigma, ab} & \Sigma_{\sigma, b^2} & \Sigma_{\sigma, wb} \\
\Sigma_{\sigma, wa} & \Sigma_{\sigma, wb} & \Sigma_{\sigma, w^2}
\end{pmatrix} \begin{pmatrix}
r_\sigma \\
s_\sigma \\
0
\end{pmatrix} \\
& = & h\begin{pmatrix}
\Sigma_{\sigma, w^2} & 0 \\
0 & \Sigma_{\sigma, w^2}
\end{pmatrix} \bfu_\sigma + h\begin{pmatrix}
\Sigma_{\sigma, a^2} & \Sigma_{\sigma, ab} \\
\Sigma_{\sigma, ab} & \Sigma_{\sigma, b^2}
\end{pmatrix} \bfu_\sigma + h^2 (r_\sigma \Sigma_{\sigma, wa} + s_\sigma \Sigma_{\sigma, wb}) \bfu_\sigma \\
& = & h \left(\Gw_\sigma + \Gab_\sigma + h \Gwab_\sigma\right) \bfu_\sigma~.
\end{IEEEeqnarray*} 
Therefore, $\delta \tilde{\hat{\bfv}} = h \left(\tGw + \tGab + h \tGwab\right) \tilde{\bfu}$. Also,
\begin{IEEEeqnarray*}{+rCl+x*}
\delta \begin{pmatrix}
0 \\ c \\ 0 \\ c
\end{pmatrix} = h\begin{pmatrix}
0 \\ \hat{s}_1 + \hat{s}_{-1} \\ 0 \\ \hat{s}_1 + \hat{s}_{-1}
\end{pmatrix} = h\begin{pmatrix}
0 & 0 & 0 & 0 \\
0 & 1 & 0 & 1 \\
0 & 0 & 0 & 0 \\
0 & 1 & 0 & 1
\end{pmatrix} \begin{pmatrix}
\hat{r}_1 \\ \hat{s}_1 \\ \hat{r}_{-1} \\ \hat{s}_{-1}
\end{pmatrix} = h\tilde{\bfC}\tilde{\hat{\bfu}}~.
\end{IEEEeqnarray*}
We can now use the identities from \Cref{lemma:deriv:linear_relations} and the fact that $\tilde{\bfv} - \tilde{\ovl{\bfv}} = \tildevopt$ is constant to compute
\begin{IEEEeqnarray*}{+rCl+x*}
\delta \tilde{\ovl{\bfv}} & = & \delta \tilde{\bfv} = \tilde{\bfB} \delta \tilde{\hat{\bfv}} + \delta \begin{pmatrix}
0 \\ c \\ 0 \\ c
\end{pmatrix} = \tilde{\bfB} h \left(\tGw + \tGab + h \tGwab\right) \tilde{\bfu} + h\tilde{\bfC}\tilde{\hat{\bfu}} \\
& = & h(\tilde{\bfB}(\tGw + \tGab + h\tGwab)\tilde{\bfB} + \tilde{\bfC}) \tilde{\hat{\bfu}} = h\tilde{\bfA}\tilde{\hat{\bfu}} = -h\tilde{\bfA}\tilde{\bfM}\tilde{\ovl{\bfv}}~.
\end{IEEEeqnarray*}

Since $\tilde{\bfP}^2 = \bfI_4$, it follows that
\begin{IEEEeqnarray*}{+rCl+x*}
\delta \ovl{\bfv} = \delta (\tilde{\bfP} \tilde{\ovl{\bfv}}) = \tilde{\bfP} \delta \tilde{\ovl{\bfv}} = -h\tilde{\bfP}\tilde{\bfA}\tilde{\bfM}\tilde{\ovl{\bfv}} = -h\tilde{\bfP}\tilde{\bfA}\tilde{\bfP}\tilde{\bfP}\tilde{\bfM}\tilde{\bfP}\tilde{\bfP}\tilde{\ovl{\bfv}} = -h\bfA\bfM\ovl{\bfv}
\end{IEEEeqnarray*}
and
\begin{IEEEeqnarray*}{+rCl+x*}
\bfA & = & \tilde{\bfP}\tilde{\bfA}\tilde{\bfP} = \tilde{\bfP}\big(\tilde{\bfB}\tilde{\bfP}\tilde{\bfP}(\tGw + \tGab + h\tGwab)\tilde{\bfP}\tilde{\bfP}\tilde{\bfB} + \tilde{\bfC}\big)\tilde{\bfP} \\
& = & \bfB(\Gw + \Gab + h\Gwab)\bfB + \bfC~. & \qedhere
\end{IEEEeqnarray*}
\end{proof}
\end{proposition}

\section{Comments} \label{sec:comments}

In this section, we provide some remarks on the interpretation of the gradient descent equations derived in \Cref{sec:gradient_descent_equations}.

\begin{figure}[!htb]
\centering
\newcommand{\repeatNode}[4]{%
\node[n, draw, rectangle, fill=white] at (#1 + 0.2, #2 - 0.2) {\begin{NoHyper} \textcolor{white}{#4} \end{NoHyper}}; %
\node[n, draw, rectangle, fill=white] at (#1 + 0.1, #2 - 0.1) {\begin{NoHyper} \textcolor{white}{#4} \end{NoHyper}};
\node[n, draw, rectangle, fill=white] (#3) at (#1, #2) {#4}
}
\begin{tikzpicture}[n/.style={minimum width=4cm, anchor=north west, align=center, text width=6.9cm}]
\node[n, draw, rectangle] (sigmaone) at (0, 0) {\small $\bfSigma_{1, k+1} = (\bfI_3+h\bfQ_{1,k})\bfSigma_{1,k}(\bfI_3+h\bfQ_{1,k})$ \\
where $\bfQ_{1, k}$ depends on $\ovl{\bfv}_k$ \\
$9$-dimensional (\Cref{prop:deriv:theta_Sigma})};
\node[n, draw, rectangle] (sigmaminusone) at (8, 0) {\small $\bfSigma_{-1, k+1} = (\bfI_3+h\bfQ_{-1,k})\bfSigma_{-1,k}(\bfI_3+h\bfQ_{-1,k})$ \\ where $\bfQ_{-1, k}$ depends on $\ovl{\bfv}_k$ \\
$9$-dimensional (\Cref{prop:deriv:theta_Sigma})};

\node[n, draw, rectangle] (v) at (4, -3) {\small $\delta \ovl{\bfv} = -h\bfA\bfM\ovl{\bfv}$ \\
where $\bfA$ depends on $\ovl{\bfv}, \bfSigma_1, \bfSigma_{-1}$ \\
$4$-dimensional (\Cref{prop:deriv:AM})};

\node[anchor=north west] (thetalabel) at (2, -7.9) {\small for $i \in I_\sigma, \sigma \in \{\pm 1\}$};
\repeatNode{0}{-6}{theta}{$\delta \bftheta_i = h\bfQ_\sigma \bftheta_i$ \\
where $\bfQ_\sigma$ depends on $\ovl{\bfv}$ \\
$3$-dimensional (\Cref{prop:deriv:theta_Sigma})};

\node[n, draw, rectangle] (c) at (8, -6) {\small $\delta c = h(\hat{s}_1 + \hat{s}_{-1})$ \\
where $\hat{s}_1, \hat{s}_{-1}$ depend on $\ovl{\bfv}$ \\
$1$-dimensional (\Cref{prop:deriv:theta_Sigma})};

\draw[Latex-Latex] (v) -- node[xshift=0.2cm, anchor=west]{affects} (sigmaone);
\draw[Latex-Latex] (v) -- node[anchor=west]{affects} (sigmaminusone);
\draw[-Latex] (v) -- node[anchor=west]{affects} (theta);
\draw[-Latex] (v) -- node[xshift=0.2cm, anchor=west]{affects} (c);
\end{tikzpicture}
\caption{Decomposition into different systems that can be used to analyze the behavior of gradient descent.} \label{fig:systems}
\end{figure}

\begin{remark}[System decomposition]
We have so far derived different \quot{systems}, i.e., results on how quantities evolve during gradient descent. These systems and their dependencies are depicted in \Cref{fig:systems}. In particular, we see that the systems for $\bfSigma_1, \bfSigma_{-1}$ and $\ovl{\bfv}$ together yield a $22$-dimensional system that does not depend on any other quantities. This $22$-dimensional system describes some central properties of the neural network parameters $W$ although its dimension does not depend on $m$. These properties include:
\begin{itemize}

\item Slope $p_\sigma$ and intercept $q_\sigma$ for both signs $\sigma \in \{\pm 1\}$.
\item The loss $L_{D, \bftau}(W)$, which can be computed from $p_\sigma$ and $q_\sigma$.
\end{itemize}
While this system has a dimension independent of $m$, the probability distribution over its initialization may well depend on $m$. If its evolution is known, the evolution $(W_k)_{k \in \bbN_0}$ can be determined by solving $m$ independent three-dimensional systems and the one-dimensional system $\delta c = h(\hat{s}_1 + \hat{s}_{-1})$. Here, we will proceed along similar lines: We will first analyze the behavior of the $22$-dimensional system and then apply our results to the three-dimensional systems.

In fact, the $22$-dimensional system can be reduced to a $14$-dimensional system: The matrices $\bfSigma_\sigma$ are always symmetric and thus effectively $6$-dimensional, which reduces the dimension from $22$ to $16$. Moreover, we always have
\begin{IEEEeqnarray*}{+rCl+x*}
\begin{pmatrix}
p_1 \\ p_{-1}
\end{pmatrix} & = & \begin{pmatrix}
1 & \alpha \\
\alpha & 1
\end{pmatrix} \begin{pmatrix}
\Sigma_{1, wa} \\
\Sigma_{-1, wa}
\end{pmatrix}~.
\end{IEEEeqnarray*}
However, removing these redundancies is not beneficial for our analysis.
\end{remark}

\begin{remark} \label{remark:deriv:interpretation_and_loss}
The components of the equation $\delta \ovl{\bfv} = -h\bfA\bfM\ovl{\bfv}$ in \Cref{prop:deriv:AM} can be interpreted as follows: Recall that
\begin{IEEEeqnarray*}{+rClrClrCl+x*}
\Gw_\sigma & = & \begin{pmatrix}
\Sigma_{\sigma, w^2} & 0 \\
0 & \Sigma_{\sigma, w^2}
\end{pmatrix}, \quad & \Gab_\sigma & = & \begin{pmatrix}
\Sigma_{\sigma, a^2} & \Sigma_{\sigma, ab} \\
\Sigma_{\sigma, ab} & \Sigma_{\sigma, b^2}
\end{pmatrix}, \quad & \Gwab_\sigma & = & (r_\sigma \Sigma_{\sigma, wa} + s_\sigma\Sigma_{\sigma, wb}) \bfI_2, \\
\tilde{\bfB} & = & \begin{pmatrix}
\bfI_2 & \alpha \bfI_2 \\
\alpha \bfI_2 & \bfI_2
\end{pmatrix}, \quad & \tilde{\bfC} & = & \begin{pmatrix}
0 & 0 & 0 & 0 \\
0 & 1 & 0 & 1 \\
0 & 0 & 0 & 0 \\
0 & 1 & 0 & 1 
\end{pmatrix}, \quad & \\
\tilde{\bfA} & = & \IEEEeqnarraymulticol{7}{l}{\tilde{\bfB} \begin{pmatrix}
\Gw_1 + \Gab_1 + h\Gwab_1 \\
& \Gw_{-1} + \Gab_{-1} + h\Gwab_{-1}
\end{pmatrix}\tilde{\bfB} + \tilde{\bfC}~.}
\end{IEEEeqnarray*}
\begin{itemize}

\item The matrix $\Gw_\sigma \matgeq 0$ describes the improvement of $\ovl{\bfv}_\sigma$ by updating the weights $(a_i)_{i \in I_\sigma}$ and $(b_i)_{i \in I_\sigma}$. The larger $|w_i|$, the larger the gradients $\frac{\partial L_{D, \bftau}}{\partial a_i}, \frac{\partial L_{D, \bftau}}{\partial b_i}$ and the more effect does a change in $a_i, b_i$ have on the overall function $f_{W, \bftau, \sigma}$. 

\item The matrix $\Gab_\sigma$ is also positive semidefinite since $\tr(\Gab_\sigma) \geq 0$ and $\det(\Gab_\sigma) = \Sigma_{\sigma, a^2}\Sigma_{\sigma, b^2} - \Sigma_{\sigma, ab}^2 \geq 0$ due to Cauchy-Schwarz. It describes the improvement of $\ovl{\bfv}_\sigma$ by updating the weights $(w_i)_{i \in I_\sigma}$. Larger values of $|a_i|, |b_i|$ mean stronger effects of changing $w_i$. If the vectors $(a_i)_{i \in I_\sigma}$ and $(b_i)_{i \in I_\sigma}$ are linearly dependent (perfectly correlated), e.g.\ at initialization because of $b_{i, 0} = 0$, then $\tGab_\sigma$ only has rank one and changing the $w_i$ cannot independently update both components of $\ovl{\bfv}_\sigma$. Recall that the components of $\ovl{\bfv}_\sigma$ are the differences of the slope and intercept of $f_{W, \bftau, \sigma}$ to the optimal linear regression slope and intercept, respectively.

\item The matrix $\tilde{\bfB}$ causes an interaction between both signs $\sigma \in \{\pm 1\}$ if the leaky parameter $\alpha$ is nonzero. If it is zero, the hidden neurons are only active for one sign $\sigma$ and do only interact indirectly via the bias $c$.

\item The matrix $\tilde{\bfC}$ describes the improvement of $v_\sigma$ by updating the bias $c$. It is not block-diagonal since $c$ is active for both signs $\sigma \in \{\pm 1\}$. However, $\tilde{\bfC}$ only has rank one since changing $c$ can only change $q_1$ and $q_{-1}$ by the same amount. $\tilde{\bfC}$ is positive semidefinite since it is symmetric and it has eigenvectors $\bfe_1, \bfe_3, (0, 1, 0, -1)$ to the eigenvalue $0$ and $(0, 1, 0, 1)$ to the eigenvalue $2$.

\item The matrix $\tGwab$ represents parts of the error that (discrete) gradient descent makes when trying to approximate negative gradient flow. It arises from the additional term $\delta g_1 \cdot \delta g_2$ in the product rule for $\delta$ (\Cref{lemma:discrete_diffcalc} (c)) and does not need to be positive semidefinite. If $h$ is too large, the matrix $\tilde{A}$ might therefore not be positive semidefinite. \qedhere %
\end{itemize}

\end{remark}

\begin{remark} [Discretization error]
We have already seen that the systems for $\bfSigma_\sigma$ and $\ovl{\bfv}$ are affected by terms that arise from the term $\delta g_1 \cdot \delta g_2$ in the \quot{discrete product rule} of \Cref{lemma:discrete_diffcalc} (c). We will see that in our scenario (with small enough step size), these \quot{disturbances} are small enough to not influence the qualitative behavior of gradient descent. There is also an invariant that holds when using negative gradient flow but breaks down when using gradient descent: In the former case, $a_i^2 + b_i^2 - w_i^2$ remains constant during the optimization for each $i \in I$. An analogous identity for linear networks has been observed by \citet{saxe_exact_2014}.
\end{remark}

\begin{remark}[Alternative systems] \label{rem:alternative_systems}
In some special cases, the approach presented here only works if we modify the systems. For example, the assumption $\bfM \matgr 0$ is not satisfied if the data set is contained in $(0, \infty)$ since this implies $\bfM_{-1} = 0$. In this case, the system $\delta \ovl{\bfv} = -h\bfA\bfM\ovl{\bfv}$ could be reduced to a two-dimensional system since $p_{-1}$ and $q_{-1}$ are irrelevant for the loss. We will also see that the argument here does not work for $|\alpha| = 1$ since this renders the matrix $\bfB$ singular. The case $\alpha = 1$ corresponds to a linear activation function $\varphi(x) = x$, which implies $p_1 = p_{-1}$ and $q_1 = q_{-1}$. Similarly, the case $\alpha = -1$ corresponds to $\varphi(x) = |x|$, which implies $p_1 = -p_{-1}$. In both cases, the dimension of $\bfv$ could be reduced. 
\end{remark}

\begin{remark}[Calculations for \Cref{rem:MD-invert}]\label{rem-check-MD-invert}
To verify the assertions made in \Cref{rem:MD-invert},
we first note that $\bfM_{D, \sigma}$ is given by 
\begin{align} \label{form-of-mds}
 \bfM_{D, \sigma} &:=
 \frac{n_\sigma}{n} \begin{pmatrix}
\alpha & \beta \\
\beta & 1
\end{pmatrix} \, ,
\end{align}
where
\begin{IEEEeqnarray*}{+rCl+x*}
\alpha & \equalDef & \frac{1}{n_\sigma} \sum_{(x, y) \in D_\sigma} x^2 = \Var D_{X, \sigma} + \left(\bbE D_{X, \sigma}\right)^2~, \\
\beta & \equalDef & \frac{1}{n_\sigma} \sum_{(x, y) \in D_\sigma} x = \bbE D_{X, \sigma}~.
\end{IEEEeqnarray*}
Consequently, we have 
\begin{IEEEeqnarray*}{+rCl+x*}
\det(\bfM_{D, \sigma}) = \left(\frac{n_\sigma}{n}\right)^2 (\alpha - \beta^2) = \left(\frac{n_\sigma}{n}\right)^2 \Var D_{X, \sigma},
\end{IEEEeqnarray*}
and since $\Var D_{X,\sigma} \neq 0$ if and only if $D_{X,\sigma}$ contains at least two distinct samples, 
it follows that $\bfM_{D_\sigma}$ is invertible if and only if $D_\sigma$ contains at least two samples with different $x$ values. 

Now assume that $\Var D_{X, \sigma} > 0$. Since $\det(\bfM_{D, \sigma})$ and $\tr(\bfM_{D, \sigma})$ are positive, the minimum and maximum eigenvalues $\lmin, \lmax$ of $\bfM_{D, \sigma}$ are both positive. Thus, 
\begin{IEEEeqnarray*}{+rCl+x*}
\lmax & \leq & \lmin + \lmax = \tr(\bfM_{D, \sigma}) = \frac{n_\sigma}{n}(\alpha + 1)~, \\
\lmin & = & \frac{\lmin \lmax}{\lmax} \geq \frac{\lmin \lmax}{\lmin + \lmax} = \frac{\det(\bfM_{D, \sigma})}{\tr(\bfM_{D, \sigma})} = \frac{\left(\frac{n_\sigma}{n}\right)^2 (\alpha - \beta^2)}{\left(\frac{n_\sigma}{n}\right)(\alpha + 1)} = \frac{n_\sigma}{n} \cdot \frac{\Var D_{X,\sigma}}{\alpha + 1}~.
\end{IEEEeqnarray*}
Since $\frac{1}{2}(\lmin + \lmax) \leq \lmax \leq \lmin + \lmax$, it is easy to see that the bounds are off by a factor of at most two.
\end{remark}

\begin{remark}[Affine regression optimum] \label{rem:affine_regression}
Here, we review some well-known properties of performing affine least-squares regression on a data set $D$. This analysis also applies to $D_1$ and $D_{-1}$. Consider
\begin{IEEEeqnarray*}{+rCl+x*}
\bfX & \equalDef & \begin{pmatrix}
x_1 & 1 \\
\vdots & \vdots \\
x_n & 1
\end{pmatrix}, \quad \bfy \equalDef \begin{pmatrix}
y_1 \\
\vdots \\
y_n
\end{pmatrix}, \bfv \equalDef \begin{pmatrix}
p \\ q
\end{pmatrix}~.
\end{IEEEeqnarray*}
We always have $\bfX^\top \bfX \matgeq 0$. Assume that $\bfX$ has full column rank such that $\bfX^\top \bfX \matgr 0$. The least-squares risk of an affine function $f_{\bfv}(x) = px + q$ is $R_D(f_{\bfv})$ with
\begin{IEEEeqnarray*}{+rCl+x*}
2n R_D(f_{\bfv}) & = & \sum_{j=1}^n (y_j - f_{\bfv}(x_j))^2 = \|\bfy - \bfX \bfv\|_2^2 = (\bfy - \bfX \bfv)^\top (\bfy - \bfX \bfv) \\
& = & \left(\bfv - (\bfX^\top \bfX)^{-1} \bfX^\top \bfy\right)^\top \bfX^\top \bfX \left(\bfv - (\bfX^\top \bfX)^{-1} \bfX^\top \bfy\right) \\
&& ~+~ \left(\bfy^\top \bfy - \bfy^\top \bfX (\bfX^\top \bfX)^{-1} \bfX^\top \bfy\right)~, \IEEEyesnumber \label{eq:loss_factorization}
\end{IEEEeqnarray*}
where we performed a completion of the square. Therefore, the optimal affine function has parameters
\begin{IEEEeqnarray*}{+rCl+x*}
\vopt & = & (\bfX^\top \bfX)^{-1} \bfX^\top \bfy = \begin{pmatrix}
\sum_j x_j^2 & \sum_j x_j \\
\sum_j x_j & \sum_j 1
\end{pmatrix}^{-1} \begin{pmatrix}
\sum_j x_j y_j \\
\sum_j y_j
\end{pmatrix} \\
& = & \left(\frac{1}{n} \begin{pmatrix}
\sum_j x_j^2 & \sum_j x_j \\
\sum_j x_j & \sum_j 1
\end{pmatrix} \right)^{-1} \left(\frac{1}{n} \begin{pmatrix}
\sum_j x_j y_j \\
\sum_j y_j
\end{pmatrix}\right).
\end{IEEEeqnarray*}
Note that applying Eq.~\eqref{eq:loss_factorization} to $D_1$ and $D_{-1}$ after some rearrangement yields
\begin{IEEEeqnarray*}{+rCl+x*}
L_{D, \bftau}(W) & = & \frac{1}{2} (\bfv - \vopt_D)^\top \bfM_D (\bfv - \vopt_D) + \text{const} = \frac{1}{2} \ovl{\bfv}^\top \bfM_D \ovl{\bfv} + \text{const}~,
\end{IEEEeqnarray*}
where the constant term is the optimal achievable loss by affine regression on $D_1$ and $D_{-1}$.
\end{remark}

\section{Stochastic Proofs} \label{sec:stochastics}

In this section, we show that $W_0$ and $D$ likely have certain properties. The results are formulated in \Cref{prop:prob_W} and \Cref{prop:sampling:appendix}, respectively. In order to obtain these results, we employ concentration inequalities. Besides Markov's inequality, we use Hoeffding's inequality:

\begin{lemma}[Hoeffding's inequality, e.g.\ Lemma A.3 in \citet{gyorfi_distribution-free_2002}] \label{lemma:Hoeffding}
Let $(\Omega, \calF, P)$ be a probability space, $a < b, n \geq 1$ and $X_1, \hdots, X_n: \Omega \to [a, b]$ be independent random variables. Then, for $\tau \geq 0$, we have
\begin{IEEEeqnarray*}{+rCl+x*}
P\left(\left|\frac{1}{n} \sum_{i=1}^n \left(X_i - \bbE X_i\right)\right| \geq (b - a) \sqrt{\frac{\tau}{2n}}\right) \leq 2e^{-\tau}~.
\end{IEEEeqnarray*}
\end{lemma}

Using Markov and Hoeffding, we can prove an asymptotic concentration result. The intuition behind this result is that for random variables $X_1, \hdots, X_n$ with mean zero and finite variance, the value $n^{-1/2}(X_1 + \hdots + X_n)$ asymptotically has a Gaussian distribution by the central limit theorem. The tail of the Gaussian distribution decreases stronger than any inverse polynomial: If $\Phi$ is the CDF of a Gaussian distribution, then $\Phi(\beta n^\varepsilon) \leq O(n^{-\gamma})$ for all $\beta, \varepsilon, \gamma > 0$, where the constant in $O(n^{-\gamma})$ depends on $\beta, \varepsilon, \gamma$. However, the central limit theorem does not tell us how close the CDF of $n^{-1/2}(X_1 + \hdots + X_n)$ is to $\Phi$, so we use Markov's and Hoeffding's inequalities instead.

\begin{lemma} \label{lemma:concentration_inequality}
Let $Q$ be a probability distribution on $\bbR$ with $\mu_p \equalDef \int |x|^p \diff Q(x) < \infty$ for all $p \in (1, \infty)$. For $n \in \bbN$, let $(\Omega_n, \calF_n, P_n)$ be probability spaces with independent $Q$-distributed random variables $X_{n1}, X_{n2}, \hdots, X_{nn}: \Omega_n \to \bbR$. Then, the random variables $S_n \equalDef \frac{1}{n} \sum_{i=1}^n X_{ni}$ satisfy
\begin{IEEEeqnarray*}{+rCl+x*}
P_n\left(\left|S_n - \bbE S_n\right| \geq \beta n^{\varepsilon-1/2} \right) \leq O(n^{-\gamma})
\end{IEEEeqnarray*}
for all $\beta, \varepsilon, \gamma > 0$, where the constant in $O(n^{-\gamma})$ may depend on $\beta, \varepsilon, \gamma$ (cf. \Cref{def:O-notation}).

\begin{proof}
Let $\beta, \varepsilon, \gamma > 0$ be fixed. For $n \in \bbN$ and $b > 0$ to be determined later, define $B \equalDef \{\max_{1 \leq i \leq n} |X_{ni}| \leq b\}$. Then, for all $p > 0$, 
\begin{IEEEeqnarray*}{+rCl+x*}
P_n(B^c) & \leq & \sum_{i=1}^n P_n(|X_{ni}| \geq b) \leq nP_n(|X_{n1}|^p \geq b^p) \stackrel{\text{Markov}}{\leq} n\frac{\bbE_{P_n} |X_{n1}|^p}{b^p} = n\frac{\mu_p}{b^p}~. \IEEEyesnumber \label{eq:concentration_inequality:1}
\end{IEEEeqnarray*}
Since $S_n = S_n \bbone_B + S_n \bbone_{B^c}$, we can now bound
\begin{IEEEeqnarray*}{+rCl+x*}
P_n(|S_n - \bbE S_n| \geq \beta n^{\varepsilon-1/2}) & \leq & \underbrace{P_n(|S_n \bbone_B - \bbE(S_n \bbone_B)| \geq \beta n^{\varepsilon-1/2}/2)}_{\mathrm{I}} \\
&& ~+~ \underbrace{P_n(|S_n \bbone_{B^c} - \bbE(S_n \bbone_{B^c})| \geq \beta n^{\varepsilon-1/2}/2)}_{\mathrm{II}}.
\end{IEEEeqnarray*}
With $\tau \equalDef \gamma \log n$ and $b \equalDef \beta n^\varepsilon \sqrt{\frac{1}{8\gamma \log n}}$, we have 
\begin{IEEEeqnarray*}{+rCl+x*}
(b - (-b)) \sqrt{\frac{\tau}{2n}} & = & 2\beta n^\varepsilon \sqrt{\frac{1}{8\gamma \log n}} \cdot \sqrt{\frac{\gamma \log n}{2n}} = \beta n^{\varepsilon-1/2}/2
\end{IEEEeqnarray*}
and hence, Hoeffding (\Cref{lemma:Hoeffding}) applied to $X_i \equalDef X_{ni} \bbone_{|X_{ni}| \leq b}$ yields
\begin{IEEEeqnarray*}{+rCl+x*}
\mathrm{I} & \leq & 2e^{-\tau} = 2n^{-\gamma}~.
\end{IEEEeqnarray*}
Moreover, we have
\begin{IEEEeqnarray*}{+rCl+x*}
|\bbE_{P_n} (S_n \bbone_{B^c})| & \leq & \|S_n \bbone_{B^c}\|_{\calL_1(P_n)} \stackrel{\text{Hölder}}{\leq} \|S_n\|_{\calL_2(P_n)} \|\bbone_{B^c}\|_{\calL_2(P_n)} \\
& \leq & \left(\frac{1}{n}\sum_{i=1}^n \|X_{ni}\|_{\calL_2(P_n)}\right) \|\bbone_{B^c}\|_{\calL_2(P_n)} = \sqrt{\mu_2} \sqrt{P_n(B^c)} \\
& \stackrel{\eqref{eq:concentration_inequality:1}}{\leq} & \sqrt{\mu_2} \sqrt{n\frac{\mu_p}{b^p}} = \sqrt{\frac{\mu_2 \mu_p}{\beta^p}} (8\gamma \log n)^{p/4} n^{(1-\varepsilon p)/2}~.
\end{IEEEeqnarray*}
If we choose $p \geq 2/\varepsilon$, we have $(1-\varepsilon p)/2 \leq -1/2 < \varepsilon-1/2$ and hence $|\bbE (S_n \bbone_{B^c})| < \beta n^{\varepsilon-1/2}/2$ for $n$ large enough. Now, let $n$ be sufficiently large. For $\omega \in B$, we have $S_n(\omega) \bbone_{B^c}(\omega) = 0$ and hence $|S_n(\omega) \bbone_{B^c}(\omega) - \bbE (S_n \bbone_{B^c})| < \beta n^{\varepsilon-1/2}/2$. Thus,
\begin{IEEEeqnarray*}{+rCl+x*}
\mathrm{II} & \leq & P(B^c) \leq n\frac{\mu_p}{b^p} = \frac{\mu_p}{\beta^p} \cdot (8\gamma \log n)^{p/2} n^{1 - \varepsilon p}~.
\end{IEEEeqnarray*}
If we choose $p > (1+\gamma)/\varepsilon$, then $1 - \varepsilon p < -\gamma$ and hence $\mathrm{II} \leq O(n^{-\gamma})$.
\end{proof}
\end{lemma}

Now, we can prove that certain properties of the initialization $W_0$ hold with high probability. We will see that in all properties except (W4), the tails of the probability distributions decrease so quickly that only the parameter $\gP$ in (W4) is relevant for the rate of convergence.

\begin{proposition} \label{prop:prob_W}
Let $\varepsilon, \gP > 0$ and let $W_0$ be distributed as in \Cref{ass:init}. Then, the properties
\begin{enumerate}[(W1), wide=0pt, leftmargin=*]

\item $b_{i, 0} = c_0 = 0$, 
\item $\max_i |w_{i, 0}| \leq m^{-1/2+\varepsilon}$,
\item $\max_i |a_{i, 0}| \leq m^{\varepsilon}$,
\item $\min_i |a_{i, 0}| \geq m^{-(1+\gP)}$,
\item $\Sigma_{\sigma, a^2, 0} \in [m\vara/4, m\vara]$ for all $\sigma \in \{\pm 1\}$,
\item $\Sigma_{\sigma, w^2, 0} \in [\varw/4, \varw]$ for all $\sigma \in \{\pm 1\}$,
\item $|\Sigma_{\sigma, wa, 0}| \leq m^\varepsilon$ for all $\sigma \in \{\pm 1\}$. 
\end{enumerate}
are satisfied with probability $\geq 1 - O(m^{-\gP})$, where the constant in $O(m^{-\gP})$ may depend on $\varepsilon$ and $\gP$ (cf. \Cref{def:O-notation}).

\begin{proof}
We will show the statement for each of the properties (W1) -- (W7) individually, the rest follows by the union bound. Let $\Za, \Zw$ be the random variables from \Cref{ass:init}.

By property (Q2) in \Cref{ass:init}, $\bbE |\Za|^p, \bbE |\Zw|^p < \infty$ for all $p \in (0, \infty)$. It can be shown (using the Minkowski and Hölder inequalities) that all other random variables used below satisfy the same property, which we will use in order to apply \Cref{lemma:concentration_inequality}.
\begin{enumerate}[(W1), wide=0pt, leftmargin=*]

\item True by \Cref{ass:init}.
\item By the Markov inequality, for $p > 0$,
\begin{IEEEeqnarray*}{+rCl+x*}
\Pinitm\left(|w_{i, 0}| \geq m^{-1/2+\varepsilon}\right) & = & \Pinitm\left(|w_{i, 0}|^p \geq m^{(-1/2 + \varepsilon)p}\right) \\
& \leq & \frac{\bbE |w_{i, 0}|^p}{m^{(-1/2 + \varepsilon)p}} = \frac{\bbE |\Zw|^p}{(\sqrt{m})^p m^{(-1/2 + \varepsilon)p}} = \bbE(|\Zw|^p) m^{-\varepsilon p}~.
\end{IEEEeqnarray*}
By choosing $p = (1+\gP)/\varepsilon$, we can use the union bound to conclude
\begin{IEEEeqnarray*}{+rCl+x*}
\Pinitm\left(\max_i |w_{i, 0}| \geq m^{-1/2 + \varepsilon}\right) & \leq & m \cdot \bbE(|\Zw|^p) m^{-\varepsilon p} \leq O(m^{1-\varepsilon p}) = O(m^{-\gP})~.
\end{IEEEeqnarray*}

\item Similar to (W2).

\item By property (Q1) of \Cref{ass:init}, $\Za$ has a probability density $\dQa$ that is bounded by $\bdQwa$. Thus, for all $\delta \geq 0$, we obtain
\begin{IEEEeqnarray*}{+rCl+x*}
\Pinitm(|a_{i, 0}| \leq \delta) & = & P(|\Za| \leq \delta) = \int_{-\delta}^{\delta} \dQa(x) \diff x \leq 2\delta \cdot \bdQwa~.
\end{IEEEeqnarray*}
Therefore,
\begin{IEEEeqnarray*}{+rCl+x*}
\Pinitm \left(\min_i |a_{i, 0}| \leq m^{-(1+\gP)}\right) & \leq & \sum_{i=1}^m \Pinitm \left(|a_{i, 0}| \leq m^{-(1+\gP)}\right) \\
& \leq & m \cdot 2 m^{-(1+\gP)} \cdot \bdQwa \leq O(m^{-\gP})~.
\end{IEEEeqnarray*}

\item For the next three properties, we need some preparation. Let 
\begin{IEEEeqnarray*}{+rCl+x*}
A_{\sigma, i} & \equalDef & \bbone_{(0, \infty)}(\sigma a_{i, 0}) a_{i, 0} \\
W_{\sigma, i} & \equalDef & \bbone_{(0, \infty)}(\sigma a_{i, 0}) w_{i, 0}~.
\end{IEEEeqnarray*}
Note that the indicator function is applied to $\sigma a_{i, 0}$ in both definitions. Then, $\Sigma_{\sigma, a^2, 0} = \sum_{i \in I_\sigma} a_{i, 0}^2 = \sum_{i=1}^m A_{\sigma, i}^2$ and similarly for $\Sigma_{\sigma, w^2, 0}$ and $\Sigma_{\sigma, wa, 0}$. 
We obtain
\begin{IEEEeqnarray*}{+rCl+x*}
\bbE_{\Pinitm} A_{\sigma, i}^2 & = & \int A_{\sigma, i}^2 \diff \Pinitm = \int \left(\bbone_{(0, \infty)}(\sigma a_{i, 0}) a_{i, 0}\right)^2 \diff \Pinitm \\
& = & \int_{\{\sigma a_{i, 0} > 0\}} a_{i, 0}^2 \diff \Pinitm = \int_{\sigma(0, \infty)} x^2 \dQa(x) \diff x \\
& \stackrel{\text{(Q1)}}{=} & \frac{1}{2} \int_\bbR x^2 \dQa(x) \diff x = \frac{\bbE(\Za^2)}{2} \stackrel{\text{(Q1)}}{=} \frac{\vara}{2}~. \\
\bbE_{\Pinitm} W_{\sigma, i}^2 & = & \bbE_{\Pinitm} \left(\left(\bbone_{(0, \infty)}(\sigma a_{i, 0})\right)^2 w_{i, 0}^2\right) \\
& \stackrel{\text{indep.}}{=} &  \left(\bbE_{\Pinitm} \left(\bbone_{(0, \infty)}(\sigma a_{i, 0})\right)^2\right) \cdot \left(\bbE_{\Pinitm} w_{i, 0}^2\right) \\
& = & \Pinitm(\sigma a_{i, 0} > 0) \cdot \bbE \left(m^{-1/2} \Zw \right)^2 \stackrel{\text{(Q1)}}{=} \frac{1}{2} \cdot \frac{\varw}{m}~. \\
\bbE_{\Pinitm} W_{\sigma, i} A_{\sigma, i} & = & \bbE_{\Pinitm} \left(\bbone_{(0, \infty)}(\sigma a_{i, 0}) w_{i, 0} a_{i, 0}\right) \\
& \stackrel{\text{indep.}}{=} & \big(\underbrace{\bbE_{\Pinitm} w_{i, 0}}_{\stackrel{\text{(Q1)}}{=} 0}\big) \cdot \big(\bbE_{\Pinitm} \bbone_{(0, \infty)}(\sigma a_{i, 0}) a_{i, 0}\big) = 0~.
\end{IEEEeqnarray*}

Now, define
\begin{IEEEeqnarray*}{+rCl+x*}
S_m \equalDef \frac{\Sigma_{\sigma, a^2, 0}}{m} = \frac{1}{m} \sum_{i=1}^m A_{\sigma, i}^2~,
\end{IEEEeqnarray*}
which is an average of $m$ i.i.d.\ variables that are $p$-integrable for every $p > 0$. Then, $\bbE_{\Pinitm} S_m = \bbE_{\Pinitm} A_{\sigma, 1}^2 = \vara/2$ and \Cref{lemma:concentration_inequality} with $\varepsilon = 1/2, \beta = \vara/4$ yields:
\begin{IEEEeqnarray*}{+rCl+x*}
\Pinitm\left(\left|S_m - \frac{\vara}{2}\right| \geq \frac{\vara}{4}\right) \leq O(m^{-\gP})~.
\end{IEEEeqnarray*}

Hence,
\begin{IEEEeqnarray*}{+rCl+x*}
\Pinitm(\Sigma_{\sigma, a^2, 0} \notin [m\vara/4, m\vara]) & = & \Pinitm\left(S_m \notin [\vara/4, \vara]\right) \\
& \leq & \Pinitm(S_m \notin [\vara/4, 3\vara/4]) \\
& \leq & O(m^{-\gP})~.
\end{IEEEeqnarray*}

\item An analogous argument yields $\Pinitm(\Sigma_{\sigma, w^2, 0} \notin [\varw/4, \varw]) \leq O(m^{-\gP})$.

\item Let $S_m \equalDef \frac{1}{m} \sum_{i=1}^m A_{\sigma, i} \cdot \sqrt{m} W_{\sigma, i} = \Sigma_{\sigma, wa, 0}/\sqrt{m}$. Then, $\bbE_{\Pinitm} S_m = 0$ and thus
\begin{IEEEeqnarray*}{+rCl+x*}
\Pinitm(|\Sigma_{\sigma, wa, 0}| \geq m^\varepsilon) & = & \Pinitm(|S_m| \geq m^{\varepsilon-1/2}) \stackrel{\text{\Cref{lemma:concentration_inequality}}}{\leq} O(m^{-\gP})~. & \qedhere
\end{IEEEeqnarray*}
\end{enumerate}
\end{proof}
\end{proposition}

Now, we want to investigate stochastic properties of the data set. In order to show that $\bfM_D^{-1}$ is likely close to $\bfM_{\Pdata}^{-1}$ (both are defined in \Cref{def:main:quantities}), we need the following lemma, which is similar for example to Theorem 2.3.4 in \citet{golub_matrix_1989}:

\begin{lemma} \label{lemma:matrix_norm_inverse} Let $\bfA, \bfB \in \bbR^{m \times m}$ and let $\|\cdot\|$ be a matrix norm on $\bbR^{m \times m}$. If $\bfA$ is invertible and $\|\bfA^{-1}\| \|\bfA-\bfB\| < 1$, then $\bfB$ is invertible with
\begin{IEEEeqnarray*}{+rCl+x*}
\|\bfB^{-1} - \bfA^{-1}\| \leq \|\bfA^{-1}\| \|\bfA-\bfB\| \|\bfB^{-1}\|, \qquad \|\bfB^{-1}\| \leq \frac{\|\bfA^{-1}\|}{1 - \|\bfA^{-1}\| \|\bfA-\bfB\|}~.
\end{IEEEeqnarray*}

\begin{proof}
We have $\bfB = \bfA(\bfI - \bfA^{-1}(\bfA-\bfB))$ and since $\|\bfA^{-1}(\bfA-\bfB)\| \leq \|\bfA^{-1}\| \|\bfA-\bfB\| < 1$, the Neumann series implies that
\begin{IEEEeqnarray*}{+rCl+x*}
(\bfI - \bfA^{-1}(\bfA-\bfB))^{-1} & = & \sum_{k=0}^\infty (\bfA^{-1}(\bfA-\bfB))^k~.
\end{IEEEeqnarray*}
Hence $\bfB$ is invertible with $\bfB^{-1} - \bfA^{-1} = \bfA^{-1}(\bfA-\bfB)\bfB^{-1}$ and
\begin{IEEEeqnarray*}{+rCl+x*}
\bfB^{-1} = (\bfI - \bfA^{-1}(\bfA-\bfB))^{-1} \bfA^{-1} = \sum_{k=0}^\infty (\bfA^{-1}(\bfA-\bfB))^k \bfA^{-1}~,
\end{IEEEeqnarray*}
which yields both bounds using the submultiplicativity of $\|\cdot\|$.
\end{proof}
\end{lemma}

Now, we can show that for large $n$, a sampled data set $D$ likely has characteristics that are close to $\Pdata$. We use the convention $\infty \cdot 0 \equalDef \infty$.

\begin{proposition} \label{prop:sampling:appendix}

Let $\Pdata$ satisfy \Cref{ass:P}, let $\varepsilon, \Cx > 0$, $m \geq 1$ and $\gx, \gamma' \geq 0$.  
If $\eta = \infty$, we further assume that $\Cx$ satisfies $\Pdata((-\Cx, \Cx) \times \bbR) = 0$.
Finally, let $D$ be a data set with $n$ data points $(x_j, y_j)$ sampled independently from $\Pdata$. Then with 
probability $1 - O(n^{-\gamma'} + nm^{-\eta \gx})$ the following hold:
\begin{enumerate}[(D1), wide=0pt, leftmargin=*]

\item $\|\vopt_D - \vopt_{\Pdata}\|_\infty \leq n^{\varepsilon-1/2}$,
\item For $\sigma \in \{\pm 1\}$, $\frac{1}{2} \lmin(\bfM_{\Pdata, \sigma}) \leq \lmin(\bfM_{D, \sigma})$ and $\lmax(\bfM_{D, \sigma}) \leq 2\lmax(\bfM_{\Pdata, \sigma})$,
\item $\underline{x}_D \geq \Cx m^{-\gx}$\, .
\end{enumerate}

\begin{proof}
We use the shorthand $P \equalDef \Pdata$. Again, we bound the probabilities for each property separately.
\begin{enumerate}[(D1), wide=0pt, leftmargin=*]

\item For $\sigma \in \{\pm 1\}$, define
\begin{IEEEeqnarray*}{+rCl+x*}
S_n \equalDef (\bfM_{D, \sigma})_{11} = \frac{1}{n} \sum_{j=1}^n \bbone_{(0, \infty)}(\sigma x_j) x_j^2~.
\end{IEEEeqnarray*}
Then, 
\begin{IEEEeqnarray*}{+rCl+x*}
\bbE_{P^n} S_n = \frac{1}{n} \sum_{j=1}^n \bbE_{D \sim P^n} (\bbone_{(0, \infty)}(\sigma x_j) x_j^2) = \bbE_{(x, y) \sim P} \bbone_{(0, \infty)}(\sigma x) x^2 = (\bfM_{P, \sigma})_{11}~.
\end{IEEEeqnarray*}
Because $P$ is bounded, it has finite moments and we can apply \Cref{lemma:concentration_inequality}: For all $\beta > 0$,
\begin{IEEEeqnarray*}{+rCl+x*}
P^n\left(|(\bfM_{D, \sigma})_{11} - (\bfM_{P, \sigma})_{11}| \geq \beta n^{(\varepsilon-1)/2}\right) \leq O(n^{-\gamma})~.
\end{IEEEeqnarray*}
We can get similar bounds for other entries of $\bfM_{D, \sigma}$ and $\hat{\bfu}^0_{D, \sigma}$. Since
\begin{IEEEeqnarray*}{+rCl+x*}
\bfM_D - \bfM_P & = & \tilde{\bfP} \begin{pmatrix}
\bfM_{D, 1} - \bfM_{P, 1} & 0 \\
0 & \bfM_{D, -1} - \bfM_{P, -1}
\end{pmatrix} \tilde{\bfP}, \quad \hat{\bfu}^0_D = \tilde{\bfP} \begin{pmatrix}
\hat{\bfu}^0_{D, 1} \\
\hat{\bfu}^0_{D, -1}
\end{pmatrix},
\end{IEEEeqnarray*}
and $\|\tilde{\bfP}\|_\infty = 1$, the union bound implies that the following properties hold with probability $1 - O(n^{-\gamma'})$:
\begin{IEEEeqnarray*}{+rCl+x*}
&& \|\bfM_{D, \pm 1} - \bfM_{P, \pm 1}\|_\infty \leq 2\beta n^{(\varepsilon-1)/2}, \quad \|\bfM_D - \bfM_P\|_\infty \leq 2\beta n^{(\varepsilon-1)/2}, \\
&& \|\hat{\bfu}^0_D - \hat{\bfu}^0_P\|_\infty \leq \beta n^{(\varepsilon - 1)/2}~. \IEEEyesnumber \label{eq:small_sampling_deviation}
\end{IEEEeqnarray*}
Now assume that \eqref{eq:small_sampling_deviation} holds. Set $\bfA \equalDef \bfM_P, \bfB \equalDef \bfM_D, \bfa \equalDef \hat{\bfu}^0_P, \bfb \equalDef \hat{\bfu}^0_D$. By condition (P1), $\bfA$ is invertible. Without loss of generality, we can assume $\varepsilon < 1/2$. Then, for $n$ large enough,
\begin{IEEEeqnarray*}{+rCl+x*}
\|\bfA^{-1}\|_\infty \|\bfA - \bfB\|_\infty & \leq & \|\bfA^{-1}\|_\infty 2\beta n^{(\varepsilon-1)/2} \leq \frac{1}{2}~.
\end{IEEEeqnarray*}
Hence, \Cref{lemma:matrix_norm_inverse} implies that $\bfB = \bfM_D$ is invertible with $\|\bfB^{-1}\|_\infty \leq 2\|\bfA^{-1}\|_\infty$ and
\begin{IEEEeqnarray*}{+rCl+x*}
\|\vopt_D - \vopt_P\|_\infty & = & \|\bfB^{-1}\bfb - \bfA^{-1}\bfa\|_\infty \\
& \leq & \|\bfB^{-1}\|_\infty \|\bfb - \bfa\|_\infty + \|\bfB^{-1} - \bfA^{-1}\|_\infty \|\bfa\|_\infty \\
& \leq & \|\bfB^{-1}\|_\infty \|\bfb - \bfa\|_\infty + \|\bfA^{-1}\|_\infty \|\bfA-\bfB\|_\infty \|\bfB^{-1}\|_\infty \|\bfa\|_\infty \\
& \leq & 2 \|\bfA^{-1}\|_\infty \left(\|\bfb - \bfa\|_\infty + \|\bfA^{-1}\|_\infty \|\bfa\|_\infty \|\bfB - \bfA\|_\infty\right) \\
& \stackrel{\eqref{eq:small_sampling_deviation}}{\leq} & 4 \|\bfA^{-1}\|_\infty (1 + \|\bfA^{-1}\|_\infty \|\bfa\|_\infty) \beta n^{(\varepsilon - 1)/2}~.
\end{IEEEeqnarray*}
We can choose $\beta > 0$ such that $4\|\bfA^{-1}\|_\infty (1 + \|\bfA^{-1}\|_\infty \|\bfa\|_\infty) \beta \leq 1$. Therefore,
\begin{IEEEeqnarray*}{+rCl+x*}
\|\vopt_{P, \sigma} - \vopt_{D, \sigma}\|_\infty & \leq & n^{(\varepsilon - 1)/2}
\end{IEEEeqnarray*}
with probability $1 - O(n^{-\gamma'})$.

\item For $\sigma \in \{\pm 1\}$ and each $\bfv \in \bbR^2$, we have 
\begin{IEEEeqnarray*}{+rCl+x*}
|\bfv^\top \bfM_{D, \sigma} \bfv - \bfv^\top \bfM_{P, \sigma} \bfv| \leq \|\bfv\|_2 \|\bfM_{D, \sigma} - \bfM_{P, \sigma} \|_2 \|\bfv\|_2 \leq \sqrt{2} \|\bfM_{D, \sigma} - \bfM_{P, \sigma}\|_\infty \|\bfv\|_2^2
\end{IEEEeqnarray*}
since $\|\cdot\|_2 \leq \sqrt{2} \|\cdot\|_\infty$ on $\bbR^{2 \times 2}$ as mentioned in \Cref{def:matrix_notation}. 
If we choose $\beta > 0$ small enough such that \eqref{eq:small_sampling_deviation} implies $\sqrt{2} \|\bfM_{D, \sigma} - \bfM_{P, \sigma}\|_\infty \leq \lmin(\bfM_{P, \sigma})/2$, it follows that
\begin{IEEEeqnarray*}{+rCl+x*}
\lmin(\bfM_{D, \sigma}) & = & \inf_{\|\bfv\|_2 = 1} \bfv^\top \bfM_{D, \sigma} \bfv \geq \inf_{\|\bfv\|_2 = 1} \bfv^\top \bfM_{P, \sigma} \bfv - |\bfv^\top \bfM_{P, \sigma} \bfv - \bfv^\top \bfM_{D, \sigma} \bfv| \\
& \geq & \lmin(\bfM_{P, \sigma}) - \sqrt{2} \|\bfM_{D, \sigma} - \bfM_{P, \sigma}\|_\infty \geq \lmin(\bfM_{P, \sigma})/2~.
\end{IEEEeqnarray*}
Since \eqref{eq:small_sampling_deviation} holds with probability $1 - O(n^{-\gamma'})$, we have $\lmin(\bfM_{D, \sigma}) \geq \lmin(\bfM_{P, \sigma})/2$ with probability $1 - O(n^{-\gamma})$. The probability for $\lmax(\bfM_{D, \sigma}) \leq 2\lmax(\bfM_{P, \sigma})$ can be bounded similarly.

\item In the case $\eta = \infty$ and $P_X((-\Cx, \Cx)) = 0$, this obviously holds with probability one since $\Cx m^{-\gx} \leq \Cx \leq \udl{x}_D$ almost surely. Otherwise, using property (P2) from \Cref{ass:P} and the union bound yields
\begin{IEEEeqnarray*}{+rCl+x*}
P^n(\udl{x}_D < \Cx m^{-\gx}) & \leq & \sum_{j=1}^n P^n(|x_j| < \Cx m^{-\gx}) \\
& \stackrel{\text{(P2)}}{\leq} & n \cdot O\left((\Cx m^{-\gx})^\eta\right) = O(nm^{-\eta \gx})~. & \qedhere
\end{IEEEeqnarray*}
\end{enumerate}
\end{proof}
\end{proposition}

\section{Reference Dynamics} \label{sec:reference_dynamics}

In this section, we define a matrix $\Aref \approx \bfA_0$ and study the asymptotic behavior of $h\sum_{k=0}^\infty \|\ovl{\bfv}_k\|$ when $\ovl{\bfv}_k$ satisfies a reference system 
\begin{IEEEeqnarray*}{+rCl+x*}
\ovl{\bfv}_{k+1} & = & \ovl{\bfv}_k - h\Aref \bfM_D \ovl{\bfv}_k,
\end{IEEEeqnarray*}
instead of the actual dynamics $\ovl{\bfv}_{k+1} = \ovl{\bfv}_k - h\bfA_k \bfM_D \ovl{\bfv}_k$. The solution of the reference system is simply $\ovl{\bfv}_k = (\bfI_4 - h\Aref \bfM_D)^k \ovl{\bfv}_0$.

\begin{definition} \label{def:Aref}
Let $\Aref \equalDef \bfB(\Gw_0 + \Gab_0)\bfB + \bfC$, where $\bfB, \Gw, \Gab, \bfC$ are defined in \Cref{def:derived_quantities}. Moreover, define the symmetric matrix
\begin{IEEEeqnarray*}{+rCl+x*}
\bfH & \equalDef & \bfM_D^{1/2} \Aref \bfM_D^{1/2} = \bfM_D^{1/2} (\Aref \bfM_D) \bfM_D^{-1/2}~. & \qedhere
\end{IEEEeqnarray*}
\end{definition}

\begin{assumption} \label{assumption:training_dynamics}
Assume that $\gpsi, \gx, \gP \geq 0$ with $\gpsi + \gx + \gP < 1/2$. Moreover, we only consider initial vectors $W_0$ which satisfy the conditions (W1) -- (W7) in \Cref{prop:prob_W}. Similar to \Cref{thm:main:general}, we further assume that
\begin{IEEEeqnarray*}{+rCl+x*}
\CM^{-1} & \leq & \lmin(\bfM_D) \leq \lmax(\bfM_D) \leq \CM \\
\psi_{D, p} & = & O(1) \\
\psi_{D, q} & = & O(m^{\gpsi - 1}) \\
h & \leq & \lmax(\bfH)^{-1} \\
0 & < & \varepsilon < \frac{1/2 - (\gpsi + \gx + \gP)}{3}~,
\end{IEEEeqnarray*}
where $\CM > 0$ is a constant and $\bfH$ depends on $D$ and $W_0$. Note that since $\eig(\bfM_D) = \eig(\bfM_{D, 1}) \cup \eig(\bfM_{D, -1})$ by construction of $\bfM_D$, the first condition is equivalent to 
\begin{IEEEeqnarray*}{+rCl+x*}
\CM^{-1} & \leq & \lmin(\bfM_{D, \sigma}) \leq \lmax(\bfM_{D, \sigma}) \leq \CM \text{ for } \sigma = \pm 1~. & \qedhere
\end{IEEEeqnarray*}
\end{assumption}

\begin{lemma} \label{lemma:eig_Aref}
Let \Cref{assumption:training_dynamics} be satisfied. The matrix $\Aref$ is of the form
\begin{IEEEeqnarray*}{+rCl+x*}
\Aref = \begin{pmatrix}
\Aref_1 \\
& \Aref_2
\end{pmatrix}
\end{IEEEeqnarray*}
with $0 \matless \Aref_1, \Aref_2 \in \bbR^{2 \times 2}$ and
\begin{IEEEeqnarray*}{+rCl+x*}
\lmin(\Aref_1) = \Theta(m), \quad \lmax(\Aref_1) = \Theta(m), \quad \lmin(\Aref_2) = \Theta(1), \quad \lmax(\Aref_2) = \Theta(1)~.
\end{IEEEeqnarray*}

\begin{proof}
Since $b_{i, 0} = 0$ by initialization property (W1) in \Cref{prop:prob_W}, we have $\Sigma_{\sigma, ab, 0} = \Sigma_{\sigma, b^2, 0} = 0$. Since the distributions of $\Za, \Zw$ have densities by (Q1) in \Cref{ass:init}, we have $\vara, \varw > 0$. This yields
\begin{IEEEeqnarray*}{+rCl+x*}
\Gw_{\sigma, 0} + \Gab_{\sigma, 0} = \begin{pmatrix}
\Sigma_{\sigma, w^2, 0} + \Sigma_{\sigma, a^2, 0} \\
& \Sigma_{\sigma, w^2, 0}
\end{pmatrix} \stackrel{\text{(W5), (W6)}}{=} \begin{pmatrix}
\Theta(m) \\
& \Theta(1)
\end{pmatrix}~.
\end{IEEEeqnarray*}
Hence,
\begin{IEEEeqnarray*}{+rCl+x*}
\Gw_0 + \Gab_0 & = & \tilde{\bfP}(\tGw_0 + \tGab_0)\tilde{\bfP} = \tilde{\bfP} \begin{pmatrix}
\Gw_{1, 0} + \Gab_{1, 0} \\
& \Gw_{-1, 0} + \Gab_{-1, 0}
\end{pmatrix} \tilde{\bfP} \\
& = & \tilde{\bfP}\begin{pmatrix}
\Theta(m) \\
& \Theta(1) \\
&& \Theta(m)  \\
&&& \Theta(1)
\end{pmatrix} \tilde{\bfP} = \begin{pmatrix}
\Theta(m) \\
& \Theta(m) \\
&& \Theta(1) \\
&&& \Theta(1)
\end{pmatrix} \\
& \defEqual & \begin{pmatrix}
\bfG_1 \\
& \bfG_2
\end{pmatrix}~.
\end{IEEEeqnarray*}
We have seen in \Cref{def:derived_quantities} that
\begin{IEEEeqnarray*}{+rCl+x*}
\bfB & = & \begin{pmatrix}
\hat{\bfB} \\
& \hat{\bfB}
\end{pmatrix}, \quad \hat{\bfB} = \begin{pmatrix}
1 & \alpha \\
\alpha & 1
\end{pmatrix}, \quad \bfC = \begin{pmatrix}
0 & 0 &  \\
0 & 0 &  \\
& & 1 & 1 \\
& & 1 & 1
\end{pmatrix} \defEqual \begin{pmatrix}
\bfzero & \\
& \hat{\bfC}
\end{pmatrix}~.
\end{IEEEeqnarray*}
Using the previous results, we obtain
\begin{IEEEeqnarray*}{+rCl+x*}
\Aref = \begin{pmatrix}
\hat{\bfB} \bfG_1 \hat{\bfB} \\
& \hat{\bfB} \bfG_2 \hat{\bfB} + \hat{\bfC}
\end{pmatrix} \defEqual \begin{pmatrix}
\Aref_1 \\
& \Aref_2
\end{pmatrix}~.
\end{IEEEeqnarray*}
The matrix $\hat{\bfB}$ is fixed and invertible since $|\alpha| \neq 1$. Moreover, $\eig(\hat{\bfC}) = \{0, 2\}$. This yields
\begin{IEEEeqnarray*}{+rCl+x*}
\eig(\Aref_1) & = & \eig(\hat{\bfB} \bfG_1 \hat{\bfB}) = \Theta(m) \\
\eig(\Aref_2) & = & \eig(\hat{\bfB} \bfG_2 \hat{\bfB} + \hat{\bfC}) = \Theta(1)~. & \qedhere
\end{IEEEeqnarray*}
\end{proof}
\end{lemma}

\begin{proposition} \label{prop:ref_sum} 
Let \Cref{assumption:training_dynamics} be satisfied. We have
\begin{IEEEeqnarray*}{+rCl+x*}
h\sum_{k=0}^\infty \|(\bfI_4 - h\Aref \bfM)^k \ovl{\bfv}_0\|_\infty & = & O(m^{\varepsilon + \gpsi - 1}) \\
h\sum_{k=0}^\infty \|(\bfI_4 - h\Aref \bfM)^k\|_\infty & = & O(1)~.
\end{IEEEeqnarray*}

\begin{proof}
We divide the proof in multiple steps:
\begin{enumerate}[(1), wide=0pt, leftmargin=*]

\item \emph{Investigate the initial vector}:

By definition, we have $\ovl{\bfv}_0 = \bfv_0 - \vopt_D$. Therefore,
\begin{IEEEeqnarray*}{+rCl+x*}
|\ovl{\bfv}_0| & \leq & |\bfv_0| + |\vopt_D| \\
& \leq & \begin{pmatrix}
|p_{1, 0}| \\ |p_{-1, 0}| \\ |q_{1, 0}| \\ |q_{-1, 0}|
\end{pmatrix} + \begin{pmatrix}
\psi_{D, p} \\
\psi_{D, p} \\
\psi_{D, q} \\
\psi_{D, q}
\end{pmatrix} = \begin{pmatrix}
|\Sigma_{1, wa, 0} + \alpha \Sigma_{-1, wa, 0}| \\ |\Sigma_{-1, wa, 0} + \alpha \Sigma_{1, wa, 0}| \\ |\Sigma_{1, wb, 0} + \alpha \Sigma_{-1, wb, 0}| \\ |\Sigma_{-1, wb, 0} + \alpha \Sigma_{1, wb, 0}|
\end{pmatrix} + \begin{pmatrix}
\psi_{D, p} \\
\psi_{D, p} \\
\psi_{D, q} \\
\psi_{D, q}
\end{pmatrix} \\
& \stackrel{\text{(W1), (W7), \ref{assumption:training_dynamics}}}{\leq} & \begin{pmatrix}
O(m^\varepsilon) \\ O(m^\varepsilon) \\ 0 \\ 0
\end{pmatrix} + \begin{pmatrix}
O(1) \\ O(1) \\ O(m^{\gamma_\psi - 1}) \\ O(m^{\gamma_\psi - 1})
\end{pmatrix} \leq \begin{pmatrix}
O(m^\varepsilon) \\ O(m^\varepsilon) \\ O(m^{\gpsi - 1}) \\ O(m^{\gpsi - 1})
\end{pmatrix}~.
\end{IEEEeqnarray*}
Thus, we can group
\begin{IEEEeqnarray*}{+rCl+x*}
\ovl{\bfv}_0 = \begin{pmatrix}
\ovl{\bfv}_{0, 1} \\
\ovl{\bfv}_{0, 2}
\end{pmatrix}
\end{IEEEeqnarray*}
with $\ovl{\bfv}_{0, 1}, \ovl{\bfv}_{0, 2} \in \bbR^2$ and $\|\ovl{\bfv}_{0, 1}\|_\infty \leq O(m^\varepsilon), \|\ovl{\bfv}_{0, 2}\|_\infty \leq O(m^{\gpsi - 1})$.

\item \emph{Diagonalization yields a simple bound:}

The matrix $\Aref \bfM$ is similar to the symmetric matrix 
\begin{IEEEeqnarray*}{+rCl+x*}
\bfH & \equalDef & \bfM^{1/2}\Aref \bfM^{1/2} = \bfM^{1/2}(\Aref \bfM) \bfM^{-1/2} \matgr 0~.
\end{IEEEeqnarray*}
The matrix $\bfH$ can thus be orthogonally diagonalized as $\bfH = \bfU\bfD\bfU^\top$ with $\bfU$ orthogonal and $\bfD$ diagonal such that $\bfD$ contains the eigenvalues of $\bfH$ in descending order. Then, $\bfI_4-h\bfD$ only contains non-negative entries due to the condition $h \leq \lmax(\bfH)^{-1}$ with its maximal entry being $1 - h\lmin(\bfH)$. Thus, $\|(\bfI_4 - h\bfD)^k\|_2 = (1 - h\lmin(\bfH))^k$. By applying $(\bfI_4 - h\Aref \bfM)\bfM^{-1/2} = \bfM^{-1/2} - h\Aref \bfM^{1/2} = \bfM^{-1/2}(\bfI_4 - h\bfH)$ inductively, we find $(\bfI_4 - h\Aref \bfM)^k \bfM^{-1/2} = \bfM^{-1/2}(\bfI_4 - h\bfH)^k$. We can now compute
\begin{IEEEeqnarray*}{+rCl+x*}
h\sum_{k=0}^\infty \|(\bfI_4 - h\Aref \bfM)^k\|_2 & = & h\sum_{k=0}^\infty \|\bfM^{-1/2} (\bfI_4 - h\bfH)^k \bfM^{1/2}\|_2 \\
& = & h\sum_{k=0}^\infty \|\bfM^{-1/2} \bfU (\bfI_4 - h\bfD)^k \bfU^\top \bfM^{1/2}\|_2 \\
& \leq & \|\bfM^{-1/2}\|_2 \|\bfM^{1/2}\|_2 \cdot h\sum_{k=0}^\infty \|(\bfI_4 - h\bfD)^k\|_2 \\
& = & \cond(\bfM^{1/2}) h\sum_{k=0}^\infty (1 - h\lmin(\bfH))^k \\
& = & \sqrt{\cond(\bfM)} \frac{h}{1 - (1 - h\lmin(\bfH))} \\
& = & \frac{\sqrt{\cond(\bfM)}}{\lmin(\bfH)} \leq O(1)~, \IEEEyesnumber \label{eq:matrix_power_sum}
\end{IEEEeqnarray*}
where $\lmin(\bfH) \geq \Omega(1)$ since for $\bfv \in \bbR^4$, we have
\begin{IEEEeqnarray*}{+rCl+x*}
\bfv^\top \bfH \bfv & = & (\bfM^{1/2} \bfv)^\top \Aref (\bfM^{1/2} \bfv) \geq \lmin(\Aref) \bfv^\top \bfM \bfv \geq \lmin(\Aref) \lmin(\bfM) \bfv^\top \bfv~, \IEEEyesnumber \label{eq:lminH}
\end{IEEEeqnarray*}
where $\lmin(\Aref) \lmin(\bfM) = \Theta(1)$ by \Cref{assumption:training_dynamics} and \Cref{lemma:eig_Aref}.

\item \emph{$\Aref \bfM$ has $2$ \quot{large} eigenvalues:}

Let
\begin{IEEEeqnarray*}{+rCl+x*}
\bfM & = & \begin{pmatrix}
\bfM_{11} & \bfM_{12} \\
\bfM_{12}^\top & \bfM_{22}
\end{pmatrix}, \qquad \bfM^{1/2} = \begin{pmatrix}
\tilde{\bfM}_{11} & \tilde{\bfM}_{12} \\
\tilde{\bfM}_{12}^\top & \tilde{\bfM}_{22}
\end{pmatrix}
\end{IEEEeqnarray*}
be the block decompositions of $\bfM$ and $\bfM^{1/2}$ into $2 \times 2$ blocks. Then, 
\begin{IEEEeqnarray*}{+rCl+x*}
\bfM^{1/2} \Aref \bfM^{1/2} & = & \begin{pmatrix}
\tilde{\bfM}_{11} \Aref_1 \tilde{\bfM}_{11} + \tilde{\bfM}_{12} \Aref_2 \tilde{\bfM}_{12}^\top & * \\
* & *
\end{pmatrix}
\end{IEEEeqnarray*}
and by Cauchy's interlacing theorem (cf.\ e.g.\ Corollary III.1.5 in \citep{bhatia_matrix_2013}), the second largest eigenvalue $\lambda_2(\bfH)$ of $\bfH$ satisfies
\begin{IEEEeqnarray*}{+rCl+x*}
\lambda_2(\bfH) & \geq & \lambda_2(\tilde{\bfM}_{11} \Aref_1 \tilde{\bfM}_{11} + \tilde{\bfM}_{12} \Aref_2 \tilde{\bfM}_{12}^\top) = \lmin(\tilde{\bfM}_{11} \Aref_1 \tilde{\bfM}_{11} + \tilde{\bfM}_{12} \Aref_2 \tilde{\bfM}_{12}^\top) \\
& \geq & \lmin(\tilde{\bfM}_{11} \Aref_1 \tilde{\bfM}_{11}) \geq \lmin(\Aref_1) \lmin(\tilde{\bfM}_{11})^2 \geq \lmin(\Aref_1) \lmin(\bfM^{1/2})^2 \\
& = & \lmin(\Aref_1) \lmin(\bfM) \geq \Theta(m)~. \IEEEyesnumber \label{eq:middle_eigenvalue}
\end{IEEEeqnarray*}

\item \emph{Lower components of eigenvectors to large eigenvalues are small:}

Let $\bfw = (\bfw_1, \bfw_2)^\top$ be an eigenvector of $\Aref \bfM$ with eigenvalue $\lambda \geq \lambda_2(\bfH) \geq \Theta(m)$. The lower part of the identity $\lambda \bfw = \Aref\bfM\bfw$ reads as
\begin{IEEEeqnarray*}{+rCl+x*}
\lambda \bfw_2 & = & \Aref_2\bfM_{12}^\top \bfw_1 + \Aref_2 \bfM_{22} \bfw_2~,
\end{IEEEeqnarray*}
which yields
\begin{IEEEeqnarray*}{+rCl+x*}
\Theta(m) \|\bfw_2\|_2 & \leq & \lambda \|\bfw_2\|_2 \leq \|\Aref_2\|_2 \|\bfM_{12}^\top\|_2 \|\bfw_1\|_2 + \|\Aref_2\|_2 \|\bfM_{22}\|_2 \|\bfw_2\|_2 \\
& \leq & \Theta(1) \|\bfw_1\|_2 + \Theta(1) \|\bfw_2\|_2
\end{IEEEeqnarray*}
and hence (for large $m$)
\begin{IEEEeqnarray*}{+rCl+x*}
\|\bfw_2\|_2 & \leq & \frac{\Theta(1)}{\Theta(m) - \Theta(1)} \|\bfw_1\|_2 \leq O(m^{-1}) \|\bfw_1\|_2~. \IEEEyesnumber \label{eq:eigenvector:beta}
\end{IEEEeqnarray*}

\item \emph{The first two eigenvectors of $\Aref \bfM$ are \quot{well-conditioned}:}

Let
\begin{IEEEeqnarray*}{+rCl+x*}
\bfU & = & \begin{pmatrix}
\bfU_1 & \bfU_2
\end{pmatrix} = \begin{pmatrix}
\bfU_{11} & \bfU_{12} \\
\bfU_{21} & \bfU_{22}
\end{pmatrix}, \quad \bfF = \begin{pmatrix}
\bfF_1 & \bfF_2
\end{pmatrix} \equalDef \bfU_1^\top \bfM^{1/2}, \\
\bfW & = & \begin{pmatrix}
\bfW_1 \\ \bfW_2
\end{pmatrix} \equalDef \bfM^{-1/2} \bfU_1~.
\end{IEEEeqnarray*} 
The columns of $\bfW$ are the eigenvectors of $\Aref\bfM$ to the $2$ largest eigenvalues:
\begin{IEEEeqnarray*}{+rCl+x*}
\Aref\bfM\bfW & = & \bfM^{-1/2}\bfM^{1/2}\Aref\bfM^{1/2}\bfU_1 = \bfM^{-1/2}\bfU\bfD\bfU^\top \bfU_1 = \bfM^{-1/2} \bfU \bfD \begin{pmatrix}
\bfI_2 \\ \bfzero
\end{pmatrix} \\
& = & \bfM^{-1/2} \bfU_1 \bfD_1 = \bfW \bfD_1~, \IEEEyesnumber \label{eq:sum_AM:eigenvectors}
\end{IEEEeqnarray*}
where $\bfD_1$ is the upper left $2 \times 2$ block of $\bfD$. Thus,
\begin{IEEEeqnarray*}{+rCl+x*}
\|\bfF\|_2 & \leq & \|\bfU_1^\top\|_2 \|\bfM^{1/2}\|_2 = 1 \cdot \lmax(\bfM^{1/2}) = \Theta(1) \\
\|\bfW\|_2 & \leq & \|\bfM^{-1/2}\|_2 \|\bfU_1\|_2 = \lmax(\bfM^{-1/2}) \cdot 1 = \Theta(1) \\
\|\bfW_2\|_2 & \leq & \|\bfW_2\|_F \stackrel{\text{\eqref{eq:eigenvector:beta}}}{\leq} O(m^{-1}) \|\bfW_1\|_F \leq O(m^{-1}) \|\bfW\|_2 \leq O(m^{-1})~.
\end{IEEEeqnarray*}
We want to show that $\bfW_1^{-1}$ exists and $\|\bfW_1^{-1}\|_2$ is sufficiently small. Observe that $\bfI_2 = \bfU_1^\top \bfU_1 = \bfF\bfW = \bfF_1\bfW_1 + \bfF_2\bfW_2$ and
\begin{IEEEeqnarray*}{+rCl+x*}
\|\bfF_2\bfW_2\|_2 & \leq & \|\bfF_2\|_2 \|\bfW_2\|_2 \leq O(m^{-1}) \leq \frac{1}{2}
\end{IEEEeqnarray*}
for large $m$. Hence, $\bfF_1\bfW_1 = \bfI_2 - \bfF_2\bfW_2$ is invertible with
\begin{IEEEeqnarray*}{+rCl+x*}
(\bfF_1\bfW_1)^{-1} = \sum_{k=0}^\infty (\bfF_2\bfW_2)^k, \qquad \|(\bfF_1\bfW_1)^{-1}\|_2 \leq \sum_{k=0}^\infty \|\bfF_2\bfW_2\|_2^k \leq 2~.
\end{IEEEeqnarray*}
Since $\bfF_1\bfW_1$ has full rank, $\bfW_1$ and $\bfF_1$ must also have full rank. Hence, $(\bfF_1\bfW_1)^{-1} = \bfW_1^{-1}\bfF_1^{-1}$ and 
\begin{IEEEeqnarray*}{+rCl+x*}
\|\bfW_1^{-1}\|_2 \leq \|(\bfF_1\bfW_1)^{-1}\|_2 \|\bfF_1\|_2 \leq O(1)~.
\end{IEEEeqnarray*}

\item \emph{Bound the sum for a \quot{similar} initial vector:}

Note that for $\tilde{\bfv}_2 \equalDef \bfW_2\bfW_1^{-1}\ovl{\bfv}_{0, 1}$, we have
\begin{IEEEeqnarray*}{+rCl+x*}
\bfW\bfW_1^{-1}\ovl{\bfv}_{0, 1} & = & \begin{pmatrix}
\bfI_2 \\
\bfW_2\bfW_1^{-1}
\end{pmatrix} \ovl{\bfv}_{0, 1} = \begin{pmatrix}
\ovl{\bfv}_{0, 1} \\ \tilde{\bfv}_2
\end{pmatrix} \IEEEyesnumber \label{eq:proj_v1}
\end{IEEEeqnarray*}
and $\tilde{\bfv}_2$ is \quot{small}:
\begin{IEEEeqnarray*}{+rCl+x*}
\|\tilde{\bfv}_2\|_2 \leq \|\bfW_2\|_2 \|\bfW_1^{-1}\|_2 \|\ovl{\bfv}_{0, 1}\|_2 \leq O(m^{-1}) O(1) O(m^\varepsilon) = O(m^{\varepsilon - 1})~.
\end{IEEEeqnarray*}

By Eq.~\eqref{eq:sum_AM:eigenvectors}, we have $\Aref \bfM \bfW = \bfW\bfD_1$, where $\bfD_1$ is the upper left $2 \times 2$ block of $\bfD$. Therefore,
\begin{IEEEeqnarray*}{+rCl+x*}
h\sum_{k=0}^\infty \|(\bfI_4 - h\Aref\bfM)^k\bfW\bfW_1^{-1} \ovl{\bfv}_{0, 1}\|_2 & = & h\sum_{k=0}^\infty \|\bfW(\bfI_2 - h\bfD_1)^k \bfW_1^{-1} \ovl{\bfv}_{0, 1}\|_2 \\
& \leq & \|\bfW\|_2 \|\bfW_1^{-1}\|_2 \|\ovl{\bfv}_{0, 1}\|_2 \cdot h\sum_{k=0}^\infty \|(\bfI_2 - h\bfD_1)^k\|_2~,
\end{IEEEeqnarray*}
where
\begin{IEEEeqnarray*}{+rCl+x*}
\|\bfW\|_2 \|\bfW_1^{-1}\|_2 \|\ovl{\bfv}_{0, 1}\|_2 & \leq & O(1) O(1) O(m^\varepsilon) = O(m^\varepsilon)
\end{IEEEeqnarray*}
and we can compute the remaining sum similar to step (2):
\begin{IEEEeqnarray*}{+rCl+x*}
h\sum_{k=0}^\infty \|(\bfI_2 - h\bfD_1)^k\|_2 & = & h\sum_{k=0}^\infty (1-h\lambda_2(\bfH))^k \leq \frac{h}{1-(1-h\lambda_2(\bfH))} \stackrel{\eqref{eq:middle_eigenvalue}}{\leq} O(m^{-1})~.
\end{IEEEeqnarray*}

\item \emph{Bound the original sum:}

Using $\ovl{\bfv}_0 = \bfW\bfW_1^{-1}\ovl{\bfv}_{0, 1} + \begin{pmatrix}
0 \\ \ovl{\bfv}_{0, 2} - \tilde{\bfv}_2
\end{pmatrix}$, we obtain
\begin{IEEEeqnarray*}{+rCl+x*}
&& h\sum_{k=0}^\infty \|(\bfI_4 - h\Aref\bfM)^k \ovl{\bfv}_0\|_2 \\
& \stackrel{\eqref{eq:proj_v1}}{\leq} & h\sum_{k=0}^\infty \|(\bfI_4 - h\Aref\bfM)^k \bfW\bfW_1^{-1}\ovl{\bfv}_{0, 1}\|_2 \\
&& ~+~ h\sum_{k=0}^\infty \left\|(\bfI_4-h\Aref\bfM)^k \begin{pmatrix}
0 \\
\ovl{\bfv}_{0, 2} - \tilde{\bfv}_2
\end{pmatrix}\right\|_2 \\
& \leq & h\sum_{k=0}^\infty \|(\bfI_4 - h\Aref\bfM)^k \bfW\bfW_1^{-1}\ovl{\bfv}_{0, 1}\|_2 \\
&& ~+~ h\sum_{k=0}^\infty \|(\bfI_4-h\Aref\bfM)^k\|_2 \cdot \left(\|\ovl{\bfv}_{0, 2}\|_2 + \|\tilde{\bfv}_2\|_2\right) \\
& \stackrel{\text{\eqref{eq:matrix_power_sum}, Step (5)}}{\leq} & O(m^{-1}) O(m^{\varepsilon}) + O(1) \left(O(m^{\gpsi - 1}) + O(m^{\varepsilon-1})\right) \\
& \leq & O(m^{\varepsilon + \gpsi - 1})~. & \qedhere
\end{IEEEeqnarray*}
\end{enumerate}
\end{proof}
\end{proposition}

\section{Training Dynamics} \label{sec:training_dynamics}

In this section, we investigate how much the weights $W_k$ change during training, which allows us to prove \Cref{thm:main:general} at the end of this section. To this end, we first define important terms.

\begin{definition} \label{def:kappa_Q_1}
For any sequence $(z_k)_{k \in \bbN_0}$, define
\begin{IEEEeqnarray*}{+rCl+x*}
\Delta_k z \equalDef \max_{0 \leq l \leq k} |z_l - z_0|~,
\end{IEEEeqnarray*}
where the supremum should be taken element-wise if $z$ is a vector or a matrix. Moreover, let
\begin{IEEEeqnarray*}{+rClrCl+x*}
\kappa_{u, k} & \equalDef & h\sum_{l=0}^k \|\bfu_l\|_\infty, \quad & \tilde{\bfQ} & \equalDef & \begin{pmatrix}
0 & 0 & 1 \\
0 & 0 & 1 \\
1 & 1 & 0
\end{pmatrix}, \quad \mathbf{1}_3 \equalDef \begin{pmatrix}
1 \\ 1 \\ 1
\end{pmatrix}, \quad \mathbf{1}_{3 \times 3} \equalDef \begin{pmatrix}
1 & 1 & 1 \\
1 & 1 & 1 \\
1 & 1 & 1
\end{pmatrix}~. & \qedhere
\end{IEEEeqnarray*}
\end{definition}

Now, we can state a general result, which resembles a first-order Taylor approximation:\footnote{In the \quot{first-order term}, the matrices are still sparse. \quot{Higher-order} approximations are not useful for our purpose.}

\begin{proposition} \label{prop:difference_bound}
Let $k \in \bbN_0$, $\sigma \in \{\pm 1\}$ and $i \in I_\sigma$. Then, with $|\cdot|$ and $\leq$ understood component-wise,
\begin{IEEEeqnarray*}{+rCl+x*}
\Delta_k \bftheta_i & \leq & \kappa_{u, k} \tilde{\bfQ} |\bftheta_{i, 0}| + 2 \kappa_{u, k}^2 e^{2\kappa_{u, k}} \|\bftheta_{i, 0}\|_\infty \mathbf{1}_3 \\
\Delta_k \bfSigma_\sigma & \leq & \kappa_{u, k}(\tilde{\bfQ} |\bfSigma_{\sigma, 0}| + |\bfSigma_{\sigma, 0}| \tilde{\bfQ}) + 8\kappa_{u, k}^2 e^{4\kappa_{u, k}} \|\bfSigma_{\sigma, 0}\|_\infty \mathbf{1}_{3 \times 3}~.
\end{IEEEeqnarray*}

\begin{proof}
The inequality
\begin{IEEEeqnarray*}{+rCl+x*}
\|(\bfA + \bfI)^2 - 2\bfA - \bfI\|_\infty & \leq & (\|\bfA\|_\infty + \|\bfI\|_\infty)^2 - 2\|\bfA\|_\infty - \|\bfI\|_\infty
\end{IEEEeqnarray*}
for arbitrary matrices $\bfA$ looks like an incorrect application of the triangle inequality due to the minus signs. However, it is correct since the subtracted terms exactly match terms in the expansion of the first term (since $\|\bfI\|_\infty = 1$):
\begin{IEEEeqnarray*}{+rCl+x*}
\|(\bfA + \bfI)^2 - 2\bfA - \bfI\|_\infty & = & \|\bfA^2 + 2\bfA + \bfI - 2\bfA - \bfI\|_\infty = \|\bfA^2\|_\infty \\
& \leq & \|\bfA\|_\infty^2 = \|\bfA\|_\infty^2 + 2\|\bfA\|_\infty + \|\bfI\|_\infty - 2\|\bfA\|_\infty - \|\bfI\|_\infty \\
& = & (\|\bfA\|_\infty + \|\bfI\|_\infty)^2 - 2\|\bfA\|_\infty - \|\bfI\|_\infty~.
\end{IEEEeqnarray*}
We can apply the same trick to obtain bounds on $|\bftheta_{i, k} - \bftheta_{i, 0}|$ and $|\bfSigma_{\sigma, k} - \bfSigma_{\sigma, 0}|$:\footnote{The bound on $\Delta_k \bftheta_i$ and $\Delta_k \bfSigma_\sigma$ then follows since the bound is increasing in $k$.} Define
\begin{IEEEeqnarray*}{+rCl+x*}
\tilde{\bfQ}_k & \equalDef & h\sum_{l=0}^k \bfQ_{\sigma, l}, \quad \tilde{s}_k \equalDef h\sum_{l=0}^k \|\bfQ_{\sigma, l}\|_\infty~.
\end{IEEEeqnarray*}
Since
\begin{IEEEeqnarray*}{+rCl+x*}
\bftheta_{i, k} & = & (\bfI_3 + h\bfQ_{\sigma, k-1}) \cdot \hdots \cdot (\bfI_3 + h\bfQ_{\sigma, 0}) \bftheta_{i, 0} \\
\Sigma_{\sigma, k} & = & (\bfI_3 + h\bfQ_{\sigma, k-1}) \cdot \hdots \cdot (\bfI_3 + h\bfQ_{\sigma, 0}) \bfSigma_{\sigma, 0} (\bfI_3 + h\bfQ_{\sigma, 0}) \cdot \hdots \cdot (\bfI_3 + h\bfQ_{\sigma, k-1}),
\end{IEEEeqnarray*}
we find with $1 + x \leq e^x$:
\begin{IEEEeqnarray*}{+rCl+x*}
&& \|\bftheta_{i, k} - \tilde{\bfQ}_{k-1} \bftheta_{i, 0} - \bftheta_{i, 0}\|_\infty \\
& \leq & (1 + h\|\bfQ_{\sigma, k-1}\|_\infty) \cdot \hdots \cdot (1 + h\|\bfQ_{\sigma, 0}\|_\infty) \|\bftheta_{i, 0}\|_\infty - \tilde{s}_{k-1} \|\bftheta_{i, 0}\|_\infty - \|\bftheta_{i, 0}\|_\infty \\
& \leq & (e^{\tilde{s}_{k-1}} - \tilde{s}_{k-1} - 1) \|\bftheta_{i, 0}\|_\infty
\end{IEEEeqnarray*}
and similarly
\begin{IEEEeqnarray*}{+rCl+x*}
&& \|\bfSigma_{\sigma, k} - \tilde{\bfQ}_{k-1} \bfSigma_{\sigma, 0} - \bfSigma_{\sigma, 0} \tilde{\bfQ}_{k-1} - \bfSigma_{\sigma, 0}\|_\infty \\
& \leq & (1 + h\|\bfQ_{\sigma, k-1}\|_\infty) \cdots (1 + h\|\bfQ_{\sigma, 0}\|_\infty) \|\bfSigma_{\sigma, 0}\|_\infty (1 + h\|\bfQ_{\sigma, 0}\|_\infty) \cdots (1 + h\|\bfQ_{\sigma, k-1}\|_\infty) \\
&& ~-~ (\tilde{s}_{k-1} \|\bfSigma_{\sigma, 0}\|_\infty + \|\bfSigma_{\sigma, 0}\|_\infty \tilde{s}_{k-1} + 1) \\
& \leq & (e^{2\tilde{s}_{k-1}} - 2\tilde{s}_{k-1} - 1) \|\bfSigma_{\sigma, 0}\|_\infty~.
\end{IEEEeqnarray*}
Observe that
\begin{IEEEeqnarray*}{+rCl+x*}
e^x - x - 1 = \sum_{k=2}^\infty \frac{x^k}{k!} = x^2 \sum_{k=0}^\infty \frac{x^k}{(k+2)!} \stackrel{(k+2)! \geq 2k!}{\leq} \frac{1}{2} x^2 \sum_{k=0}^\infty \frac{x^k}{k!} = \frac{1}{2} x^2 e^x~.
\end{IEEEeqnarray*}
Obviously, 
\begin{IEEEeqnarray*}{+rCl+x*}
|\tilde{\bfQ}_k| & \leq & h\sum_{l=0}^k |\bfQ_{\sigma, l}| = h\begin{pmatrix}
0 & 0 & \sum_{l=0}^k |r_{\sigma, l}| \\
0 & 0 & \sum_{l=0}^k |s_{\sigma, l}| \\
\sum_{l=0}^k |r_{\sigma, l}| & \sum_{l=0}^k |s_{\sigma, l}|
\end{pmatrix} \leq \kappa_{u, k} \tilde{\bfQ}~.
\end{IEEEeqnarray*}
We also have $\tilde{s}_k \leq 2\kappa_{u, k}$ since
\begin{IEEEeqnarray*}{+rCl+x*}
\|\bfQ_{\sigma, l}\|_\infty & = & \max_{i \in \{1, \hdots, 3\}} \sum_{j=1}^3 |(\bfQ_{\sigma, l})_{ij}| = |r_{\sigma, l}| + |s_{\sigma, l}| \leq 2\|\bfu_l\|_\infty~.
\end{IEEEeqnarray*}
Aggregating the previous results and using $\kappa_{u, k-1} \leq \kappa_{u, k}$ yields
\begin{IEEEeqnarray*}{+rCl+x*}
|\bftheta_{i, k} - \bftheta_{i, 0}| & \leq & |\tilde{\bfQ}_{k-1}| |\bftheta_{i, 0}| + \|\bftheta_{i, k} - \tilde{\bfQ}_{k-1} \bftheta_{i, 0} - \bftheta_{i, 0}\|_\infty \bfone_3 \\
& \leq & \kappa_{u, k} \tilde{\bfQ} |\bftheta_{i, 0}| + 2\kappa_{u, k}^2 e^{2\kappa_{u, k}} \|\bftheta_{i, 0}\|_\infty \bfone_3~. \\
|\bfSigma_{\sigma, k} - \bfSigma_{\sigma, 0}| & \leq & |\tilde{\bfQ}_{k-1}| |\bfSigma_{\sigma, 0}| + |\bfSigma_{\sigma, 0}| |\tilde{\bfQ}_{k-1}| \\
&& ~+~ \|\bfSigma_{\sigma, k} - \tilde{\bfQ}_{k-1} \bfSigma_{\sigma, 0} - \bfSigma_{\sigma, 0} \tilde{\bfQ}_{k-1} - \bfSigma_{\sigma, 0}\|_\infty \bfone_{3 \times 3} \\
& \leq & \kappa_{u, k} (\tilde{\bfQ} |\bfSigma_{\sigma, 0}| + |\bfSigma_{\sigma, 0}| \tilde{\bfQ}) + 8\kappa_{u, k}^2 e^{4\kappa_{u, k}} \|\bfSigma_{\sigma, 0}\|_\infty \bfone_{3 \times 3}~. & \qedhere
\end{IEEEeqnarray*}
\end{proof}
\end{proposition}

\begin{corollary} \label{cor:difference_bound_3}
Let \Cref{assumption:training_dynamics} be satisfied. If $\kappa_{u, k} \leq O(m^{\varepsilon + \gpsi - 1})$ for some $k \in \bbN_0$ with bound independent of $k$, we have %
\begin{IEEEeqnarray*}{+rCl+x*}
\Delta_k \bftheta_i & \leq & O(m^{2\varepsilon + \gpsi}) \begin{pmatrix}
O(m^{-3/2}) \\
O(m^{-3/2}) \\
O(m^{-1})
\end{pmatrix}, \\
\Delta_k \bfSigma_\sigma & \leq & O(m^{2\varepsilon + \gpsi}) \begin{pmatrix}
O(m^{-1}) & O(m^{-1}) & O(1) \\
O(m^{-1}) & O(m^{-3/2}) & O(m^{-1}) \\
O(1) & O(m^{-1}) & O(m^{-1})
\end{pmatrix}
\end{IEEEeqnarray*}
with a bound independent of $k$.

\begin{proof}
Note that since $\varepsilon + \gpsi < 1$, we have $e^{4\kappa_{u, k}} \leq O(1)$.
\begin{enumerate}[(a), wide=0pt, leftmargin=*]

\item By properties (W1), (W2), and (W3) in \Cref{prop:prob_W}, we have
\begin{IEEEeqnarray*}{+rCl+x*}
|\bftheta_{i, 0}| = \begin{pmatrix}
|a_{i, 0}| \\ |b_{i, 0}| \\ |w_{i, 0}|
\end{pmatrix} \leq \begin{pmatrix}
m^\varepsilon \\ 0 \\ m^{\varepsilon-1/2}
\end{pmatrix}~.
\end{IEEEeqnarray*}
We can now apply \Cref{prop:difference_bound} to obtain
\begin{IEEEeqnarray*}{+rCl+x*}
|\bftheta_{i, k} - \bftheta_{i, 0}| & \leq & \kappa_{u, k} \tilde{\bfQ} |\bftheta_{i, 0}| + 2\kappa_{u, k}^2 e^{2\kappa_{u, k}} \|\bftheta_{i, 0}\|_\infty \mathbf{1}_3 \\
& \leq & \kappa_{u, k} \begin{pmatrix}
m^{\varepsilon-1/2} \\
m^{\varepsilon-1/2} \\
m^{\varepsilon}
\end{pmatrix} + 2\kappa_{u, k}^2 e^{2\kappa_{u, k}} m^\varepsilon \mathbf{1}_3 \\
& \leq & O(m^{2\varepsilon + \gpsi}) \left(\begin{pmatrix}
O(m^{-3/2}) \\
O(m^{-3/2}) \\
O(m^{-1})
\end{pmatrix} + O(m^{\varepsilon + \gpsi}) \begin{pmatrix}
O(m^{-2}) \\
O(m^{-2}) \\
O(m^{-2})
\end{pmatrix}\right) \\
& \stackrel{\varepsilon + \gpsi \leq 1/2}{=} & O(m^{2\varepsilon + \gpsi}) \begin{pmatrix}
O(m^{-3/2}) \\
O(m^{-3/2}) \\
O(m^{-1})
\end{pmatrix}~.
\end{IEEEeqnarray*}

\item By properties (W1), (W5), (W6) and (W7) in \Cref{prop:prob_W}, we have
\begin{IEEEeqnarray*}{+rCl+x*}
|\bfSigma_{\sigma, 0}| & = & \begin{pmatrix}
|\Sigma_{\sigma, a^2, 0}| & |\Sigma_{\sigma, ab, 0}| & |\Sigma_{\sigma, wa, 0}| \\
|\Sigma_{\sigma, ab, 0}| & |\Sigma_{\sigma, b^2, 0}| & |\Sigma_{\sigma, wb, 0}| \\
|\Sigma_{\sigma, wa, 0}| & |\Sigma_{\sigma, wb, 0}| & |\Sigma_{\sigma, w^2, 0}|
\end{pmatrix} = \begin{pmatrix}
O(m) & 0 & O(m^\varepsilon) \\
0 & 0 & 0 \\
O(m^\varepsilon) & 0 & O(1)
\end{pmatrix}~.
\end{IEEEeqnarray*}
Since $\varepsilon \leq 1$, we can conclude $\|\bfSigma_{\sigma, 0}\|_\infty \leq O(m)$ and
\begin{IEEEeqnarray*}{+rCl+x*}
\tilde{\bfQ} |\bfSigma_{\sigma, 0}| + |\bfSigma_{\sigma, 0}| \tilde{\bfQ} & = & \begin{pmatrix}
O(m^\varepsilon) & 0 & O(1) \\
O(m^\varepsilon) & 0 & O(1) \\
O(m) & 0 & O(m^\varepsilon)
\end{pmatrix} + \begin{pmatrix}
O(m^\varepsilon) & O(m^\varepsilon) & O(m) \\
0 & 0 & 0 \\
O(1) & O(1) & O(m^\varepsilon)
\end{pmatrix} \\
& = & \begin{pmatrix}
O(m^\varepsilon) & O(m^\varepsilon) & O(m) \\
O(m^\varepsilon) & 0 & O(1) \\
O(m) & O(1) & O(m^\varepsilon)
\end{pmatrix}~.
\end{IEEEeqnarray*}
We can now apply \Cref{prop:difference_bound} to obtain
\begin{IEEEeqnarray*}{+rCl+x*}
&& |\bfSigma_{\sigma, k} - \bfSigma_{\sigma, 0}| \\
& \leq & \kappa_{u, k} (\tilde{\bfQ} |\bfSigma_{\sigma, 0}| + |\bfSigma_{\sigma, 0}| \tilde{\bfQ}) + 8\kappa_{u, k}^2 e^{4\kappa_{u, k}} \|\bfSigma_{\sigma, 0}\|_\infty \mathbf{1}_{3 \times 3} \\
& \leq & O(m^{2\varepsilon + \gpsi}) \left(\begin{pmatrix}
O(m^{-1}) & O(m^{-1}) & O(1) \\
O(m^{-1}) & 0 & O(m^{-1}) \\
O(m^{-1}) & O(m^{-1}) & O(m^{-1})
\end{pmatrix} + O(m^{\varepsilon + \gpsi - 2}) \bfone_{3 \times 3} \right) \\
& \stackrel{\varepsilon + \gpsi \leq 1/2}{=} & O(m^{2\varepsilon + \gpsi}) \begin{pmatrix}
O(m^{-1}) & O(m^{-1}) & O(1) \\
O(m^{-1}) & O(m^{-3/2}) & O(m^{-1}) \\
O(1) & O(m^{-1}) & O(m^{-1})
\end{pmatrix}~. & \qedhere
\end{IEEEeqnarray*}
\end{enumerate}
\end{proof}
\end{corollary}

\begin{remark} \label{rem:second_moment_diff_bounds}
We will prove in \Cref{prop:sum-O} that the assumption of \Cref{cor:difference_bound_3} is satisfied. Although the first inequality of \Cref{cor:difference_bound_3} already provides bounds on the change of the individual weights, the second inequality is interesting as well because its bounds are stronger than what one would expect only from the individual weight bounds in the first inequality: For the sake of simplicity, pretend that $\varepsilon = \gamma_\psi = 0$. Then, for example, one could argue using the first inequality that
\begin{IEEEeqnarray*}{+rCl+x*}
\Delta_k \bfSigma_{\sigma, a^2} & = & \max_{0 \leq l \leq k} \left|\sum_{i \in I_\sigma} (a_{i, l}^2 - a_{i, 0}^2)\right| \leq \sum_{i \in I_\sigma} \max_{0 \leq l \leq k} |a_{i, l}^2 - a_{i, 0}^2| \\
& = & \sum_{i \in I_\sigma} \max_{0 \leq l \leq k} |a_{i, l} + a_{i, 0}| \cdot |a_{i, l} - a_{i, 0}| \leq \sum_{i \in I_\sigma} (|a_{i, 0}| + \Delta_k a_i) \Delta_k a_i \\
& \leq & \sum_{i \in I_\sigma} (O(1) + O(m^{-3/2})) O(m^{-3/2}) = |I_\sigma| O(m^{-3/2}) \leq O(m^{-1/2})~,
\end{IEEEeqnarray*}
which is weaker than the bound $\Delta_k \bfSigma_{\sigma, a^2} \leq O(m^{-1})$ obtained by the second inequality. These stronger bounds will be crucial in proving that the assumption $\kappa_{u, k} \leq O(m^{\varepsilon + \gpsi - 1})$ of \Cref{cor:difference_bound_3} is satisfied. Also, note that for $\varepsilon = \gamma_\psi = 0$, the weakest bound
\begin{IEEEeqnarray*}{+rCl+x*}
\Delta_k \Sigma_{\sigma, wa} \leq O(1)
\end{IEEEeqnarray*}
cannot be improved: $\Sigma_{\sigma, wa} = p_\sigma$ is the slope of $f_{W, \bftau, \sigma}$, which initially satisfies $|\Sigma_{\sigma, wa, 0}| \leq O(1)$ by (W7) and needs to converge to an $\popt_\sigma$ that is independent of $\Sigma_{\sigma, wa, 0}$ and also satisfies $|\popt_\sigma| \leq \psi_{D, p} \leq O(1)$. Our proof works since $\Sigma_{\sigma, wa}$ only occurs in $h\Gwab$ with a small factor $h$, but neither in $\Gw$ nor $\Gab$. %
\end{remark}

We will soon use \Cref{cor:difference_bound_3} to prove its own assumption $\kappa_{u, k} \leq O(m^{\varepsilon + \gpsi - 1})$. To this end, we first need a lemma that connects the reference system $\delta \ovl{\bfv} = -h\Aref \bfM \ovl{\bfv}$ to the actual system $\delta \ovl{\bfv} = -h \bfA \bfM \ovl{\bfv}$.

\begin{lemma} \label{lemma:L1-bound:step}
For $m \geq 1$, let $\|\cdot\|$ denote an arbitrary vector norm on $\bbR^m$ and its induced matrix norm. Let $k \in \bbN_0$, $\bfK_0, \hdots, \bfK_{k-1} \in \bbR^{m \times m}$ and $\tilde{\bfK} \in \bbR^{m \times m}$. If
\begin{IEEEeqnarray*}{+rCl+x*}
\delta_{k-1} \equalDef \sum_{l=0}^{k-1} \|\tilde{\bfK}^l\| \cdot \sup_{l \in \{0, \hdots, k-1\}} \|\bfK_l - \tilde{\bfK}\| < 1~,
\end{IEEEeqnarray*}
where $\delta_{-1} \equalDef 0$, then each sequence $\bfv_0, \hdots, \bfv_k \in \bbR^m$ with $\bfv_{l+1} = \bfK_l \bfv_l$ for all $l \in \{0, \hdots, k-1\}$ satisfies
\begin{IEEEeqnarray*}{+rCl+x*}
\sum_{l=0}^k \|\bfv_l\| & \leq & \frac{1}{1-\delta_{k-1}} \sum_{l=0}^k \|\tilde{\bfK}^l \bfv_0\|~.
\end{IEEEeqnarray*}

\begin{proof}
Clearly, for $l \in \{0, \hdots, k-1\}$,
\begin{IEEEeqnarray*}{+rCl+x*}
\bfv_{l+1} & = & \tilde{\bfK} \bfv_l + (\bfK_l - \tilde{\bfK})\bfv_l
\end{IEEEeqnarray*}
and hence, by induction on $l$,
\begin{IEEEeqnarray*}{+rCl+x*}
\bfv_l & = & \tilde{\bfK}^l \bfv_0 + \sum_{l'=0}^{l-1} \tilde{\bfK}^{l-1-l'} (\bfK_{l'} - \tilde{\bfK}) \bfv_{l'}
\end{IEEEeqnarray*}
for all $l \in \{0, \hdots, k\}$. Summing norms on both sides yields
\begin{IEEEeqnarray*}{+rCl+x*}
\sum_{l=0}^k \|\bfv_l\| & \leq & \sum_{l=0}^k \|\tilde{\bfK}^l \bfv_0\| + \sum_{l=0}^k \sum_{l'=0}^{l-1} \|\tilde{\bfK}^{l-1-l'} (\bfK_{l'} - \tilde{\bfK}) \bfv_{l'}\| \\
& = & \sum_{l=0}^k \|\tilde{\bfK}^l \bfv_0\| + \sum_{l'=0}^{k-1} \sum_{l=l'+1}^k \|\tilde{\bfK}^{l-1-l'} (\bfK_{l'} - \tilde{\bfK}) \bfv_{l'}\| \\
& \leq & \sum_{l=0}^k \|\tilde{\bfK}^l \bfv_0\| + \sum_{l'=0}^{k-1} \left(\sum_{l=0}^{k-1-l'} \|\tilde{\bfK}^l\|\right) \cdot \sup_{l \in \{0, \hdots, k-1\}} \|\bfK_l - \tilde{\bfK}\| \cdot \|\bfv_{l'}\| \\
& \leq & \sum_{l=0}^k \|\tilde{\bfK}^l \bfv_0\| + \delta_{k-1} \sum_{l'=0}^k \|\bfv_{l'}\|~.
\end{IEEEeqnarray*}
Hence $(1-\delta_{k-1}) \sum_{l=0}^k \|\bfv_l\| \leq \sum_{l=0}^k \|\tilde{\bfK}^l \bfv_0\|$ and since $\delta_{k-1} < 1$, the inequality is preserved when dividing by $1-\delta_{k-1}$.
\end{proof}
\end{lemma}

\begin{proposition} \label{prop:sum-O}
Let \Cref{assumption:training_dynamics} be satisfied. We have
\begin{IEEEeqnarray*}{+rCl+x*}
\kappa_{u, k} \leq O(m^{\varepsilon + \gpsi - 1})~,
\end{IEEEeqnarray*}
where $\kappa_{u, k}$ was defined in \Cref{def:kappa_Q_1} and the bound $O(m^{\varepsilon + \gpsi - 1})$ is independent of $k \in \bbN_0$.

\begin{proof}
By \Cref{prop:deriv:AM}, we know that $\ovl{\bfv}_{k+1} = (\bfI_4 - h\bfA_k \bfM_D)\ovl{\bfv}_k$. We want to bound $\kappa_{u, k} = h\sum_{l=0}^k \|\bfu_l\|_\infty = h\sum_{l=0}^k \|\bfB\bfM_D\ovl{\bfv}_l\|_\infty$ by comparing it to the reference system $\delta \ovl{\bfv} = -h\Aref \bfM_D \ovl{\bfv}$ using \Cref{lemma:L1-bound:step}. Hence, we define
\begin{IEEEeqnarray*}{+rCl+x*}
\delta_k & \equalDef & \sum_{l=0}^k \|(\bfI_4 - h\Aref \bfM_D)^l\|_\infty \cdot \sup_{0 \leq l \leq k} \|(\bfI_4 - h\Aref \bfM_D) - (\bfI_4 - h\bfA_l\bfM_D)\|_\infty \\
& = & h\sum_{l=0}^k \|(I - h\Aref \bfM_D)^l\|_\infty \cdot \sup_{0 \leq l \leq k} \|(\bfA_l - \Aref)\bfM_D\|_\infty~. \IEEEyesnumber \label{eq:def_delta_k}
\end{IEEEeqnarray*}
For $m$ large enough, we want to prove by induction that $\delta_k \leq 1/2$ for all $k \in \bbN_0$. Trivially, $\delta_{-1} = 0 \leq 1/2$. Now let $k \in \bbN_0$ with $\delta_{k-1} \leq 1/2$.

\begin{enumerate}[(1), wide=0pt, leftmargin=*]

\item By \Cref{lemma:deriv:linear_relations}, we have $\tilde{\bfu}_k = -\tilde{\bfB}\tilde{\bfM}_D \tilde{\ovl{\bfv}}_k$ and hence $\bfu_k = -\bfB\bfM_D \ovl{\bfv}_k$. Thus,
\begin{IEEEeqnarray*}{+rCl+x*}
\kappa_{u, k} & = & h\sum_{l=0}^k \|\bfu_l\|_\infty \leq \|\bfB\|_\infty \|\bfM_D\|_\infty \cdot h\sum_{l=0}^k \|\ovl{\bfv}_l\|_\infty~.
\end{IEEEeqnarray*}
Because $\delta_{k-1} \leq 1/2$, we can apply \Cref{lemma:L1-bound:step} and obtain
\begin{IEEEeqnarray*}{+rCl+x*}
h\sum_{l=0}^k \|\ovl{\bfv}_l\|_\infty & \leq & \frac{1}{1-\delta_{k-1}} h\sum_{l=0}^k \|(\bfI_4 - h\Aref \bfM_D)^l \ovl{\bfv}_0\|_\infty \\
& \leq & 2 h\sum_{l=0}^\infty \|(\bfI_4 - h\Aref \bfM_D)^l \ovl{\bfv}_0\|_\infty \\
& \stackrel{\text{\Cref{prop:ref_sum}}}{\leq} & O(m^{\varepsilon + \gpsi - 1})~.
\end{IEEEeqnarray*}
Norm equivalence (cf.\ \Cref{def:matrix_notation}) yields
\begin{IEEEeqnarray*}{+rCl+x*}
\|\bfM_D\|_\infty \leq O(\|\bfM_D\|_2) = O(\lmax(\bfM_D)) \stackrel{\text{\Cref{assumption:training_dynamics}}}{=} O(1)~. \IEEEyesnumber \label{eq:norm-M_D-is-bounded}
\end{IEEEeqnarray*}
Hence, we can write
\begin{IEEEeqnarray*}{+rCl+x*}
\kappa_{u, k} & = & O(m^{\varepsilon + \gpsi - 1})~, \IEEEyesnumber \label{eq:kappa_u_k-is-small}
\end{IEEEeqnarray*}
where, in accordance with \Cref{def:O-notation}, the constant in $O(m^{\varepsilon + \gpsi - 1})$ does not depend on the induction step $k$.

\item Let us investigate the components of Eq.~\eqref{eq:def_delta_k}:
\begin{IEEEeqnarray*}{+rCl+x*}
h\sum_{l=0}^k \|(\bfI_4 - h\Aref \bfM_D)^l\|_\infty & \leq & h\sum_{l=0}^\infty \|(\bfI_4 - h\Aref \bfM_D)^l\|_\infty \stackrel{\text{\Cref{prop:ref_sum}}}{=} O(1) \\
(\bfA_l - \Aref)\bfM_D & = & \bfB\left((\Gw_l - \Gw_0) + (\Gab_l - \Gab_0) + h\Gwab_l\right)\bfB\bfM_D \\
\Rightarrow \|(\bfA_l - \Aref)\bfM_D\|_\infty & \stackrel{\eqref{eq:norm-M_D-is-bounded}}{=} & O(1) \cdot (\|\Gw_l - \Gw_0\|_\infty + \|\Gab_l - \Gab_0\|_\infty + h\|\Gwab_l\|_\infty).
\end{IEEEeqnarray*}

First of all, for $0 \leq l \leq k$,
\begin{IEEEeqnarray*}{+rCl+x*}
\|\Gw_l - \Gw_0\|_\infty & \leq & \max_{\sigma \in \{\pm 1\}} \Delta_k \Sigma_{\sigma, w^2} \stackrel{\text{\Cref{cor:difference_bound_3}}}{\leq} O(m^{2\varepsilon + \gpsi - 1})~.
\end{IEEEeqnarray*}
Similarly, for $0 \leq l \leq k$,
\begin{IEEEeqnarray*}{+rCl+x*}
\|\Gab_l - \Gab_0\|_\infty & \leq & \max_{\sigma \in \{\pm 1\}} (\Delta_k \Sigma_{\sigma, a^2} + \Delta_k \Sigma_{\sigma, ab} + \Delta_k \Sigma_{\sigma, b^2}) \leq O(m^{2\varepsilon + \gamma_\psi - 1})~.
\end{IEEEeqnarray*}

Observe that
\begin{IEEEeqnarray*}{+rCl+x*}
h|r_{\sigma, l}| & \leq & h\|\bfu_l\|_\infty \leq h\sum_{l'=0}^k \|\bfu_{l'}\|_\infty = \kappa_{u, k} \stackrel{\eqref{eq:kappa_u_k-is-small}}{=} O(m^{\varepsilon + \gpsi - 1})
\end{IEEEeqnarray*}
and similarly $h|s_{\sigma, l}| \leq O(m^{\varepsilon + \gpsi - 1})$. Thus, we find
\begin{IEEEeqnarray*}{+rCl+x*}
h\|\Gwab_l\|_\infty & = & \max_{\sigma \in \{\pm 1\}} h|r_{\sigma, l} \Sigma_{\sigma, wa, l} + s_{\sigma, l} \Sigma_{\sigma, wb, l}| \\
& = & O(m^{\varepsilon + \gpsi - 1}) \cdot \left(\max_{\sigma \in \{\pm 1\}} |\Sigma_{\sigma, wa, l}| + |\Sigma_{\sigma, wb, l}|\right)~.
\end{IEEEeqnarray*}
Similar to the other calculations, we can compute for $0 \leq l \leq k$
\begin{IEEEeqnarray*}{+rCl+x*}
|\Sigma_{\sigma, wa, l}| & \leq & |\Sigma_{\sigma, wa, 0}| + \Delta_k \Sigma_{\sigma, wa} \stackrel{\text{(W7)}}{\leq} O(m^\varepsilon) + O(m^{2\varepsilon + \gpsi}) = O(m^{2\varepsilon + \gpsi}) \\
|\Sigma_{\sigma, wb, l}| & \leq & |\Sigma_{\sigma, wb, 0}| + \Delta_k \Sigma_{\sigma, wb} \leq 0 + O(m^{2\varepsilon + \gpsi - 1}) = O(m^{2\varepsilon + \gpsi - 1})~,
\end{IEEEeqnarray*}
which yields $h\|\Gwab_l\|_\infty \leq O(m^{3\varepsilon + 2\gpsi - 1})$.

We can now revisit the beginning of step (2) to obtain 
\begin{IEEEeqnarray*}{+rCl+x*}
\|(\bfA_l - \Aref)\bfM_D\|_\infty \leq O(m^{3\varepsilon + 2\gpsi - 1})
\end{IEEEeqnarray*}
and
\begin{IEEEeqnarray*}{+rCl+x*}
\delta_k & \stackrel{\eqref{eq:def_delta_k}}{=} & h\sum_{l=0}^k \|(\bfI_4 - h\Aref \bfM_D)^l\|_\infty \cdot \sup_{0 \leq l \leq k} \|(\bfA_l - \Aref)\bfM_D\|_\infty \\
& = & O(1) \cdot O(m^{3\varepsilon + 2\gpsi - 1}) = O(m^{3\varepsilon + 2\gpsi - 1})~.
\end{IEEEeqnarray*}
\end{enumerate}
We have shown that $\delta_{k-1} \leq 1/2$ implies $\delta_k \leq O(m^{3\varepsilon + 2\gpsi - 1})$, where the constant in $O(m^{3\varepsilon + 2\gpsi - 1})$ does not depend on $k$. Since $3\varepsilon + 2\gpsi < 1$ by \Cref{assumption:training_dynamics}, we have 
$\lim_{m \to \infty} m^{3\varepsilon + 2\gpsi - 1} = 0$ and there exists $m_0 \in \bbN_0$ such that for all $m \geq m_0$ and $k \in \bbN_0$, $\delta_{k-1} \leq 1/2$ implies $\delta_k \leq 1/2$ and the induction works. 
Thus, for all $m \geq m_0$ and $k \in \bbN_0$, we know that $\delta_{k-1} \leq 1/2$ and we can apply step (1) to obtain 
\begin{IEEEeqnarray*}{+rCl+x*}
\kappa_{u, k} & = & O(m^{\varepsilon + \gpsi - 1})~. & \qedhere
\end{IEEEeqnarray*}
\end{proof}
\end{proposition}

We can now prove our main theorem:

\begin{proof}[Proof of \Cref{thm:main:general}]

Since $\gpsi + \gx + \gP < 1/2$, we can assume without loss of generality that
\begin{IEEEeqnarray*}{+rCl+x*}
0 < \varepsilon < \frac{1/2 - (\gpsi + \gx + \gP)}{3}~.
\end{IEEEeqnarray*}
Moreover, if (W1) -- (W7) from \Cref{prop:prob_W} are satisfied, we have (similar to Eq.~\eqref{eq:lminH} in the proof of \Cref{prop:ref_sum})
\begin{IEEEeqnarray*}{+rCl+x*}
\lmax(\bfH) & = & \lmax(\bfM_D^{1/2} \Aref \bfM_D^{1/2}) \leq \|\bfM_D^{1/2}\|_2^2 \lmax(\Aref) = \lmax(\bfM_D) \lmax(\Aref)  \\
& \stackrel{\text{\Cref{lemma:eig_Aref}}}{\leq} & \CM \Theta(m)~.
\end{IEEEeqnarray*}
Hence, there exists a constant $\Ch^{-1} > 0$ with $\lmax(\bfH) \leq \Ch^{-1} m$. For this choice, assumption \eqref{step-size-assumption} yields 
\begin{IEEEeqnarray*}{+rCl+x*}
h \leq \Ch m^{-1} \leq \frac{1}{\lmax(\bfH)}~.
\end{IEEEeqnarray*}
Therefore, \Cref{assumption:training_dynamics} is satisfied whenever the initialization satisfies (W1) -- (W7).

By our choice of $\varepsilon$, we have
\begin{IEEEeqnarray*}{+rCl+x*}
2\varepsilon + \gpsi - 3/2 & < & -1-\gP - \gx \leq -1-\gP~. \IEEEyesnumber \label{eq:eps}
\end{IEEEeqnarray*}
For $\udl{x} \equalDef \Cx m^{-\gx}$, we then obtain using (W4), \Cref{cor:difference_bound_3} and \Cref{prop:sum-O}:
\begin{IEEEeqnarray*}{+rCl+x*}
(|a_{i, 0}| - |a_{i, 0} - a_{i, k}|) \udl{x} & \geq & \left(\Omega(m^{-1-\gP}) - O(m^{2\varepsilon + \gpsi - 3/2})\right) \Theta(m^{-\gx}) \stackrel{\eqref{eq:eps}}{=} \Omega(m^{-1-\gP-\gx}) \\
|b_{i, k}| & = & |b_{i, k} - b_{i, 0}| \leq O(m^{2\varepsilon + \gpsi - 3/2})~.
\end{IEEEeqnarray*}
Using \eqref{eq:eps} again, we find that there exists $m_0$ such that for all $m \geq m_0$ and all $i \in I, k \in \bbN_0$,
\begin{IEEEeqnarray*}{+rCl+x*}
|b_{i, k}| < (|a_{i, 0}| - |a_{i, 0} - a_{i, k}|) \udl{x}~,
\end{IEEEeqnarray*}
which, in terms of \Cref{lemma:linearization_region}, means $W_k \in \calS_{W_0}(\udl{x})$. Since $\udl{x} \leq \udl{x}_D$ by assumption, we also have $W_k \in \calS_{W_0}(\udl{x}_D)$. But then, \Cref{lemma:linearization_region} tells us that $\nabla L_{D, \bftau}(W_k) = \nabla L_D(W_k)$ and that $f_W|_{[\udl{x}, \infty)} = f_{W, \bftau, 1}|_{[\udl{x}, \infty)}$ as well as $f_W|_{(-\infty, -\udl{x}]} = f_{W, \bftau, -1}|_{(-\infty, -\udl{x}]}$ are affine. Hence, $(W_k)_{k \in \bbN_0}$ satisfies the original gradient descent iteration
\begin{IEEEeqnarray*}{+rCl+x*}
W_{k+1} = W_k - h\nabla L_D(W_k)
\end{IEEEeqnarray*}
and (v) is satisfied. Moreover, by \Cref{cor:difference_bound_3}, we obtain (i), (ii) and (iii) (up to a factor $2$ in front of $\varepsilon$, which can be resolved by shrinking $\varepsilon$):
\begin{IEEEeqnarray*}{+rCl+x*}
|a_{i, k} - a_{i, 0}| & \leq & O(m^{2\varepsilon + \gpsi - 3/2}) \\
|b_{i, k} - b_{i, 0}| & \leq & O(m^{2\varepsilon + \gpsi - 3/2}) \\
|w_{i, k} - w_{i, 0}| & \leq & O(m^{2\varepsilon + \gpsi - 1})~.
\end{IEEEeqnarray*}
In order to find a similar bound for $c$, we recall from \Cref{lemma:deriv:linear_relations} that $\delta c = h(\hat{s}_1 + \hat{s}_{-1})$, from \Cref{def:kappa_Q_1} that $\kappa_{u, k} = h\sum_{l=0}^k \|\bfu_l\|_\infty$ and from \Cref{def:derived_quantities} that
\begin{IEEEeqnarray*}{+rCl+x*}
\bfu & = & \bfB \begin{pmatrix}
\hat{r}_1 \\ \hat{r}_{-1} \\ \hat{s}_1 \\ \hat{s}_{-1}
\end{pmatrix} \quad \text{ and therefore } \quad \begin{pmatrix}
\hat{r}_1 \\ \hat{r}_{-1} \\ \hat{s}_1 \\ \hat{s}_{-1}
\end{pmatrix} = \bfB^{-1} \bfu~,
\end{IEEEeqnarray*}
since $\bfB$ and invertible due to $|\alpha| \neq 1$. Because $\bfB$ is fixed, we therefore obtain
\begin{IEEEeqnarray*}{+rCl+x*}
|c_k - c_0| & \leq & \sum_{l=0}^{k-1} |c_{l+1} - c_l| = \sum_{l=0}^{k-1} |\delta c_l| = h\sum_{l=0}^{k-1} |\hat{s}_{1, l} + \hat{s}_{-1, l}| \leq O(\kappa_{u, k}) \\
& \stackrel{\text{\Cref{prop:sum-O}}}{\leq} & O(m^{2\varepsilon + \gpsi - 1})~,
\end{IEEEeqnarray*}
which shows (iv) after rescaling $\varepsilon$.

All of this holds under the assumption that $m \geq m_0$ and (W1) -- (W7), where $m_0$ is independent of $W_0, k, h$. By \Cref{prop:prob_W}, the assumption holds with probability $\geq 1 - O(m^{-\gP})$.
\end{proof}

\section{Multi-Dimensional Inputs} \label{sec:multi-d}

In the following, we investigate the case where 
\begin{itemize}

\item the one-dimensional $x$ values of $D$ are projected onto a line in a $d$-dimensional input space ($d \in \bbN$), and
\item a two-layer neural network with $d$-dimensional input is trained on this (degenerate) data set.
\end{itemize}
In \Cref{rem:ddim}, we show that this $d$-dimensional case can be reduced to the case $d = 1$ and hence, comparable conclusions hold:

\begin{remark} \label{rem:ddim}
Let $d \geq 1$ and let $\bfz \in \bbR^d$ with $\|\bfz\|_2 = 1$. For a data set $D \in (\bbR \times \bbR)^n$, we consider an embedded data set $\tilde{D} = ((\bfz x_1, y_1), \hdots, (\bfz x_n, y_n)) \in (\bbR^d \times \bbR)^n$ and a neural network function
\begin{IEEEeqnarray*}{+rCl+x*}
f_{\tilde{W}}(\bfx) & \equalDef & \tilde{c} + \sum_{i=1}^m \tilde{w}_i \varphi(\tilde{\bfa}_i^\top \bfx + \tilde{b}_i)~.
\end{IEEEeqnarray*}
Then, 
\begin{IEEEeqnarray*}{+rCl+x*}
f_{\tilde{W}}(\bfz x) & \equalDef & \tilde{c} + \sum_{i=1}^m \tilde{w}_i \varphi(\tilde{\bfa}_i^\top \bfz x + \tilde{b}_i) = \tilde{c} + \sum_{i=1}^m \tilde{w}_i \varphi((\bfz^\top \tilde{\bfa}_i) x + \tilde{b}_i) = f_{\bfZ \tilde{W}}(x)~, \IEEEyesnumber \label{eq:ddim-gd}
\end{IEEEeqnarray*}
where
\begin{IEEEeqnarray*}{+rCl+x*}
\bfZ \tilde{W} & \equalDef & \begin{pmatrix}
\bfz^\top \\
& \ddots \\
& & \bfz^\top \\
&&& \bfI_m \\
&&&& 1 \\
&&&&& \bfI_m
\end{pmatrix} \begin{pmatrix}
\tilde{\bfa}_1 \\ \vdots \\ \tilde{\bfa}_m \\ \tilde{\bfb} \\ \tilde{c} \\ \tilde{\bfw}
\end{pmatrix} = \begin{pmatrix}
\bfz^\top \tilde{\bfa}_1 \\ \vdots \\ \bfz^\top \tilde{\bfa}_m \\ \tilde{\bfb} \\ \tilde{c} \\ \tilde{\bfw}
\end{pmatrix} \in \bbR^{3m+1}~.
\end{IEEEeqnarray*}
If we naturally extend the definition of $L_D$ in the usual way for $d$-dimensional inputs, Eq.~\eqref{eq:ddim-gd} yields $L_{\tilde{D}}(\tilde{W}) = L_D(\bfZ \tilde{W})$. Moreover, since $\|\bfz\|_2 = 1$, we have $\bfZ \bfZ^\top = \bfI_{3m+1}$. Using these two insights, we obtain for $W_k \equalDef \bfZ \tilde{W}_k$:
\begin{IEEEeqnarray*}{+rCl+x*}
W_{k+1} & = & \bfZ \tilde{W}_{k+1} = \bfZ (\tilde{W}_k - h \nabla_{\tilde{W}_k} L_{\tilde{D}}(\tilde{W}_k)) = \bfZ \tilde{W}_k - h \bfZ \nabla_{\tilde{W}_k} L_D(\bfZ \tilde{W}_k) \\
& = & \bfZ \tilde{W}_k - h \bfZ \bfZ^\top \nabla L_D(\bfZ \tilde{W}_k) = \bfZ \tilde{W}_k - h \nabla L_D(\bfZ \tilde{W}_k) = W_k - h\nabla L_D(W_k)~.
\end{IEEEeqnarray*}
Hence, $(W_k)_{k \in \bbN_0}$ satisfy the gradient descent equation for the original data set $D$. Moreover, if we initialize $\tilde{W}_0$ analogous to \Cref{ass:init}, i.e.,
\begin{IEEEeqnarray*}{+rCl+x*}
\tilde{b}_i & = & 0, \quad \tilde{c} = 0, \quad \tilde{w}_i \sim \frac{1}{\sqrt{m}} \tZw, \quad \tilde{a}_{il} \sim \tZa \IEEEyesnumber \label{eq:multi_d_init} %
\end{IEEEeqnarray*}
with independent variables, then the initial vector $\tilde{W}_0 = \bfZ W_0$ satisfies
\begin{IEEEeqnarray*}{+rCl+x*}
b_i & = & \tilde{b}_i = 0, \quad c = \tilde{c} = 0, \quad w_i = \tilde{w}_i \sim \frac{1}{\sqrt{m}} \tZw, \quad a_i = \sum_{l=1}^d z_l \tilde{a}_{il} \sim \Za
\end{IEEEeqnarray*}
with independent variables and suitable $\Za$. The random variables $(\Za, \tZw)$ satisfy (Q1) and (Q2) from \Cref{ass:init} (we only need to verify them for $\Za$):
\begin{enumerate}[(Q1), wide=0pt, leftmargin=*]

\item It is well-known that the sum of independent $\bbR$-valued random-variables $X, Y$ with densities $p_X, p_Y$ has density
\begin{IEEEeqnarray*}{+rCl+x*}
p_{X + Y}(x) & = & \int_\bbR p_X(x-y) p_Y(y) \diff y~.
\end{IEEEeqnarray*}
Hence, if we know that there exists a bound $B \in (0, \infty)$ with $p_X(x) \leq B$ for all $x \in \bbR$, then
\begin{IEEEeqnarray*}{+rCl+x*}
p_{X + Y}(x) & \leq & B \int_{-\infty}^\infty p_Y(y) \diff y = B~.
\end{IEEEeqnarray*}
Moreover, if $p_X$ and $p_Y$ are symmetric, then $p_{X+Y}$ is also symmetric:
\begin{IEEEeqnarray*}{+rCl+x*}
p_{X + Y}(x) & = & \int_\bbR p_X(x-y) p_Y(y) \diff y = \int_\bbR p_X(y-x) p_Y(-y) \diff y \\
& \stackrel{y' \equalDef -y}{=} & \int_\bbR p_X((-x) - y') p_Y(y') \diff y' = p_{X+Y}(-x)~.
\end{IEEEeqnarray*}
This directly yields (Q1) for $\Za$.
\item Since $\Za$ can be written as a linear combination of random variables that satisfy (Q2), $\Za$ must also satisfy (Q2) by the Minkowski inequality.
\end{enumerate}
By Eq.~\eqref{eq:ddim-gd}, we obtain
\begin{IEEEeqnarray*}{+rCl+x*}
f_{\tilde{W}_k}(\bfz x) & = & f_{W_k}(x)
\end{IEEEeqnarray*}
for all $k \in \bbN_0$ and $x \in \bbR$. Especially, we can apply Theorems \ref{thm:main:general}, \ref{thm:main:fixed_D} and \ref{thm:main:inconsistency} and obtain that under the assumptions of these theorems, the kinks of $x \mapsto f_{\tilde{W}_k}(\bfz x)$ do not cross the data points with high probability. 

Since the assumptions of the theorems are (up to modifying constants) invariant under multiplying the $x_j$ by a positive constant, we can also allow $\|\bfz\|_2 \neq 1$ as long as $\bfz \neq 0$.
\end{remark}

\section{Inconsistency Proofs} \label{sec:appendix:inconsistency}

In this section, we give proofs of the inconsistency results in \Cref{sec:inconsistency}.

\begin{proof}[Proof of \Cref{cor:main:inconsistency}]
Let $\gpsi \in (0, 1/2)$ be sufficiently large such that $\frac{1}{2 - 2\gpsi} \geq 1 - \varepsilon$. We can then choose $\Cn > 0$ sufficiently large such that
\begin{IEEEeqnarray*}{+rCl+x*}
m_n & \leq & \Cn n^{\frac{1}{2 - 2\gpsi}}
\end{IEEEeqnarray*}
for all $n \geq 1$. Choose $\Cx > 0$ such that $\Pdata_X([-\Cx, \Cx]) = 0$. Moreover, let $\gx = 0$ and $\gP > 0$ such that $\gP + \gx + \gpsi < 1/2$.

Let $\Ch$ be the corresponding constant from \Cref{thm:main:inconsistency}. Since $m_n \to \infty$ and $h_n < o(m_n^{-1})$, there exists an $n_0$ such that for all $n \geq n_0$, we have
\begin{IEEEeqnarray*}{+rCl+x*}
h_n & \leq & \Ch m_n^{-1}~.
\end{IEEEeqnarray*}
By \Cref{thm:main:inconsistency} we hence obtain for all $n \geq n_0$ that $f_{W_k}$ is affine on $(-\infty, -\Cx]$ and $[\Cx, \infty)$ with probability $\geq 1 - \CP m_n^{-\gP} \to 1 \quad (n \to \infty)$. But such a function satisfies $f_{W_k} \in \calF_{\Cx}$ and, because $\Pdata_X([-\Cx, \Cx]) = 0$, we obtain
\begin{IEEEeqnarray*}{+rCl+x*}
R_{\Pdata}(f_{W_k}) & \geq & R_{\Pdata, \Cx}^* = R_{\Pdata, 0}^* \stackrel{\text{(P4)}}{>} R_{\Pdata}^*~,
\end{IEEEeqnarray*}
which yields inconsistency.
\end{proof}

\begin{proof}[Proof of \Cref{cor:main:inconsistency:multi-d}]
Consider an NN as in \Cref{cor:main:inconsistency}
that is inconsistent on a distribution $\Pdata$ on $\bbR \times \bbR$. Furthermore, 
fix an arbitrary vector $\bfz \in \bbR^d$ with $\|\bfz\|_2 = 1$.

For $(x, y) \sim \Pdata$, let $\tPdata$ denote the distribution of $(x\bfz, y)$. It is easy to show that the optimal population risks satisfy
\begin{IEEEeqnarray*}{+rCl+x*}
R_{\tPdata}^* = R_{\Pdata}^*~. \IEEEyesnumber \label{eq:optimal_risks_equal}
\end{IEEEeqnarray*}
Let $D \sim (\Pdata)^n$, i.e., $D$ consists of $n$ i.i.d.\ data points $(x_j, y_j) \sim \Pdata$, then $\tilde{D} \sim (\tPdata)^n$ with $\tilde{D}$ defined in \Cref{rem:ddim}. Let $\tilde{W}_0$ be independent from $\tilde{D}$ and initialized analogous to \Cref{ass:init} as discussed in \Cref{rem:ddim}. Let 
\begin{IEEEeqnarray*}{+rCl+x*}
\tilde{W}_{k+1} & = & \tilde{W}_k - h_n \nabla L_{\tilde{D}}(\tilde{W}_k)~.
\end{IEEEeqnarray*}
Let $W_k \equalDef \bfZ \tilde{W}_k$. As shown in \Cref{rem:ddim}, $W_0$ satisfies \Cref{ass:init} and $(W_k)_{k \in \bbN_0}$ arises from gradient descent on $D$:
\begin{IEEEeqnarray*}{+rCl+x*}
W_{k+1} & = & W_k - h_n \nabla L_D(W_k)~.
\end{IEEEeqnarray*}
Moreover, $f_{\tilde{W}_k}(x\bfz) = f_{W_k}(x)$ for all $x \in \bbR, k \in \bbN_0$ and therefore
\begin{IEEEeqnarray*}{+rCl+x*}
R_{\tPdata}(f_{\tilde{W}_k}) & = & R_{\Pdata}(f_{W_k}) \IEEEyesnumber \label{eq:risks_equal}
\end{IEEEeqnarray*}
for all $k \in \bbN_0$. Since the NN with one-dimensional input is inconsistent on $\Pdata$, the NN with $d$-dimensional input must be inconsistent on $\tPdata$ by \eqref{eq:optimal_risks_equal} and \eqref{eq:risks_equal}.
\end{proof}

In order to prove \Cref{cor:main:inconsistency-finite-eta}, we first show that if functions that are affine on $\bbR \setminus \{0\}$ cannot approach the Bayes risk $R_{\Pdata}^*$, there exists $\delta > 0$ such that the same holds for functions that are affine on $\bbR \setminus [-\delta, \delta]$.

\begin{lemma} \label{lemma:delta_affine_bayes_risk}
Let $\Pdata$ be a bounded distribution on $\bbR \times \bbR$ satisfying (P1). For $\delta \geq 0$, define 
\begin{IEEEeqnarray*}{+rCl+x*}
\calF_\delta \equalDef \{f: \bbR \to \bbR \mid f \text{ affine on } (-\infty, -\delta) \text{ and } (\delta, \infty)\}, \quad R_{\Pdata, \delta}^* \equalDef \inf_{f \in \calF_\delta} R_{\Pdata}(f)~.
\end{IEEEeqnarray*}
Then, $\lim_{\delta \searrow 0} R_{\Pdata, \delta}^* = R_{\Pdata, 0}^*$.

\begin{proof}
In the case $\eta = \infty$, we obviously have $R_{\Pdata, \delta}^* = R_{\Pdata, 0}^*$ for all $\delta \geq 0$ with $\Pdata_X((-\delta, \delta)) = 0$ and we are done. 

Obviously, $R_{\Pdata, \delta}^*$ is non-increasing in $\delta$. In order to derive a contradiction, assume that $\lim_{\delta \searrow 0} R_{\Pdata, \delta}^* < R_{\Pdata, 0}^*$.

For $\delta' > 0, \sigma \in \{\pm 1\}$, consider $P_\sigma \equalDef \Pdata(\cdot \mid \sigma X > \delta')$. For sufficiently small $\delta'$, $P$ is well-defined and $P_X$ is not only concentrated on a single $x$ value due to (P1). Now, fix such a $\delta' > 0$.

For $f \in \calF_\delta$, define $\bfv_\sigma(f) \in \bbR^{2 \times 2}$ as the slope and intercept of $f$ on $\sigma(\delta, \infty)$. As in \Cref{def:main:quantities}, we can construct an invertible matrix $\bfM_{P_\sigma} = \bfM_{P_\sigma, \sigma}$ and a vector $\vopt_{P_\sigma} = \vopt_{P_\sigma, \sigma}$. Analogous to \Cref{rem:affine_regression}, we obtain for $\delta \leq \delta'$:
\begin{IEEEeqnarray*}{+rCl+x*}
R_P(f) & \geq & (\bfv_\sigma(f) - \vopt_{P_\sigma})^\top \bfM_{P_\sigma} (\bfv_\sigma(f) - \vopt_{P_\sigma})~.
\end{IEEEeqnarray*}
Hence, for $f \in \calF_\delta$ with $\delta \leq \delta'$ and $R_{\Pdata}(f) \leq R_{\Pdata, 0}^*$, we obtain
\begin{IEEEeqnarray*}{+rCl+x*}
R_{\Pdata, 0}^* & \geq & R_{\Pdata}(f) \geq \frac{1}{2}\bbE_{(x, y) \sim \Pdata} \bbone_{\sigma(\delta', \infty)}(x) (y - f(x))^2 \\
& = & \frac{1}{2}\Pdata_X(\sigma(\delta', \infty)) \cdot \bbE_{(x, y) \sim P_\sigma} (y - f(x))^2 \\
& \geq & \frac{1}{2}\Pdata_X(\sigma(\delta', \infty)) \cdot (\bfv_\sigma(f) - \vopt_{P_\sigma})^\top \bfM_{P_\sigma} (\bfv_\sigma(f) - \vopt_{P_\sigma})~.
\end{IEEEeqnarray*}
Since $\bfM_{P_\sigma} \matgr 0$ and $\Pdata_X(\sigma(\delta', \infty)) > 0$, there has to exist a constant $C$ such that $\|\bfv_\sigma(f)\|_\infty \geq C$ for all such $f$ and $\sigma \in \{\pm 1\}$.

Now, pick $f \in \calF_\delta$, $0 < \delta \leq \delta'$, with $R_{\Pdata}(f) \leq R_{\Pdata, 0}^*$ and let $f_0 \in \calF_{\mathrm{hsal}}$ be its affine continuation (i.e., $f_0(x) = f(x)$ for $|x| > \delta$). Then, $|f_0(x)| \leq C(1+|x|)$ for $x > 0$ and therefore
\begin{IEEEeqnarray*}{+rCl+x*}
R_{\Pdata, 0}^* & \leq & R_{\Pdata}(f_0) \\
& = & \frac{1}{2}\bbE_{(x, y) \sim \Pdata} \bbone_{[-\delta, \delta]}(x) (y - f_0(x))^2 + \frac{1}{2}\bbE_{(x, y) \sim \Pdata} \bbone_{[-\delta, \delta]^c}(x) (y - f_0(x))^2 \\
& \leq & \frac{1}{2} \bbE_{(x, y) \sim \Pdata} \bbone_{[-\delta, \delta]}(x) 2(y^2 + f_0(x)^2) + R_{\Pdata}(f) \\
& \leq & \underbrace{\bbE_{(x, y) \sim \Pdata} \bbone_{[-\delta, \delta]}(x) y^2}_{\to 0 \quad (\delta \to 0)} + \underbrace{\Pdata_X([-\delta, \delta]) \cdot (C(1+|\delta|))^2}_{\to 0 \quad (\delta \to 0)} + R_{\Pdata}(f)~.
\end{IEEEeqnarray*}
Since we can choose $R_{\Pdata}(f)$ arbitrarily close to $R_{\Pdata, \delta}^*$, it follows that $R_{\Pdata, 0}^* \leq \lim_{\delta \searrow 0} R_{\Pdata, \delta}^*$, which contradicts our initial assumption.
\end{proof}
\end{lemma}

\begin{proof}[Proof of \Cref{cor:main:inconsistency-finite-eta}]
Since
\begin{IEEEeqnarray*}{+rCl+x*}
\frac{1}{\gamma \eta} & < & \frac{1}{2}~, \\
1 - \frac{1}{2\gamma} + \frac{1}{\eta \gamma} & = & 1 - \frac{1 - \frac{2}{\eta}}{2\gamma} = \frac{2\gamma - \left(1 - \frac{2}{\eta}\right)}{2\gamma} < \frac{1}{2}~,
\end{IEEEeqnarray*}
there exist some
\begin{IEEEeqnarray*}{+rCl+x*}
\gpsi & \geq & \max\left\{0, 1 - \frac{1}{2\gamma}\right\} \\
\gx & > & \frac{1}{\gamma \eta} \\
\gP & > & 0
\end{IEEEeqnarray*}
such that $\gpsi + \gx + \gP < 1/2$. We then have
\begin{IEEEeqnarray*}{+rCl+x*}
\gamma & \leq & \frac{1}{2 - 2\gpsi} \\
1 - \gamma \eta \gx & < & 0
\end{IEEEeqnarray*}
and therefore
\begin{IEEEeqnarray*}{+rCl+x*}
m_n & \leq & O\left(n^{\frac{1}{2 - 2\gpsi}}\right) \\
O(nm_n^{-\eta \gx}) & = & o(1) \\
O(m_n^{-\gP}) & = & o(1)~.
\end{IEEEeqnarray*}
Let $\Ch$ be the corresponding constant from \Cref{thm:main:inconsistency}. Since $m_n \to \infty$ and $h_n < o(m_n^{-1})$, there exists an $n_0$ such that for all $n \geq n_0$, we have
\begin{IEEEeqnarray*}{+rCl+x*}
h_n & \leq & \Ch m_n^{-1}~.
\end{IEEEeqnarray*}
Hence, by \Cref{thm:main:inconsistency}, we obtain for all $n \geq n_0$ that $f_{W_k} \in \calF_{\Cx m_n^{-\gx}}$ with probability $\geq 1 - \CP (m_n^{-\gP} + nm_n^{-\eta \gx}) \to 1$ for $n \to \infty$. By assumption (P4), we have $R_{\Pdata, 0}^* > R_{\Pdata}^*$ and by \Cref{lemma:delta_affine_bayes_risk}, there exists $\delta > 0$ such that $R_{\Pdata, \delta}^* > R_{\Pdata}^*$. For $n$ sufficiently large,  we have $\Cx m_n^{-\gx} \leq \delta$ and therefore 
\begin{IEEEeqnarray*}{+rCl+x*}
R_{\Pdata}(f_{W_k}) & \geq & R_{\Pdata, \delta}^* > R_{\Pdata}^*
\end{IEEEeqnarray*}
with probability $\geq 1 - \CP (m_n^{-\gP} + nm_n^{-\eta \gx}) \to 1$ for  $n \to \infty$. This shows inconsistency.
\end{proof}

\section{Miscellaneous} \label{sec:misc}

In this section, we prove a fact that has been mentioned in the main paper.

\begin{lemma} \label{lemma:adding_points}
Let $D = ((x_1, y_1), \hdots, (x_n, y_n)) \in (\bbR \times \bbR)^n$ with $n \geq 1$ and $x_j \neq 0$ for all $j$. Then, by adding three points to $D$, we can achieve that $\bfM_{D, \sigma}$ is invertible for both $\sigma \in \{\pm 1\}$ and that $\psi_{D, q} = 0$.

\begin{proof}
By adding at most one point to $D$, we can ensure that both $D_1$ and $D_{-1}$ are nonempty. Now consider the case of $D_1$ ($D_{-1}$ can be handled analogously). For $x' \equalDef 1 + \max_{(x, y) \in D_1} x$ and $y' \in \bbR$ yet to be specified, consider the data set $\tilde{D} \equalDef D \cup \{(x', y')\}$. Since the kernels of two different matrices $\bfM_{x_j} \matgeq 0$ and $\bfM_{x'} \matgeq 0$ only intersect in zero, we have
\begin{IEEEeqnarray*}{+rCl+x*}
\bfM_{\tilde{D}, 1} & = & \frac{1}{n+1} \left(\bfM_{x'} + \sum_{(x, y) \in D_1} \bfM_x\right) \matgr 0~,
\end{IEEEeqnarray*}
i.e., $\bfM_{\tilde{D}, 1}$ is invertible. Moreover, we have
\begin{IEEEeqnarray*}{+rCl+x*}
\begin{pmatrix}
\popt_{\tilde{D}, 1} \\
\qopt_{\tilde{D}, 1}
\end{pmatrix} = \vopt_{\tilde{D}, 1} & = & \bfM_{\tilde{D}, 1}^{-1} \bfu^0_{\tilde{D}, 1} = \frac{1}{n+1} \left(\bfu_{(x', y')} + \sum_{(x, y) \in D_1} \bfu^0_{(x, y)}\right) \\
& = & \frac{y'}{n+1} \bfM_{\tilde{D}, 1}^{-1} \begin{pmatrix}
x' \\
1
\end{pmatrix} + \frac{n}{n+1} \bfM_{\tilde{D}, 1}^{-1} \bfu^0_{D, 1}
\end{IEEEeqnarray*}
We need to show that we can choose $y'$ such that $\qopt_{\tilde{D}, 1} = 0$. Assume the contrary, which means that there exists $z \in \bbR$ with
\begin{IEEEeqnarray*}{+rCl+x*}
\bfM_{\tilde{D}, 1}^{-1} \begin{pmatrix}
x' \\ 1
\end{pmatrix} & = & \begin{pmatrix}
z' \\ 0
\end{pmatrix}
\end{IEEEeqnarray*}
or, equivalently,
\begin{IEEEeqnarray*}{+rCl+x*}
\begin{pmatrix}
x' \\ 1
\end{pmatrix} = \bfM_{\tilde{D}, 1} \begin{pmatrix}
z' \\ 0
\end{pmatrix} = \frac{z'}{n+1} \left(\begin{pmatrix}
x' \\
1
\end{pmatrix} + \sum_{(x, y) \in D_1} \begin{pmatrix}
x \\ 1
\end{pmatrix}\right)~.
\end{IEEEeqnarray*}
Since $D_1$ is nonempty by assumption and all $(x, y) \in D_1$ satisfy $x < x'$, we obtain the desired contradiction.

Overall, we can therefore satisfy $M_{D, \sigma} \matgr 0$ for both $\sigma \in \{\pm 1\}$ by adding at most point to $D$ and we can then satisfy $\psi_{D, q} = 0$ by adding at most two more points to $D$.
\end{proof}
\end{lemma}

\section{NTK Relation Proofs} \label{sec:ntk_proofs}

\begin{proposition} \label{prop:ntk_relations}
Let $W \in \bbR^{3m+1}$ be a parameter vector that induces the same activation pattern as $W_0$ on $D$, i.e., 
\begin{IEEEeqnarray*}{+rCl+x*}
\sgn(a_i x_j + b_i) & = & \sgn(a_{i, 0} x_j + b_{i, 0})
\end{IEEEeqnarray*}
for all $i \in I, j \in J$, such that $f_{W, \tau, \sgn(x_j)}(x_j) = f_W(x_j)$ for all $j \in J$. By the proof of \Cref{lemma:linearization_region}, this is satisfied e.g.\ for $W \in \calS_{W_0}(\udl{x}_D)$. Then, following the notation from \Cref{sec:ntk} and \Cref{def:derived_quantities}, we have
\begin{IEEEeqnarray*}{+rCl+x*}
\bfK & = & \bfX \bfA \bfX^\top \\
\bfM_D & = & \frac{1}{n} \bfX^\top \bfX \\
\ovl{\bfv} & = & (\bfX^\top \bfX)^{-1} \bfX^\top \ovl{\bff}~,
\end{IEEEeqnarray*}
where $h = 0$ is used in the definition of $\bfA$.
\end{proposition}

\begin{proof}
We first show $\bfK_{jk} = [\bfX \bfA \bfX^\top]_{jk}$ for $j, k \in J$. Recall from \Cref{sec:ntk} and \Cref{def:derived_quantities} that
\begin{IEEEeqnarray*}{+rCl+x*}
\bfX & = & \begin{pmatrix}
x_1 & 0 & 1 & 0 \\
\vdots & \vdots & \vdots & \vdots \\
x_{n'} & 0 & 1 & 0 \\
0 & x_{n'+1} & 0 & 1 \\
\vdots & \vdots & \vdots & \vdots \\
0 & x_n & 0 & 1
\end{pmatrix}~, \\
\bfA & = & \begin{pmatrix}
\Sigma_{1, a^2} + \Sigma_{1, w^2} & 0 & \Sigma_{1, ab} & 0 \\
0 & \Sigma_{-1, a^2} + \Sigma_{-1, a^2} & 0 & \Sigma_{-1, ab} \\
\Sigma_{1, ab} & 0 & 1 + \Sigma_{1, b^2} + \Sigma_{1, w^2} & 1 \\
0 & \Sigma_{-1, ab} & 1 & 1 + \Sigma_{-1, b^2} + \Sigma_{-1, w^2}
\end{pmatrix}~.
\end{IEEEeqnarray*}
We consider several cases:
\begin{itemize}
\item Case 1: $x_j > 0$ and $x_k < 0$. Then, since no neuron is activated for both $x_j$ and $x_k$, 
\begin{IEEEeqnarray*}{+rCl+x*}
K_{jk} & = & \sum_{i \in I} \left(\frac{\partial f_W(x_j)}{\partial a_i} \frac{\partial f_W(x_k)}{\partial a_i} + \frac{\partial f_W(x_j)}{\partial b_i} \frac{\partial f_W(x_k)}{\partial b_i} + \frac{\partial f_W(x_j)}{\partial w_i} \frac{\partial f_W(x_k)}{\partial w_i} \right) \\
&& ~+~ \frac{\partial f_W(x_j)}{\partial c} \frac{\partial f_W(x_k)}{\partial c} \\
& = & 0 + 1 \cdot 1 = 1 = \begin{pmatrix}
x_j \\ 0 \\ 1 \\ 0
\end{pmatrix}^\top \bfA \begin{pmatrix}
0 \\ x_k \\ 0 \\ 1
\end{pmatrix} = [(\bfX \bfA \bfX)^\top]_{jk}~.
\end{IEEEeqnarray*}
\item Case 2: $x_j, x_k > 0$. Then,
\begin{IEEEeqnarray*}{+rCl+x*}
K_{jk} & = & \sum_{i \in I_1} \left(\frac{\partial f_W(x_j)}{\partial a_i} \frac{\partial f_W(x_k)}{\partial a_i} + \frac{\partial f_W(x_j)}{\partial b_i} \frac{\partial f_W(x_k)}{\partial b_i} + \frac{\partial f_W(x_j)}{\partial w_i} \frac{\partial f_W(x_k)}{\partial w_i} \right) \\
&& ~+~ \frac{\partial f_W(x_j)}{\partial c} \frac{\partial f_W(x_k)}{\partial c} \\
& = & \sum_{i \in I_1} \left(w_i^2 x_j x_k + w_i^2 + (a_i x_j + b_i)(a_i x_k + b_i)\right) + 1 \\
& = & \Sigma_{1, w^2} (x_j x_k + 1) + \Sigma_{1, a^2} x_j x_k + \Sigma_{1, ab} (x_j + x_k) + \Sigma_{1, b^2} + 1 \\
& = & \begin{pmatrix}
x_j \\ 0 \\ 1 \\ 0
\end{pmatrix}^\top \bfA \begin{pmatrix}
x_k \\ 0 \\ 1 \\ 0
\end{pmatrix} = [(\bfX \bfA \bfX)^\top]_{jk}~.
\end{IEEEeqnarray*}
\item Case 3: $x_j, x_k < 0$. This can be handled just like Case 2.
\end{itemize}

The identity $\bfM_D = \frac{1}{n} \bfX^\top \bfX$ is easy to verify using the definitions of $\bfM_D$ and $\bfX$.

It remains to show the last identity. For $j \in J$ and $\sigma = \sgn(x_j)$, we have
\begin{IEEEeqnarray*}{+rCl+x*}
(\bfX \bfv)_j & = & p_\sigma x_j + q_\sigma \stackrel{\text{\Cref{lemma:deriv:linear_relations}}}{=} f_{W, \bftau, \sigma}(x_j) = f_W(x_j)~.
\end{IEEEeqnarray*}
Therefore,
\begin{IEEEeqnarray*}{+rCl+x*}
(\bfX^\top \bfX)^{-1} \bfX^\top \begin{pmatrix}
f_W(x_1) \\ \vdots \\ f_W(x_n)
\end{pmatrix} & = & (\bfX^\top \bfX)^{-1} \bfX^\top \bfX \bfv = \bfv~. \IEEEyesnumber \label{eq:base_change_v}
\end{IEEEeqnarray*}
Using \Cref{def:derived_quantities} and \Cref{def:main:quantities}, we also obtain
\begin{IEEEeqnarray*}{+rCl+x*}
(\bfX^\top \bfX)^{-1} & = & (n\bfM_D)^{-1} = (n\tilde{\bfP} \tilde{\bfM}_D \tilde{\bfP})^{-1} = \frac{1}{n} \tilde{\bfP} \tilde{\bfM}_D^{-1} \tilde{\bfP} \\
\tilde{\bfM}_D^{-1} & = & \begin{pmatrix}
\bfM_{D, 1} \\
& \bfM_{D, -1}
\end{pmatrix}^{-1} = \begin{pmatrix}
\bfM_{D, 1}^{-1} \\
& \bfM_{D, -1}^{-1}
\end{pmatrix} \\
\frac{1}{n} \tilde{\bfP} \bfX^\top \begin{pmatrix}
y_1 \\ \vdots \\ y_n
\end{pmatrix} & = & \frac{1}{n} \tilde{\bfP} \begin{pmatrix}
\sum_{j \in J_1} x_j y_j \\
\sum_{j \in J_{-1}} x_j y_j \\
\sum_{j \in J_1} y_j \\
\sum_{j \in J_{-1}} y_j
\end{pmatrix} = \frac{1}{n} \begin{pmatrix}
\sum_{j \in J_1} x_j y_j \\
\sum_{j \in J_1} y_j \\
\sum_{j \in J_{-1}} x_j y_j \\
\sum_{j \in J_{-1}} y_j
\end{pmatrix} = \begin{pmatrix}
\hat{\bfu}^0_{D, 1} \\ \hat{\bfu}^0_{D, -1}
\end{pmatrix}
\end{IEEEeqnarray*}
and therefore
\begin{IEEEeqnarray*}{+rCl+x*}
(\bfX^\top \bfX)^{-1} \bfX^\top \begin{pmatrix}
y_1 \\ \vdots \\ y_n
\end{pmatrix} & = & \tilde{\bfP} \begin{pmatrix}
\bfM_{D, 1}^{-1} \\
& \bfM_{D, -1}^{-1}
\end{pmatrix} \begin{pmatrix}
\hat{\bfu}^0_{D, 1} \\ \hat{\bfu}^0_{D, -1}
\end{pmatrix} = \tilde{\bfP} \begin{pmatrix}
\vopt_{D, 1} \\ \vopt_{D, -1}
\end{pmatrix} = \tilde{\bfP} \tildevopt \\
& = & \vopt~. \IEEEyesnumber \label{eq:base_change_vopt}
\end{IEEEeqnarray*}
Subtracting \eqref{eq:base_change_vopt} from \eqref{eq:base_change_v} yields the desired identity
\begin{IEEEeqnarray*}{+rCl+x*}
(\bfX^\top \bfX)^{-1} \bfX^\top \ovl{\bff} & = & \ovl{\bfv}~. & \qedhere
\end{IEEEeqnarray*}
\end{proof}

\section{Experimental Details for \Cref{sec:practical}} \label{sec:appendix_experiments}

\begin{table}[tb]
\centering
{\tiny
\setlength{\tabcolsep}{4pt}
\begin{tabular}{ccccccccc}%
Param/Opt/Bias-init & $P^{\mathrm{data}}_{\mathrm{ex}}$ & $P^{\mathrm{data}}_{1}$ & $P^{\mathrm{data}}_{2}$ & $P^{\mathrm{data}}_{4}$ & $P^{\mathrm{data}}_{8}$ & $P^{\mathrm{data}}_{16}$ & $P^{\mathrm{data}}_{32}$ & $P^{\mathrm{data}}_{64}$\\
\hline
He/SGD/Zero & \cellcolor{red!100} $9288 \pm 53$ & \cellcolor{red!100} $1020 \pm 95$ & \cellcolor{red!92} $1266 \pm 67$ & \cellcolor{red!63} $1968 \pm 25$ & \cellcolor{red!46} $2424 \pm 21$ & \cellcolor{red!41} $2877 \pm 26$ & \cellcolor{red!38} $3293 \pm 23$ & \cellcolor{red!46} $3846 \pm 48$\\
He/SGD/PyTorch & \cellcolor{red!86} $7226 \pm 42$ & \cellcolor{red!83} $500 \pm 3$ & \cellcolor{red!79} $1017 \pm 7$ & \cellcolor{red!43} $1638 \pm 20$ & \cellcolor{red!41} $2319 \pm 19$ & \cellcolor{red!43} $2929 \pm 19$ & \cellcolor{red!39} $3317 \pm 21$ & \cellcolor{red!46} $3859 \pm 44$\\
He/SGD/$U(1)$ & \cellcolor{red!84} $6986 \pm 38$ & \cellcolor{red!82} $482 \pm 4$ & \cellcolor{red!74} $928 \pm 5$ & \cellcolor{red!37} $1540 \pm 6$ & \cellcolor{red!38} $2251 \pm 10$ & \cellcolor{red!37} $2753 \pm 10$ & \cellcolor{red!38} $3268 \pm 28$ & \cellcolor{red!46} $3843 \pm 35$\\
He/SGD/$U_+(1)$ & \cellcolor{red!92} $8057 \pm 21$ & \cellcolor{red!83} $507 \pm 5$ & \cellcolor{red!84} $1102 \pm 8$ & \cellcolor{red!80} $2325 \pm 11$ & \cellcolor{red!86} $3556 \pm 14$ & \cellcolor{red!89} $4484 \pm 15$ & \cellcolor{red!91} $5315 \pm 16$ & \cellcolor{red!90} $5750 \pm 17$\\
He/SGD/$U_-(1)$ & \cellcolor{red!50} $3768 \pm 65$ & \cellcolor{red!66} $233 \pm 4$ & \cellcolor{red!51} $637 \pm 5$ & \cellcolor{red!19} $1300 \pm 5$ & \cellcolor{red!22} $1948 \pm 8$ & \cellcolor{red!24} $2445 \pm 9$ & \cellcolor{red!16} $2686 \pm 9$ & \cellcolor{red!15} $2896 \pm 14$\\
He/SGD/$U_{k-}(1)$ & \cellcolor{red!25} $2385 \pm 58$ & \cellcolor{red!48} $109 \pm 3$ & \cellcolor{red!49} $612 \pm 6$ & \cellcolor{red!20} $1320 \pm 6$ & \cellcolor{red!20} $1910 \pm 7$ & \cellcolor{red!22} $2402 \pm 9$ & \cellcolor{red!15} $2661 \pm 9$ & \cellcolor{red!14} $2882 \pm 15$\\
He/SGD/$X_{k-}$ & \cellcolor{red!5} $1641 \pm 11$ & \cellcolor{red!49} $115 \pm 3$ & \cellcolor{red!45} $576 \pm 5$ & \cellcolor{red!19} $1310 \pm 6$ & \cellcolor{red!25} $1996 \pm 9$ & \cellcolor{red!29} $2554 \pm 9$ & \cellcolor{red!23} $2862 \pm 9$ & \cellcolor{red!23} $3111 \pm 15$\\
He/SGD/He+5 & \cellcolor{red!79} $6404 \pm 32$ & \cellcolor{red!83} $490 \pm 4$ & \cellcolor{red!77} $983 \pm 11$ & \cellcolor{red!45} $1674 \pm 9$ & \cellcolor{red!51} $2540 \pm 13$ & \cellcolor{red!50} $3117 \pm 11$ & \cellcolor{red!44} $3450 \pm 12$ & \cellcolor{red!49} $3940 \pm 17$\\
He/SGDM/Zero & \cellcolor{red!89} $7678 \pm 417$ & \cellcolor{red!86} $554 \pm 6$ & \cellcolor{red!72} $897 \pm 5$ & \cellcolor{red!60} $1930 \pm 14$ & \cellcolor{red!84} $3490 \pm 154$ & \cellcolor{red!62} $3496 \pm 71$ & \cellcolor{red!51} $3694 \pm 46$ & \cellcolor{red!41} $3691 \pm 45$\\
He/SGDM/PyTorch & \cellcolor{red!38} $3044 \pm 68$ & \cellcolor{red!71} $291 \pm 6$ & \cellcolor{red!66} $821 \pm 4$ & \cellcolor{red!37} $1550 \pm 8$ & \cellcolor{red!55} $2659 \pm 33$ & \cellcolor{red!60} $3408 \pm 83$ & \cellcolor{red!46} $3539 \pm 41$ & \cellcolor{red!42} $3703 \pm 43$\\
He/SGDM/$U(1)$ & \cellcolor{red!29} $2549 \pm 58$ & \cellcolor{red!52} $127 \pm 3$ & \cellcolor{red!45} $576 \pm 5$ & \cellcolor{red!16} $1270 \pm 6$ & \cellcolor{red!24} $1982 \pm 7$ & \cellcolor{red!29} $2555 \pm 9$ & \cellcolor{red!23} $2858 \pm 11$ & \cellcolor{red!24} $3138 \pm 15$\\
He/SGDM/$U_+(1)$ & \cellcolor{red!47} $3595 \pm 54$ & \cellcolor{red!59} $175 \pm 3$ & \cellcolor{red!46} $581 \pm 4$ & \cellcolor{red!9} $1192 \pm 6$ & \cellcolor{red!28} $2052 \pm 8$ & \cellcolor{red!35} $2714 \pm 11$ & \cellcolor{red!30} $3053 \pm 11$ & \cellcolor{red!33} $3425 \pm 14$\\
He/SGDM/$U_-(1)$ & \cellcolor{red!5} $1649 \pm 7$ & \cellcolor{red!28} $45 \pm 2$ & \cellcolor{red!18} $364 \pm 5$ & \cellcolor{red!14} $1243 \pm 6$ & \cellcolor{red!22} $1948 \pm 8$ & \cellcolor{red!23} $2423 \pm 9$ & \cellcolor{red!15} $2657 \pm 11$ & \cellcolor{red!10} $2782 \pm 14$\\
He/SGDM/$U_{k-}(1)$ & \cellcolor{red!5} $1644 \pm 7$ & \cellcolor{red!18} $29 \pm 1$ & \cellcolor{red!25} $412 \pm 9$ & \cellcolor{red!20} $1313 \pm 7$ & \cellcolor{red!20} $1911 \pm 7$ & \cellcolor{red!21} $2385 \pm 9$ & \cellcolor{red!14} $2631 \pm 11$ & \cellcolor{red!10} $2769 \pm 15$\\
He/SGDM/$X_{k-}$ & \cellcolor{red!0} $1520 \pm 5$ & \cellcolor{red!20} $32 \pm 1$ & \cellcolor{red!12} $331 \pm 6$ & \cellcolor{red!16} $1264 \pm 7$ & \cellcolor{red!25} $1998 \pm 9$ & \cellcolor{red!27} $2511 \pm 17$ & \cellcolor{red!20} $2785 \pm 9$ & \cellcolor{red!18} $2982 \pm 17$\\
He/SGDM/He+5 & \cellcolor{red!40} $3108 \pm 97$ & \cellcolor{red!65} $230 \pm 7$ & \cellcolor{red!65} $797 \pm 6$ & \cellcolor{red!31} $1463 \pm 7$ & \cellcolor{red!43} $2360 \pm 10$ & \cellcolor{red!44} $2946 \pm 11$ & \cellcolor{red!36} $3215 \pm 11$ & \cellcolor{red!31} $3360 \pm 18$\\
He/Adam/Zero & \cellcolor{red!2} $1554 \pm 6$ & \cellcolor{red!39} $72 \pm 2$ & \cellcolor{red!31} $449 \pm 7$ & \cellcolor{red!13} $1232 \pm 5$ & \cellcolor{red!13} $1779 \pm 7$ & \cellcolor{red!26} $2496 \pm 9$ & \cellcolor{red!27} $2973 \pm 10$ & \cellcolor{red!28} $3265 \pm 15$\\
He/Adam/PyTorch & \cellcolor{red!1} $1544 \pm 6$ & \cellcolor{red!32} $53 \pm 2$ & \cellcolor{red!28} $430 \pm 7$ & \cellcolor{red!13} $1234 \pm 6$ & \cellcolor{red!13} $1786 \pm 8$ & \cellcolor{red!24} $2458 \pm 9$ & \cellcolor{red!27} $2963 \pm 10$ & \cellcolor{red!28} $3266 \pm 16$\\
He/Adam/$U(1)$ & \cellcolor{red!1} $1550 \pm 6$ & \cellcolor{red!26} $41 \pm 3$ & \cellcolor{red!24} $405 \pm 6$ & \cellcolor{red!13} $1233 \pm 6$ & \cellcolor{red!13} $1791 \pm 7$ & \cellcolor{red!13} $2219 \pm 9$ & \cellcolor{red!14} $2632 \pm 9$ & \cellcolor{red!16} $2918 \pm 15$\\
He/Adam/$U_+(1)$ & \cellcolor{red!4} $1632 \pm 15$ & \cellcolor{red!43} $87 \pm 8$ & \cellcolor{red!41} $536 \pm 10$ & \cellcolor{red!7} $1160 \pm 7$ & \cellcolor{red!18} $1874 \pm 10$ & \cellcolor{red!28} $2533 \pm 13$ & \cellcolor{red!30} $3055 \pm 11$ & \cellcolor{red!32} $3395 \pm 19$\\
He/Adam/$U_-(1)$ & \cellcolor{red!0} $1521 \pm 6$ & \cellcolor{red!8} $19 \pm 1$ & \cellcolor{red!2} $279 \pm 4$ & \cellcolor{red!5} $1141 \pm 7$ & \cellcolor{red!10} $1733 \pm 7$ & \cellcolor{red!5} $2061 \pm 8$ & \cellcolor{red!3} $2377 \pm 9$ & \cellcolor{red!2} $2584 \pm 13$\\
He/Adam/$U_{k-}(1)$ & \cellcolor{red!0} $1512 \pm 5$ & \cellcolor{red!0} $13 \pm 1$ & \cellcolor{red!0} $268 \pm 4$ & \cellcolor{red!7} $1160 \pm 7$ & \cellcolor{red!9} $1713 \pm 7$ & \cellcolor{red!5} $2057 \pm 8$ & \cellcolor{red!3} $2376 \pm 9$ & \cellcolor{red!2} $2582 \pm 13$\\
He/Adam/$X_{k-}$ & \cellcolor{red!0} $1498 \pm 4$ & \cellcolor{red!0} $13 \pm 1$ & \cellcolor{red!0} $266 \pm 4$ & \cellcolor{red!7} $1162 \pm 7$ & \cellcolor{red!12} $1771 \pm 8$ & \cellcolor{red!10} $2160 \pm 8$ & \cellcolor{red!10} $2533 \pm 9$ & \cellcolor{red!11} $2788 \pm 15$\\
He/Adam/He+5 & \cellcolor{red!1} $1534 \pm 5$ & \cellcolor{red!30} $50 \pm 2$ & \cellcolor{red!27} $425 \pm 6$ & \cellcolor{red!14} $1249 \pm 7$ & \cellcolor{red!13} $1785 \pm 7$ & \cellcolor{red!19} $2339 \pm 9$ & \cellcolor{red!22} $2838 \pm 11$ & \cellcolor{red!24} $3134 \pm 16$\\
NTK/SGD/Zero & \cellcolor{red!90} $7813 \pm 41$ & \cellcolor{red!84} $529 \pm 3$ & \cellcolor{red!66} $821 \pm 4$ & \cellcolor{red!9} $1186 \pm 4$ & \cellcolor{red!5} $1651 \pm 6$ & \cellcolor{red!5} $2057 \pm 8$ & \cellcolor{red!2} $2357 \pm 8$ & \cellcolor{red!4} $2633 \pm 11$\\
NTK/SGD/PyTorch & \cellcolor{red!87} $7371 \pm 30$ & \cellcolor{red!83} $501 \pm 3$ & \cellcolor{red!66} $818 \pm 4$ & \cellcolor{red!9} $1182 \pm 5$ & \cellcolor{red!4} $1644 \pm 6$ & \cellcolor{red!5} $2055 \pm 7$ & \cellcolor{red!2} $2348 \pm 8$ & \cellcolor{red!5} $2644 \pm 12$\\
NTK/SGD/$U(1)$ & \cellcolor{red!83} $6909 \pm 26$ & \cellcolor{red!76} $367 \pm 5$ & \cellcolor{red!57} $705 \pm 4$ & \cellcolor{red!9} $1182 \pm 5$ & \cellcolor{red!7} $1689 \pm 7$ & \cellcolor{red!9} $2123 \pm 8$ & \cellcolor{red!5} $2421 \pm 8$ & \cellcolor{red!12} $2810 \pm 15$\\
NTK/SGD/$U_+(1)$ & \cellcolor{red!87} $7416 \pm 25$ & \cellcolor{red!77} $386 \pm 9$ & \cellcolor{red!67} $823 \pm 4$ & \cellcolor{red!18} $1287 \pm 9$ & \cellcolor{red!14} $1793 \pm 11$ & \cellcolor{red!12} $2194 \pm 15$ & \cellcolor{red!6} $2454 \pm 13$ & \cellcolor{red!19} $3002 \pm 23$\\
NTK/SGD/$U_-(1)$ & \cellcolor{red!75} $5903 \pm 31$ & \cellcolor{red!77} $378 \pm 4$ & \cellcolor{red!56} $691 \pm 4$ & \cellcolor{red!3} $1125 \pm 5$ & \cellcolor{red!6} $1672 \pm 6$ & \cellcolor{red!7} $2102 \pm 9$ & \cellcolor{red!4} $2408 \pm 8$ & \cellcolor{red!7} $2692 \pm 13$\\
NTK/SGD/$U_{k-}(1)$ & \cellcolor{red!72} $5618 \pm 33$ & \cellcolor{red!58} $166 \pm 4$ & \cellcolor{red!51} $629 \pm 4$ & \cellcolor{red!3} $1124 \pm 5$ & \cellcolor{red!5} $1662 \pm 6$ & \cellcolor{red!7} $2096 \pm 9$ & \cellcolor{red!4} $2403 \pm 8$ & \cellcolor{red!7} $2688 \pm 13$\\
NTK/SGD/$X_{k-}$ & \cellcolor{red!60} $4505 \pm 35$ & \cellcolor{red!57} $159 \pm 5$ & \cellcolor{red!53} $661 \pm 5$ & \cellcolor{red!4} $1136 \pm 4$ & \cellcolor{red!6} $1663 \pm 6$ & \cellcolor{red!7} $2088 \pm 7$ & \cellcolor{red!4} $2399 \pm 8$ & \cellcolor{red!7} $2693 \pm 13$\\
NTK/SGD/He+5 & \cellcolor{red!81} $6659 \pm 26$ & \cellcolor{red!80} $446 \pm 4$ & \cellcolor{red!64} $786 \pm 4$ & \cellcolor{red!9} $1189 \pm 5$ & \cellcolor{red!5} $1658 \pm 6$ & \cellcolor{red!7} $2084 \pm 8$ & \cellcolor{red!2} $2353 \pm 8$ & \cellcolor{red!6} $2677 \pm 13$\\
NTK/SGDM/Zero & \cellcolor{red!73} $5761 \pm 230$ & \cellcolor{red!83} $500 \pm 4$ & \cellcolor{red!67} $826 \pm 4$ & \cellcolor{red!14} $1239 \pm 7$ & \cellcolor{red!10} $1727 \pm 8$ & \cellcolor{red!8} $2113 \pm 11$ & \cellcolor{red!7} $2458 \pm 13$ & \cellcolor{red!7} $2694 \pm 17$\\
NTK/SGDM/PyTorch & \cellcolor{red!17} $2061 \pm 68$ & \cellcolor{red!71} $297 \pm 13$ & \cellcolor{red!66} $815 \pm 5$ & \cellcolor{red!12} $1215 \pm 5$ & \cellcolor{red!10} $1728 \pm 8$ & \cellcolor{red!9} $2132 \pm 10$ & \cellcolor{red!8} $2484 \pm 13$ & \cellcolor{red!8} $2722 \pm 21$\\
NTK/SGDM/$U(1)$ & \cellcolor{red!9} $1768 \pm 42$ & \cellcolor{red!69} $268 \pm 10$ & \cellcolor{red!57} $707 \pm 5$ & \cellcolor{red!4} $1136 \pm 5$ & \cellcolor{red!6} $1670 \pm 6$ & \cellcolor{red!8} $2103 \pm 8$ & \cellcolor{red!5} $2417 \pm 9$ & \cellcolor{red!4} $2624 \pm 13$\\
NTK/SGDM/$U_+(1)$ & \cellcolor{red!37} $2954 \pm 45$ & \cellcolor{red!76} $368 \pm 9$ & \cellcolor{red!60} $732 \pm 4$ & \cellcolor{red!6} $1150 \pm 6$ & \cellcolor{red!9} $1715 \pm 6$ & \cellcolor{red!11} $2166 \pm 8$ & \cellcolor{red!6} $2454 \pm 8$ & \cellcolor{red!4} $2629 \pm 13$\\
NTK/SGDM/$U_-(1)$ & \cellcolor{red!3} $1602 \pm 39$ & \cellcolor{red!51} $121 \pm 9$ & \cellcolor{red!45} $572 \pm 7$ & \cellcolor{red!4} $1133 \pm 5$ & \cellcolor{red!4} $1643 \pm 7$ & \cellcolor{red!5} $2055 \pm 9$ & \cellcolor{red!3} $2378 \pm 9$ & \cellcolor{red!4} $2616 \pm 15$\\
NTK/SGDM/$U_{k-}(1)$ & \cellcolor{red!2} $1564 \pm 6$ & \cellcolor{red!37} $68 \pm 2$ & \cellcolor{red!38} $508 \pm 7$ & \cellcolor{red!4} $1128 \pm 5$ & \cellcolor{red!4} $1638 \pm 7$ & \cellcolor{red!5} $2049 \pm 8$ & \cellcolor{red!3} $2377 \pm 9$ & \cellcolor{red!3} $2611 \pm 14$\\
NTK/SGDM/$X_{k-}$ & \cellcolor{red!1} $1552 \pm 6$ & \cellcolor{red!38} $70 \pm 2$ & \cellcolor{red!40} $523 \pm 7$ & \cellcolor{red!6} $1157 \pm 4$ & \cellcolor{red!4} $1637 \pm 6$ & \cellcolor{red!5} $2054 \pm 7$ & \cellcolor{red!3} $2388 \pm 8$ & \cellcolor{red!4} $2630 \pm 14$\\
NTK/SGDM/He+5 & \cellcolor{red!19} $2126 \pm 77$ & \cellcolor{red!68} $256 \pm 13$ & \cellcolor{red!64} $784 \pm 5$ & \cellcolor{red!10} $1196 \pm 5$ & \cellcolor{red!7} $1689 \pm 7$ & \cellcolor{red!8} $2106 \pm 9$ & \cellcolor{red!6} $2447 \pm 8$ & \cellcolor{red!5} $2644 \pm 12$\\
NTK/Adam/Zero & \cellcolor{red!3} $1605 \pm 7$ & \cellcolor{red!34} $59 \pm 3$ & \cellcolor{red!32} $461 \pm 8$ & \cellcolor{red!12} $1219 \pm 5$ & \cellcolor{red!5} $1655 \pm 6$ & \cellcolor{red!9} $2124 \pm 7$ & \cellcolor{red!8} $2491 \pm 10$ & \cellcolor{red!9} $2747 \pm 14$\\
NTK/Adam/PyTorch & \cellcolor{red!3} $1600 \pm 9$ & \cellcolor{red!29} $47 \pm 2$ & \cellcolor{red!29} $440 \pm 7$ & \cellcolor{red!12} $1216 \pm 5$ & \cellcolor{red!6} $1677 \pm 6$ & \cellcolor{red!10} $2145 \pm 8$ & \cellcolor{red!11} $2568 \pm 9$ & \cellcolor{red!12} $2818 \pm 15$\\
NTK/Adam/$U(1)$ & \cellcolor{red!4} $1628 \pm 7$ & \cellcolor{red!37} $67 \pm 4$ & \cellcolor{red!33} $466 \pm 8$ & \cellcolor{red!4} $1132 \pm 6$ & \cellcolor{red!8} $1705 \pm 7$ & \cellcolor{red!11} $2175 \pm 9$ & \cellcolor{red!12} $2591 \pm 9$ & \cellcolor{red!13} $2852 \pm 14$\\
NTK/Adam/$U_+(1)$ & \cellcolor{red!5} $1664 \pm 8$ & \cellcolor{red!40} $78 \pm 5$ & \cellcolor{red!40} $530 \pm 9$ & \cellcolor{red!13} $1237 \pm 8$ & \cellcolor{red!17} $1846 \pm 9$ & \cellcolor{red!25} $2471 \pm 12$ & \cellcolor{red!23} $2857 \pm 11$ & \cellcolor{red!23} $3108 \pm 17$\\
NTK/Adam/$U_-(1)$ & \cellcolor{red!2} $1580 \pm 7$ & \cellcolor{red!27} $44 \pm 2$ & \cellcolor{red!18} $364 \pm 6$ & \cellcolor{red!0} $1093 \pm 6$ & \cellcolor{red!1} $1594 \pm 6$ & \cellcolor{red!1} $1984 \pm 7$ & \cellcolor{red!1} $2334 \pm 8$ & \cellcolor{red!0} $2539 \pm 13$\\
NTK/Adam/$U_{k-}(1)$ & \cellcolor{red!2} $1559 \pm 7$ & \cellcolor{red!13} $24 \pm 1$ & \cellcolor{red!11} $322 \pm 5$ & \cellcolor{red!0} $1085 \pm 6$ & \cellcolor{red!0} $1585 \pm 6$ & \cellcolor{red!1} $1974 \pm 7$ & \cellcolor{red!1} $2333 \pm 9$ & \cellcolor{red!0} $2539 \pm 12$\\
NTK/Adam/$X_{k-}$ & \cellcolor{red!1} $1528 \pm 6$ & \cellcolor{red!13} $23 \pm 1$ & \cellcolor{red!15} $348 \pm 5$ & \cellcolor{red!2} $1116 \pm 5$ & \cellcolor{red!3} $1625 \pm 6$ & \cellcolor{red!3} $2023 \pm 8$ & \cellcolor{red!2} $2360 \pm 8$ & \cellcolor{red!2} $2583 \pm 14$\\
NTK/Adam/He+5 & \cellcolor{red!3} $1594 \pm 6$ & \cellcolor{red!27} $43 \pm 2$ & \cellcolor{red!30} $447 \pm 7$ & \cellcolor{red!10} $1203 \pm 4$ & \cellcolor{red!7} $1691 \pm 7$ & \cellcolor{red!10} $2159 \pm 9$ & \cellcolor{red!12} $2589 \pm 9$ & \cellcolor{red!13} $2852 \pm 15$\\
\end{tabular}
}
\caption{For different combinations of data distribution, parameterization, optimizer and bias initialization, this table shows the mean RMSE over 100 runs and the estimated standard deviation of the mean estimator, multiplied by $10^4$ and rounded to the nearest integer, respectively. Cell colors are interpolated on a logarithmic scale from white (best result in column) to red (worst result in column). The bias initialization methods He+5 and $X_{k-}$ and the distribution $\Pdata_{\mathrm{ex}}$ are explained in \Cref{sec:appendix_experiments}, while the other terms are explained in \Cref{sec:practical}. 
} \label{tab:experiment_results}
\end{table}

In \Cref{tab:experiment_results}, we provide the detailed experimental results for the plots shown in \Cref{sec:practical} and some more results. In particular, we provide results for two other bias initializations:
\begin{itemize}
\item Random point combination (\textbf{He+5}): This is one of the variants of the initialization methods proposed by \cite{steinwart_sober_2019}, given by $b_i = -\sum_{k=1}^5 \lambda_k \langle \bfa_i, \bfx_{j_k}\rangle$, where $j_k$ are random indices and $\lambda$ is randomly drawn from a simplex.
\item Negative random point kink initialization ($X_{k-}$): Here, we choose $b_i = -\langle \bfa_i, \bfx \rangle$, where $\bfx$ is a random data point. If $b_i$ is positive, we update $b_i \leftarrow -b_i$ and $\bfa_i \leftarrow -\bfa_i$, such that $b_i$ becomes negative and the kink hyperplane does not change.
\end{itemize}

We also provide results for the data-generating distribution $\Pdata_{\mathrm{ex}}$ used in \Cref{fig:nn_ex}: We sample
\begin{IEEEeqnarray*}{+rCl+x*}
x & = & \sqrt{\frac{196}{105}} \sigma \beta~, \\
y & = & \frac{1}{0.727} \left(\cos\left(7\pi\left(x - \frac{x}{\sqrt{1+x^2}}\right)\right) + \frac{1}{5}x^2 + \frac{1}{10}\varepsilon + 0.074\right)~,
\end{IEEEeqnarray*}
where $\sigma \sim \calU\{-1, 1\}$, $\beta \sim \Beta(5, 2)$ and $\varepsilon \sim \calN(0, 1)$ are independent. The factor $\sqrt{196/105}$ is chosen such that $\Var(x) = 1$. The parameter $5$ in the Beta distribution ensures $\eta = 5$ in the sense of \Cref{ass:P}. The distribution of $y$ is designed to yield visually interesting results, make $y$ approximately normalized and ensure $\qopt_{\Pdata, \pm 1} \approx 0$.

\vskip 0.2in
\bibliography{zotero_references,other_references}

\end{document}